\documentclass[12pt]{article}
\usepackage{amsmath, amssymb, amsthm}
\usepackage{booktabs,longtable}
\usepackage{geometry}
\usepackage{graphicx}
\usepackage{setspace}
\usepackage{hyperref}
\usepackage[authoryear,round]{natbib}
\setcitestyle{aysep={,},citesep={;}}
\usepackage{multirow}
\usepackage{array}
\usepackage{bm}
\usepackage{mathrsfs}
\usepackage{caption}
\usepackage{subcaption}
\usepackage{float}
\usepackage{enumitem}
\usepackage{etoolbox}
\usepackage{needspace}
\usepackage{placeins}
\usepackage{xcolor}

\floatstyle{ruled}
\newfloat{algorithm}{htbp}{loa}
\floatname{algorithm}{Algorithm}
\floatstyle{plain}

\newtheorem{theorem}{Theorem}[section]
\newtheorem{lemma}[theorem]{Lemma}
\newtheorem{corollary}[theorem]{Corollary}
\newtheorem{proposition}[theorem]{Proposition}

\newtheorem{remark}[theorem]{Remark}

\newtheorem{example}[theorem]{Example}

\AtBeginEnvironment{theorem}{\Needspace{5\baselineskip}}
\AtBeginEnvironment{lemma}{\Needspace{5\baselineskip}}
\AtBeginEnvironment{proposition}{\Needspace{7\baselineskip}}
\AtBeginEnvironment{corollary}{\Needspace{6\baselineskip}}
\AtBeginEnvironment{remark}{\Needspace{4\baselineskip}}

\newcommand{\E}{\mathbb E}
\newcommand{\Pp}{\mathbb P}
\newcommand{\R}{\mathbb R}
\newcommand{\ind}{\mathbf 1}
\newcommand{\KL}{\operatorname{KL}}
\newcommand{\supp}{\operatorname{supp}}

\newcommand{\nrm}[1]{\left\lVert#1\right\rVert}

\newcommand{\eventref}[1]{\hyperlink{event:#1}{E_{#1}}}
\newcommand{\eventdef}[1]{\hypertarget{event:#1}{E_{#1}}}
\hypersetup{hidelinks,pdftitle={High-dimensional transfer learning with unaligned ordinal responses},pdfauthor={Ruopei Fu and Ying Yang}}
\begin{document}
\begin{center}
\begin{Large}
High-dimensional transfer learning with unaligned ordinal responses
\end{Large}
\par\vspace{0.7em}
Ruopei Fu\textsuperscript{1} \quad and \quad Ying Yang\textsuperscript{2}
\par\vspace{0.35em}
{\small \textsuperscript{1}Department of Mathematical Sciences, Tsinghua University\par}
{\small \textsuperscript{2}Department of Statistics and Data Science, Tsinghua University\par}
\vspace{0.35em}
{\small \href{mailto:frp23@mails.tsinghua.edu.cn}{frp23@mails.tsinghua.edu.cn}\quad
\href{mailto:yangying@tsinghua.edu.cn}{yangying@tsinghua.edu.cn}\par}
\end{center}

\begin{abstract}
In this work, we study high-dimensional transfer learning with unaligned ordinal responses, allowing different domain-specific regression coefficients, cumulative links and thresholds. We develop the sample-split and full-target algorithms that pool source and target likelihoods and then estimate a target correction. Under suitable conditions, our algorithms achieve minimax-optimal rates for parameter estimation and excess risk. Simulations favor the full-target algorithm over the sample-split algorithm for finite-sample estimation and prediction. An application to a medical dataset further demonstrates the predictive gains of our full-target algorithm over the target-only and pooled baselines when transferring information from the five-category source data to binary hospital mortality.
\end{abstract}

\section{Introduction}

Transfer learning aims to improve the performance on a target task by transferring knowledge from different but similar tasks (sources). It has been successfully applied to a variety of fields such as computer vision \citep{li2013computer}, speech recognition \citep{wang2015speech}, remote sensing \citep{zhao2017geo} and medical imaging \citep{kim2022medical}.

The theoretical properties of transfer learning have also been extensively studied under different statistical models, such as binary classification \citep{tcai2021,kpotufe2021}, high-dimensional linear models \citep{li2022linear,bastani2021}, generalized linear models \citep[GLMs;][]{tian2023,li2024glm} and functional regression \citep{ma2023rkhs,lin2022functional}. In these works, the source and target tasks are generally formulated under a common statistical model, while the transferability is characterized through the similarity between their underlying parameters or distributions.

Several studies consider source-target heterogeneity. \citet{zhao2023heterogeneous} considered heterogeneous transfer learning for high-dimensional generalized linear models with disparate datasets. \citet{gu2025angle} developed a robust angle-based transfer learning procedure that allows transfer based on similarities in parameter directions using source summary statistics. \citet{peng2025hetero} considered heterogeneous high-dimensional linear regression models with dataset-specific nuisance parameters. Their framework allows heterogeneity across datasets while retaining a common regression structure for the parameter of interest. \citet{chang2024feature} and \citet{zheng2026hfs} investigated heterogeneous feature spaces, where the covariates observed in the source and target domains do not coincide. These studies primarily address heterogeneity in regression parameters, nuisance structures, or feature spaces.

In many real-world applications, however, the source and target responses may be structurally heterogeneous. An abundant source dataset may record fine-grained outcomes, whereas the target task may observe only a coarse outcome. For example, a source study may record detailed disease stages while the target task concerns a binary outcome such as mortality or recurrence. Similarly, fine-grained credit ratings may be available in a source dataset, whereas the target task focuses on a coarse default outcome. Our medical data application in Section \ref{sec:Real-Data} pairs a five-level outcome with binary hospital mortality. Such settings motivate transfer between response spaces with different ordinal structures.

Fewer articles have considered heterogeneity in the response or label space. \citet{shi2010transfer} developed Heterogeneous Spectral Mapping (HeMap), which considers domains with different input and output spaces and seeks a shared latent representation to facilitate cross-domain transfer. Such approaches are primarily based on representation learning or cross-domain mappings. More recently, \citet{zheng2026discrete} studied transfer learning for high-dimensional discrete responses within the generalized linear model framework and proposed an importance-weighted procedure based on source-observation-level conditional probability ratios. 

Despite these advances, a high-dimensional statistical framework for transfer learning between \emph{unaligned ordinal responses} remains largely unexplored.
In this paper, we consider transfer learning between structurally heterogeneous ordinal response models. Specifically, the source and target outcomes are modeled by separate cumulative-link models with domain-specific regression coefficients, cumulative links and thresholds. The source response has $K_s$ ordinal categories and the target response has $K_t$ categories, where $K_s$ and $K_t$ can differ. (In particular, the source and target categories can be unaligned even when $K_s=K_t$.) The framework accommodates two forms of cross-domain heterogeneity: the high-dimensional regression coefficients can be different, and the response mechanisms can differ through cumulative links, class counts and thresholds. We assume that the high-dimensional regression coefficients are close in $\ell_1$-distance and develop a two-step transfer-learning framework that pools source and target likelihoods and then corrects the transfer bias. We propose two algorithms that differ in their use of target data. The full-target algorithm reuses all target observations in both stages, while the sample-split algorithm uses separate target subsamples for pooling and correction.

This formulation introduces new statistical challenges arising from the heterogeneity of the ordinal response spaces. In addition to the discrepancy between the regression coefficients, the source and target models may have different ordinal threshold structures and numbers of categories. Thus, transfer learning must account for both coefficient heterogeneity and ordinal-threshold estimation. To characterize these effects, we derive $\ell_1$- and $\ell_2$-error bounds and corresponding minimax lower bounds. The resulting rates separate the costs of high-dimensional estimation, source-target transfer difficulty, and ordinal-threshold estimation. Under suitable conditions, both algorithms achieve minimax-optimal expected rates for parameter estimation and excess risk.

Our work therefore complements the recent literature on heterogeneous and discrete-response transfer learning. Existing studies consider transferring information across heterogeneous datasets or high-dimensional discrete-response models; our focus is instead on unaligned ordinal response spaces, where the source and target classification systems themselves differ.

The main contributions of this paper are summarized as follows:\\
\textbf{Methodological Innovation}: We introduce a high-dimensional transfer-learning framework for unaligned ordinal response spaces, where the source and target may have different numbers of ordinal categories and different category definitions. Our framework accommodates domain-specific regression coefficients, cumulative links and thresholds, and provides the full-target and sample-split algorithms based on penalized pooling and target correction.\\
\textbf{Theoretical Guarantees}: We derive $\ell_1$- and $\ell_2$-error bounds and excess-risk bounds for both algorithms, together with corresponding minimax lower bounds. The rates separate the costs of high-dimensional estimation, source--target discrepancy and target-threshold estimation. Under suitable conditions, both algorithms achieve minimax-optimal expected estimation and excess-risk rates.\\
\textbf{Empirical Validation}: We validate the practical utility of our framework through numerical simulations and a medical data application. The simulations favor the full-target algorithm over the sample-split algorithm, and the real-data results demonstrate its predictive gains over the pooled and target-only algorithms.

The rest of the article is organized as follows. Section \ref{sec:Problem-Formulation} details the problem formulation, including the heterogeneous latent variable model, data assumptions, and the sample-split algorithm based on penalized maximum likelihood. Section \ref{sec:Results} presents the regularity assumptions, estimation and excess-risk bounds, minimax lower bounds and the full-target algorithm. Section \ref{sec:Simulation} provides numerical simulations to evaluate finite-sample estimation and prediction performance. Section \ref{sec:Real-Data} studies hospital-mortality prediction, emphasizing the effect of additional source observations at fixed target sample sizes. Section \ref{sec:Discussion} concludes the paper and discusses directions for future research. The \hyperlink{supplement:start}{Supplementary Material} contains additional analysis for real data and all proofs.

\section{Problem Formulation}\label{sec:Problem-Formulation}

\subsection{Preliminaries}\label{subsec:preliminaries}
For an integer $K\ge2$, write $[K]=\{1,\ldots,K\}$ and $[0]=\varnothing$. Let $\Delta^{K-1}=\{u\in[0,1]^K:\sum_ku_k=1\}$ be the probability simplex. For $v\in\R^r$, let $\|v\|_q=(\sum_{j=1}^r|v_j|^q)^{1/q}$ for $1\le q<\infty$, $\|v\|_\infty=\max_j|v_j|$, $\|v\|_0=\#\{j:v_j\ne0\}$, and $\supp(v)=\{j:v_j\ne0\}$. For an index set $S\subseteq[r]$, $v_S$ is the subvector indexed by $S$ and $S^c=[r]\setminus S$. For a vector with an existing subscript, we append the index set after a comma: $v_{\beta,S}=(v_\beta)_S$. Write $0_r$ for the zero vector, $I_r$ for the identity matrix, and $e_j$ for the $j$th coordinate unit vector in the relevant Euclidean space. The inner product is $\langle u,v\rangle=u^\top v$.

For a matrix $M=(M_{ij})$, let $\|M\|_1=\max_j\sum_i|M_{ij}|$ and $\|M\|_{\rm op}=\sup_{\|w\|_2=1}\|Mw\|_2$ denote its matrix $\ell_1$ norm and spectral norm, respectively. For symmetric matrices, $A\preceq B$ means that $B-A$ is positive semidefinite. For nonnegative quantities, $u\lesssim v$ means $u\le Cv$ for a fixed constant, and $u\asymp v$ means both $u\lesssim v$ and $v\lesssim u$. Write $a\wedge b=\min(a,b)$.

For distributions $P$ and $Q$ with densities $f$ and $g$ relative to a common measure $\nu$, define the Kullback--Leibler (KL) divergence by
\[
 \KL(P,Q)=\int f\log(f/g)\,d\nu.
\]

A real random variable $Z$ is sub-Gaussian with variance proxy $\sigma^2$ when
\[
 \E\exp\{u(Z-\E Z)\}\le\exp(u^2\sigma^2/2),\qquad u\in\R.
\]
A vector $X\in\R^p$ is sub-Gaussian with variance proxy $\sigma_x^2$ when every projection $v^\top(X-\E X)$ has variance proxy $\sigma_x^2\|v\|_2^2$:
\[
 \E\exp\{u v^\top(X-\E X)\}\le\exp(u^2\sigma_x^2\|v\|_2^2/2),\qquad u\in\R,\ v\in\R^p.
\]
The Orlicz norms are $\|Z\|_{\psi_r}=\inf\{c>0:\E\exp(|Z|^r/c^r)\le2\}$ for $r=1,2$. They describe sub-exponential and sub-Gaussian tails, respectively. The projection bound implies $\|v^\top(X-\E X)\|_{\psi_2}\le C\sigma_x\|v\|_2$ for a universal constant $C$. Throughout, $c,C$ denote positive constants whose values may change from line to line. Their dependence on the model is specified with the results, and the bounds hold uniformly over the stated parameter classes.

\subsection{Heterogeneous Cumulative-Link Models}\label{subsec:model}
Let $d\in\{s,t\}$ index the source ($s$) and target ($t$) domains. Throughout the paper, a subscript $d$ or a superscript $d$ or $(d)$ identifies the corresponding domain. For each $d$, let $K_d\ge2$ be the fixed number of ordered response categories, $X^{(d)}\in\R^p$ the covariate vector, and $Y^d\in[K_d]$ the ordinal response. Let $\beta_d\in\R^p$ be the regression coefficient vector and $\tau^d=(\tau^d_1,\ldots,\tau^d_{K_d-1})^\top$ the unknown finite threshold vector, with $-\infty=\tau^d_0<\tau^d_1<\cdots<\tau^d_{K_d-1}<\tau^d_{K_d}=\infty$. The covariate dimension $p\ge 2$. 

The latent error $\epsilon_d$ is independent of $X^{(d)}$ and has a known density $f_d$ that is strictly positive on $\R$. Its distribution function is $F_d(z)=\int_{-\infty}^z f_d(u)\,du$. The error distributions may differ between the two domains. A latent variable $V^d$ determines the response through
\begin{equation}\label{eq:latent}
 V^d=X^{(d)\top}\beta_d+\epsilon_d,\qquad Y^d=k\ \Longleftrightarrow\ \tau^d_{k-1}<V^d\le\tau^d_k.
\end{equation}
For $x\in\R^p$ and $k\in[K_d]$, the conditional probabilities are
\begin{equation}\label{eq:model}
 \pi_{d,k}(x;\beta_d,\tau^d)=\Pp(Y^d=k\mid X^{(d)}=x)=F_d(\tau^d_k-x^\top\beta_d)-F_d(\tau^d_{k-1}-x^\top\beta_d).
\end{equation}
Write $\pi_d(x;\beta_d,\tau^d)=(\pi_{d,1}(x;\beta_d,\tau^d),\ldots,\pi_{d,K_d}(x;\beta_d,\tau^d))^\top\in\Delta^{K_d-1}$ for the conditional probability vector.
In each domain, this is a cumulative-link ordinal regression model \citep{mccullagh1980,tutz2022ordinal}, since
\[
 F_d^{-1}\{\Pp(Y^d\le j\mid X^{(d)}=x)\}=\tau_j^d-x^\top\beta_d,\qquad j\in[K_d-1].
\]
The link is $F_d^{-1}$, and the coefficient vector is common to all cutpoints within a domain.
For any $K_d\ge2$, two standard choices are the logistic and normal distribution functions,
\[
 F_{\rm L}(z)=(1+e^{-z})^{-1},\qquad F_{\rm N}(z)=\int_{-\infty}^z(2\pi)^{-1/2}e^{-u^2/2}\,du.
\]
We emphasize that the two domains may use different cumulative links.

Write $(\beta_d^*,\tau^{d*})$ for the true parameters. We impose source sparsity and an $\ell_1$ discrepancy radius:
\begin{equation}\label{eq:shift}
 \|\beta_s^*\|_0\le s_1,\qquad \delta^*=\beta_t^*-\beta_s^*,\qquad \|\delta^*\|_1\le h.
\end{equation}
Here $s_1\in[1,p]$ is an integer and $h\ge0$ quantifies the total magnitude of the coefficient shift. Transfer exploits the similarity between the coefficient vectors, while domain-specific links and thresholds accommodate the different response mechanisms.

\subsection{Estimation Procedure}\label{subsec:algorithm}
We estimate the regression coefficients and thresholds by penalized maximum likelihood, first pooling source and target data and then estimating a target correction.

We observe source data $\{(X_i^{(s)},Y_i^s)\}_{i=1}^{n_s}$ and target data $\{(X_i^{(t)},Y_i^t)\}_{i=1}^{n_t}$, with $n_s\ge1$ and $n_t\ge2$. Each pair $(X_i^{(d)},Y_i^d)$ is a copy of $(X^{(d)},Y^d)$, and all observations are mutually independent. The distributions of $X^{(s)}$ and $X^{(t)}$ may differ. Put $n=n_s+n_t$ for the total sample size. For $x\in\R^p$, define the single-observation negative log-likelihoods by
\[
 l_s(\beta,\tau^s;x,y)=-\log\{F_s(\tau^s_y-x^\top\beta)-F_s(\tau^s_{y-1}-x^\top\beta)\},\qquad y\in[K_s],
\]
\[
 l_t(\beta,\tau^t;x,y)=-\log\{F_t(\tau^t_y-x^\top\beta)-F_t(\tau^t_{y-1}-x^\top\beta)\},\qquad y\in[K_t].
\]
Thus $l_d(\beta,\tau^d;x,y)=-\log\pi_{d,y}(x;\beta,\tau^d)$ for $d=s,t$. The source and target empirical losses are
\begin{equation}\label{eq:loss_source}
 L_s(\beta,\tau^s)=n_s^{-1}\sum_{i=1}^{n_s}l_s(\beta,\tau^s;X_i^{(s)},Y_i^s).
\end{equation}
\begin{equation}\label{eq:loss_target}
 L_t(\beta,\tau^t)=n_t^{-1}\sum_{i=1}^{n_t}l_t(\beta,\tau^t;X_i^{(t)},Y_i^t).
\end{equation}
Each loss is normalized by its domain sample size.

\Needspace{5\baselineskip}
Define the augmented pooled and target parameters by
\begin{equation}\label{eq:augmented}
 \begin{aligned}\phi&=(\beta^\top,(\tau^s)^\top,(\tau^t)^\top)^\top\in\R^{p+K_s+K_t-2},\\
 \psi&=(\beta^\top,(\tau^t)^\top)^\top\in\R^{p+K_t-1}.\end{aligned}
\end{equation}
The true target parameter is $\psi^*=(\beta_t^{*\top},(\tau^{t*})^\top)^\top$.

For arbitrary positive numbers $b,t,\gamma$ and an integer $K\ge2$, define the set families
\begin{equation}\label{eq:compact}
 \begin{aligned}\mathcal B(b)&=\{\beta\in\R^p:\|\beta\|_2\le b\},\\
 \mathcal T_K(t,\gamma)&=\{\tau\in\R^{K-1}:\|\tau\|_\infty\le t,\ \tau_{j+1}-\tau_j\ge\gamma,\ j\in[K-2]\}.\end{aligned}
\end{equation}
For $K=2$, there is a single finite threshold and the gap constraint is empty. 

To specify the parameter spaces, we fix positive numbers $B,T,g$ (independent of $n,p$) throughout the paper. We assume $2T>(K_d-2)g$ for $d=s,t$, which is equivalent to requiring that $\mathcal T_{K_s}(T,g)$ and $\mathcal T_{K_t}(T,g)$ have nonempty interior. The compact convex parameter spaces are
\[
 \Phi=\mathcal B(B)\times\mathcal T_{K_s}(T,g)\times\mathcal T_{K_t}(T,g),\qquad \Psi=\mathcal B(B)\times\mathcal T_{K_t}(T,g).
\]
\Needspace{3\baselineskip}
Partition the target indices into fixed disjoint sets $I_1,I_2$, independently of the observations, with sizes $n_{t,1}=\lfloor n_t/2\rfloor$ and $n_{t,2}=n_t-n_{t,1}$. The target loss on each part is
\begin{equation}\label{eq:targetSplitLoss}
 L_{t,k}(\beta,\tau^t)=n_{t,k}^{-1}\sum_{i\in I_k}l_t(\beta,\tau^t;X_i^{(t)},Y_i^t),\qquad k=1,2.
\end{equation}
For the augmented target parameter $\psi$, write $l_t(\psi;x,y)=l_t(\beta,\tau^t;x,y)$, $L_t(\psi)=L_t(\beta,\tau^t)$ and $L_{t,k}(\psi)=L_{t,k}(\beta,\tau^t)$. Define the pooling sample size and domain weights by
\begin{equation}\label{eq:poolSampleSizes}
 n_{\rm p}=n_s+n_{t,1},\qquad \rho_s=n_s/n_{\rm p},\qquad \rho_t=n_{t,1}/n_{\rm p}.
\end{equation}
The pooled empirical and population losses are
\begin{equation}\label{eq:poolrisk}
 L_{\rm pool}(\phi)=\rho_sL_s(\beta,\tau^s)+\rho_tL_{t,1}(\beta,\tau^t),\qquad R_{\rm pool}(\phi)=\E L_{\rm pool}(\phi).
\end{equation}
For $d\in\{s,t\}$, write $R_d(\beta,\tau^d)=\E L_d(\beta,\tau^d)$ for the domain population loss; in the target domain, $R_t(\psi)=\E L_{t,k}(\psi)$ for $k=1,2$. For $n_t\ge2$, $n_{\rm p}\asymp n$ and $n_{t,k}\asymp n_t$.

Motivated by penalized $M$-estimation \citep{negahban2012} and two-step transfer procedures \citep{li2022linear,bastani2021,tian2023}, we propose the sample-split algorithm, which pools the source and first target part, then estimates the target correction on the second part.

\begin{algorithm}[htbp]
\caption{Sample-Split Algorithm}\label{alg:transfer_learning}
\small
\textbf{Require:} Source sample; target parts $I_1,I_2$; optimization sets $\Phi,\Psi$; penalties $\lambda_{\rm pool},\lambda_t>0$.
\begin{enumerate}[label=\arabic*:,leftmargin=1.8em,itemsep=0pt]
\item \textbf{Pooling step.} Using all source observations and the target observations in $I_1$, obtain $\hat\phi=(\hat\beta_{\rm pool}^\top,\hat\tau^{s\top},\hat\tau^{t,\rm pool\top})^\top$ from
\[
 \hat\phi\in\arg\min_{\phi\in\Phi}\{L_{\rm pool}(\phi)+\lambda_{\rm pool}\|\beta\|_1\}.
\]
\item \textbf{Target correction step.} Using the target observations in $I_2$, obtain $\hat\psi=(\hat\beta_t^\top,\hat\tau^{t\top})^\top$ from
\[
 \hat\psi\in\arg\min_{\psi\in\Psi}\{L_{t,2}(\psi)+\lambda_t\|\beta-\hat\beta_{\rm pool}\|_1\}.
\]
\end{enumerate}
\textbf{Ensure:} The target estimator $\hat\psi$.
\end{algorithm}
The thresholds are unpenalized in both steps. Compactness and continuity guarantee the existence of minimizers. The second penalty controls the coefficient correction relative to $\hat\beta_{\rm pool}$. The target thresholds are re-estimated during correction. The correction observations in $I_2$ are independent of the pooling observations, which facilitates the theoretical analysis. 

\section{Main Results}\label{sec:Results}
\subsection{Assumptions and Parameter Space}\label{subsec:assumptions}
We impose the following regularity conditions on the fixed model and parameter spaces.

Recall that estimation is performed over the compact convex parameter spaces $\Phi$ and $\Psi$ defined in Section \ref{subsec:algorithm}, with fixed positive numbers $B,T,g$ independent of $n,p$ and satisfying $2T>(K_d-2)g$ for $d=s,t$. For the theoretical analysis, fix $r_0\in(0,\min\{B,T\})$, independently of $n,p$. The assumptions below require the true coefficients and thresholds to satisfy the constraints defining these spaces with a positive margin $r_0$.

\begin{description}[style=nextline,leftmargin=1cm]
\item[\textbf{C1.}] \hypertarget{condition:C1}{} For $d=s,t$, $X^{(d)}$ has mean $\mu_d=\E X^{(d)}\in\R^p$ with $\|\mu_d\|_2\le M_\mu$ and sub-Gaussian variance proxy $\sigma_x^2$ in the centered-projection sense of Section \ref{subsec:preliminaries}. Its covariance $\Sigma_d=\E[(X^{(d)}-\mu_d)(X^{(d)}-\mu_d)^\top]$ satisfies $c_\Sigma I_p\preceq\Sigma_d\preceq C_\Sigma I_p$. The constants $0\le M_\mu<\infty$, $\sigma_x,c_\Sigma,C_\Sigma>0$ are fixed with respect to $n,p$; the domain means may differ.
\item[\textbf{C2.}] \hypertarget{condition:C2}{} Equation \eqref{eq:shift} holds, and $\|\beta_s^*\|_2,\|\beta_t^*\|_2\le B-r_0$.
\item[\textbf{C3.}] \hypertarget{condition:C3}{} The class counts are fixed and $\tau^{d*}\in\mathcal T_{K_d}(T-r_0,g+r_0)$ for $d=s,t$. These true-parameter threshold sets have nonempty interior.
\item[\textbf{C4.}] \hypertarget{condition:C4}{} Each domain follows \eqref{eq:model} with its known distribution function $F_d$ and strictly positive density $f_d$, and $\epsilon_d$ is independent of $X^{(d)}$. The negative log-density $V_d=-\log f_d$ satisfies
\begin{equation}\label{eq:linkCondition}
 V_d\in C^2(\R),\qquad 0<V_d''(z)\le M,\qquad z\in\R,
\end{equation}
where $M<\infty$ is a fixed positive constant common to the domains. The error distributions and $M$ do not vary with $n$ or $p$.
\item[\textbf{C5.}] \hypertarget{condition:C5}{} Population pooling stability. Let $\bar\phi$ denote the unique population pooled minimizer using the weights in \eqref{eq:poolSampleSizes}, with coefficient block $\bar\beta$ and threshold blocks $\bar\tau^s,\bar\tau^t$. We assume that $\bar\phi$ is interior to $\Phi$ and satisfies the following bound with a fixed $C_{\rm pool}\ge1$, uniform over the parameter class and independent of $n,p,s_1,h$:
\begin{equation}\label{eq:bias}
 \bar\phi=\arg\min_{\phi\in\Phi}R_{\rm pool}(\phi),\qquad \|\bar\beta-\beta_t^*\|_1\le C_{\rm pool}h.
\end{equation}
\end{description}

\begin{remark}[Interpretation of C4]\label{rem:C4}
Condition C4 gives smooth, strictly log-concave error densities with bounded curvature of their negative logarithms. Continuity and pointwise positivity of $V_d''$ supply a positive lower bound on each fixed compact interval. Write $V_{\rm N}$ and $V_{\rm L}$ for the negative log-densities of the standard normal and logistic distributions, respectively. Since
\[
 V_{\rm N}''(z)=1,\qquad V_{\rm L}''(z)=\frac{2e^z}{(1+e^z)^2},
\]
the standard normal and logistic links satisfy C4.

A further family satisfying C4 is
\begin{equation}\label{eq:linkFamily}
 f_{\lambda,\gamma}(z)=Z_{\lambda,\gamma}^{-1}e^{-\lambda z^2/2}\left(\frac{e^{z/2}+e^{-z/2}}{2}\right)^{-2\gamma},\qquad \lambda,\gamma\ge0,\quad \lambda+\gamma>0,
\end{equation}
where $Z_{\lambda,\gamma}=\int_{\R}e^{-\lambda u^2/2}\{(e^{u/2}+e^{-u/2})/2\}^{-2\gamma}\,du$ is the finite normalization constant and $(\lambda,\gamma)$ is known and fixed. Its negative log-density $V_{\lambda,\gamma}=-\log f_{\lambda,\gamma}$ satisfies $V_{\lambda,\gamma}''(z)=\lambda+2\gamma e^z/(1+e^z)^2$. The choices $(\lambda,\gamma)=(1,0)$ and $(0,1)$ give the standard normal and logistic densities.

Under C1--C4, the likelihood scores have uniform sub-exponential coefficient tails and sub-Gaussian threshold tails; see Lemma~\ref{lem:linkRegularity} in Supplement~\ref{app:pool}. 
\end{remark}

\begin{remark}[Interpretation of C5]\label{rem1}
C5 requires population pooling to preserve the scale of the source--target coefficient discrepancy, with a bias constant uniform in dimension. It is a stability property of the source and target distributions and the pooling weights. Its interiority requirement gives the population score equations used in the pooling concentration argument. Its coefficient bound controls both the correction $\beta_t^*-\bar\beta$ and the approximate-sparsity tail: for $S=\supp(\beta_s^*)$,
\[
 \|\bar\beta_{S^c}\|_1\le\|\bar\beta-\beta_s^*\|_1\le(C_{\rm pool}+1)h.
\]
Thus the transfer radius remains an $\ell_1$ radius after population pooling.

A related population stability condition appears in Assumption 4 of \citet{tian2023}, which bounds induced $\ell_1$ norms of relative integrated information matrices to control pooled coefficient bias. For the ordinal model, Proposition~\ref{prop:profilebias} in Supplement~\ref{app:extensions} gives a corresponding sufficient condition for C5 after minimizing over the unknown thresholds.

Under C1--C4, C5 holds with $C_{\rm pool}=1$ when $\beta_s^*=\beta_t^*$. Each domain risk is minimized at its true parameters, so $\bar\phi=(\beta_s^{*\top},\tau^{s*\top},\tau^{t*\top})^\top$ is interior and has zero coefficient bias. This case allows different covariate distributions, links, class counts and threshold vectors.

A concrete case with nonzero coefficient discrepancy uses $X^{(d)}\sim N(\mu_d,\sigma_d^2I_p)$, $d=s,t$, with positive variances and means satisfying C1 and collinear true coefficient vectors. If their Euclidean discrepancy is within a sufficiently small fixed radius, the pooled coefficient lies on the segment joining the two true coefficients and C5 holds with $C_{\rm pool}=1$. Example~\ref{ex:C5Gaussian} in Supplement~\ref{app:extensions} gives the precise statement and proof. The means, spherical variances, links, class counts and threshold vectors may differ between domains.
\end{remark}

Define the full domain parameter $\theta=(\beta_s^\top,\tau^{s\top},\beta_t^\top,\tau^{t\top})^\top\in\R^{2p+K_s+K_t-2}$ and its target component $\psi(\theta)=(\beta_t^\top,\tau^{t\top})^\top\in\R^{p+K_t-1}$. The parameter space used for the upper and lower bounds is
\begin{equation}\label{eq:parameterSpace}
\Theta(s_1,h)=\left\{\theta:
\begin{aligned}
&\|\beta_s\|_0\le s_1,\ \|\beta_t-\beta_s\|_1\le h,\\
&\beta_s,\beta_t\in\mathcal B(B-r_0),\\
&\tau^d\in\mathcal T_{K_d}(T-r_0,g+r_0),\ d=s,t,\\
&\text{C5 holds}
\end{aligned}
\right\}.
\end{equation}
All the theoretical guarantees and minimax risks discussed below are evaluated uniformly over the parameter space $\Theta(s_1,h)$.

\Needspace{7\baselineskip}
\subsection{Convergence Rates for Estimation}\label{subsec:theorems}
We analyze Algorithm \ref{alg:transfer_learning} with the penalty and sample-size conditions below. Theorems \ref{thm2} and \ref{thm3} give upper bounds and minimax lower bounds for estimation, respectively.

\Needspace{6\baselineskip}
Fix a failure probability $\varepsilon\in(0,1/4)$. For a sufficiently large positive constant $A_\lambda$, choose the penalties in Algorithm \ref{alg:transfer_learning} as
\begin{equation}\label{eq:tuning}
 \lambda_{\rm pool}=A_\lambda\sqrt{\frac{\log(p/\varepsilon)}{n_{\rm p}}},\qquad \lambda_t=A_\lambda\sqrt{\frac{\log(p/\varepsilon)}{n_{t,2}}}.
\end{equation}
Assume that the sample sizes, sparsity level $s_1$ and coefficient discrepancy $h$ satisfy
\begin{equation}\label{eq:rateConditions}
 \min(n_s,n_{t,1},n_{t,2})\ge C_n\log(p/\varepsilon),\qquad \frac{s_1\log(p/\varepsilon)}{n}\le c,\qquad h\sqrt{\frac{\log(p/\varepsilon)}{n_t}}\le c,
\end{equation}
where $C_n>0$ is sufficiently large and $c>0$ is sufficiently small. The constants $A_\lambda,C_n,c,C$ are independent of $n_s,n_t,p,s_1,h,\varepsilon$. Since $n_{\rm p}\asymp n$ and $n_{t,k}\asymp n_t$, we express the rates in terms of $n$ and $n_t$.

\begin{lemma}[Error Bound of the Pooling Step]\label{thm1}
Under C1--C5 and \eqref{eq:rateConditions}, with probability at least $1-\varepsilon$,
\begin{align}\label{eq:poolrates}
 \|\hat\beta_{\rm pool}-\bar\beta\|_2^2+\sum_{d=s,t}\rho_d\|\hat\tau^{d,\rm pool}-\bar\tau^d\|_2^2&\le C\left\{\frac{s_1\log(p/\varepsilon)}{n}+h\sqrt{\frac{\log(p/\varepsilon)}{n}}\right\},\\
 \|\hat\beta_{\rm pool}-\bar\beta\|_1&\le C\left\{s_1\sqrt{\frac{\log(p/\varepsilon)}{n}}+h\right\}.\nonumber
\end{align}
Consequently $\|\hat\beta_{\rm pool}-\beta_t^*\|_2\le C\{\sqrt{s_1\log(p/\varepsilon)/n}+h\}$ and $\|\hat\beta_{\rm pool}-\beta_t^*\|_1\le C\{s_1\sqrt{\log(p/\varepsilon)/n}+h\}$. For $d=s,t$,
\begin{equation}\label{eq:pooltau}
 \|\hat\tau^{d,\rm pool}-\bar\tau^d\|_2\le \frac{C}{\sqrt{\rho_d}}\left\{\frac{s_1\log(p/\varepsilon)}{n}+h\sqrt{\frac{\log(p/\varepsilon)}{n}}\right\}^{1/2},
\end{equation}
where $\hat\tau^{s,\rm pool}=\hat\tau^s$.
\end{lemma}
\begin{proof}See Supplement~\ref{app:pool}.\end{proof}

\begin{theorem}[Target Estimation Bounds]\label{thm2}
Under C1--C5 and \eqref{eq:rateConditions}, Algorithm \ref{alg:transfer_learning} with the penalties in \eqref{eq:tuning} satisfies, with probability at least $1-2\varepsilon$,
\begin{align}\label{eq:targetrates}
 \|\hat\psi-\psi^*\|_2&\le C\left\{\sqrt{\frac{s_1\log(p/\varepsilon)}{n}}+\left(h^{1/2}\left\{\frac{\log(p/\varepsilon)}{n_t}\right\}^{1/4}\wedge h\right)+\sqrt{\frac{(K_t-1)\log(1/\varepsilon)}{n_t}}\right\},\\
 \|\hat\psi-\psi^*\|_1&\le C\left\{s_1\sqrt{\frac{\log(p/\varepsilon)}{n}}+h+(K_t-1)\sqrt{\frac{\log(1/\varepsilon)}{n_t}}\right\}.\nonumber
\end{align}
For the expected bounds, take $\varepsilon=\{p(n_t+1)\}^{-4}$ in \eqref{eq:tuning} and assume \eqref{eq:rateConditions} with this choice. Then
\begin{align}\label{eq:expectedRates}
 \E\|\hat\psi-\psi^*\|_2&\le C\left\{\sqrt{\frac{s_1\log\{p(n_t+1)\}}{n}}+\left(h^{1/2}\left\{\frac{\log\{p(n_t+1)\}}{n_t}\right\}^{1/4}\wedge h\right)+\sqrt{\frac{K_t-1}{n_t}}\right\},\\
 \E\|\hat\psi-\psi^*\|_1&\le C\left\{s_1\sqrt{\frac{\log\{p(n_t+1)\}}{n}}+h+\frac{K_t-1}{\sqrt{n_t}}\right\}.\nonumber
\end{align}
\end{theorem}
\begin{proof}See Supplement~\ref{app:target}.\end{proof}

The $\ell_2$ bound in \eqref{eq:expectedRates} implies $\E\|\hat\psi-\psi^*\|_2\to0$ if
\[
 \frac{s_1\log\{p(n_t+1)\}}{n}\to0,\qquad \left(h\sqrt{\frac{\log\{p(n_t+1)\}}{n_t}}\wedge h^2\right)\to0,\qquad n_t\to\infty.
\]
The $\ell_1$ bound in \eqref{eq:expectedRates} implies $\E\|\hat\psi-\psi^*\|_1\to0$ if
\[
 \frac{s_1^2\log\{p(n_t+1)\}}{n}\to0,\qquad h\to0,\qquad n_t\to\infty.
\]
In each case, convergence in probability follows from Markov's inequality.

\Needspace{8\baselineskip}
In minimax statements, $\E_\theta$ denotes expectation under the full parameter $\theta$, and the infimum ranges over all measurable estimators based on the source and target samples. Fix $\alpha\in(0,1)$. For the lower bounds, suppose
\begin{equation}\label{eq:lowerRegime}
 s_1\le cp^\alpha,\quad h\sqrt{\frac{n_t}{\log p}}\le cp^\alpha,\quad \frac{s_1\log p}{n}\le c(B-r_0)^2,\quad \left(h\sqrt{\frac{\log p}{n_t}}\wedge h^2\right)\le c(B-r_0)^2.
\end{equation}
Here $c>0$ is sufficiently small and independent of $n_s,n_t,p,s_1,h$. The constants in the lower bounds are also independent of these quantities.

\Needspace{11\baselineskip}
\begin{theorem}[Minimax Estimation Lower Bounds]\label{thm3}
For each fixed pair of distributions $X^{(d)}\sim N(\mu_d,I_p)$, $d=s,t$, with $\|\mu_d\|_2\le M_\mu$, and under \eqref{eq:lowerRegime}, the minimax errors over $\Theta(s_1,h)$ satisfy
\begin{align}\label{eq:lowerRates}
 \inf_{\hat\psi}\sup_{\theta\in\Theta}\E_\theta\|\hat\psi-\psi(\theta)\|_2
 &\ge c\left\{\sqrt{\frac{s_1\log p}{n}}+\left(h^{1/2}\left\{\frac{\log p}{n_t}\right\}^{1/4}\wedge h\right)+\sqrt{\frac{K_t-1}{n_t}}\right\},\\
 \inf_{\hat\psi}\sup_{\theta\in\Theta}\E_\theta\|\hat\psi-\psi(\theta)\|_1
 &\ge c\left\{s_1\sqrt{\frac{\log p}{n}}+h+\frac{K_t-1}{\sqrt{n_t}}\right\}.\nonumber
\end{align}
\end{theorem}
\begin{proof}See Supplement~\ref{app:lower}.\end{proof}

\begin{remark}[Interpretation of the convergence rates]\label{rem:rateInterpretation}
The bounds in Theorems \ref{thm2}--\ref{thm3} separate three sources of statistical error. The first terms describe estimation of the shared high-dimensional coefficient structure using the pooled sample. The middle terms quantify the difficulty caused by the source--target coefficient discrepancy $h$. The terms involving $K_t-1$ arise from estimating the target thresholds. Note that the source-threshold contribution is absorbed into the pooling term.

Under the conditions of both Theorems \ref{thm2} and \ref{thm3}, with Gaussian covariates and $\log n_t\lesssim\log p$, we have $\log\{p(n_t+1)\}\asymp\log p$, so the expected $\ell_2$ and $\ell_1$ rates match the minimax lower bounds.
\end{remark}

\begin{remark}[Comparison with GLM and linear transfer rates]\label{rem:rateComparison}
Under their respective assumptions, \citet[Theorem 1]{tian2023} give the GLM coefficient $\ell_2$ order $\sqrt{s_1\log p/n}+\{h^{1/2}(\log p/n_t)^{1/4}\wedge h\}$, and the oracle linear Trans-Lasso bound of \citet[Theorem 1]{li2022linear} has squared $\ell_2$ order $s_1\log p/n+\{s_1\log p/n_t\wedge h\sqrt{\log p/n_t}\wedge h^2\}$. Theorem \ref{thm2} has the GLM coefficient orders under $\log n_t\lesssim\log p$, together with the threshold-estimation cost described in Remark \ref{rem:rateInterpretation}. Our parameter class uses a sparse source reference and an $\ell_1$-bounded target shift, while the GLM result assumes target sparsity. Corollary~\ref{cor:targetSparse} in Supplement~\ref{app:extensions} supplies the target-sparse comparison for $h\le s_1\sqrt{\log p/n_t}$ under the stated conditions.
\end{remark}

\subsection{Minimax Optimality for Excess Risk}
Let $P_\psi$ denote the joint distribution of a target observation $(X^{(t)},Y^t)$ under parameter $\psi$, with the distribution of $X^{(t)}$ held fixed. Write $\pi_t(x;\psi)=\pi_t(x;\beta,\tau^t)$ and $\pi_t(x;\theta)=\pi_t(x;\psi(\theta))$. For a new observation from $P_{\psi^*}$, write $X$ for its covariate vector and define the target risk and excess risk using the negative log-likelihood loss:
\begin{equation}\label{eq:populationrisk}
 R_t(\psi)=\E[-\log\pi_{t,Y^t}(X;\psi)],\qquad \mathcal E_t(\psi)=R_t(\psi)-R_t(\psi^*).
\end{equation}
When evaluated at a fitted estimator, the expectation over the new observation is conditional on its training data. Write $\E_X$ for expectation over the new target covariate $X$, independent of the training data.

\Needspace{8\baselineskip}
\begin{theorem}[Excess Risk Bounds]\label{thm4}
Under the conditions for the probability bounds in Theorem \ref{thm2}, with probability at least $1-2\varepsilon$,
\begin{equation}\label{eq:excess_risk_upper}
 \mathcal E_t(\hat\psi)\le C\left\{\frac{s_1\log(p/\varepsilon)}{n}+\left(h\sqrt{\frac{\log(p/\varepsilon)}{n_t}}\wedge h^2\right)+\frac{(K_t-1)\log(1/\varepsilon)}{n_t}\right\}.
\end{equation}

Under the conditions for the expected bounds in Theorem \ref{thm2},
\begin{equation}\label{eq:expectedRisk}
 \E\mathcal E_t(\hat\psi)\le C\left\{\frac{s_1\log\{p(n_t+1)\}}{n}+\left(h\sqrt{\frac{\log\{p(n_t+1)\}}{n_t}}\wedge h^2\right)+\frac{K_t-1}{n_t}\right\}.
\end{equation}

Under the conditions of Theorem \ref{thm3}, the minimax excess risk over data-dependent predictors $\hat q:\R^p\to\Delta^{K_t-1}$ satisfies
\begin{equation}\label{eq:excess_risk_lower}
 \inf_{\hat q}\sup_{\theta\in\Theta}\E_\theta\E_X\KL\{\pi_t(X;\theta),\hat q(X)\}
 \ge c\left\{\frac{s_1\log p}{n}+\left(h\sqrt{\frac{\log p}{n_t}}\wedge h^2\right)+\frac{K_t-1}{n_t}\right\}.
\end{equation}
\end{theorem}
\begin{proof}See Supplement~\ref{app:risk}.\end{proof}

\begin{remark}
Under $\log n_t\lesssim\log p$ and the conditions for both upper and lower bounds, Theorem \ref{thm4} gives minimax-optimal expected excess risk on the Gaussian covariate model class. Its lower bound applies to all data-dependent conditional probability predictors.
\end{remark}

\subsection{Full-Target Algorithm}\label{subsec:fullTarget}
A natural variant of the sample-split algorithm (Algorithm \ref{alg:transfer_learning}) is to use all target observations in both pooling and correction. Reuse of target data in two-stage transfer also appears in the GLM procedure of \citet{tian2023}. With the full-sample losses from Section \ref{subsec:algorithm}, define
\[
 L_F(\phi)=\frac{n_s}{n}L_s(\beta,\tau^s)+\frac{n_t}{n}L_t(\beta,\tau^t),\qquad R_F(\phi)=\E L_F(\phi).
\]
We then propose the full-target algorithm below.

\begin{algorithm}[H]
\caption{Full-Target Algorithm}\label{alg:fullTarget}
\small
\textbf{Require:} Source sample; full target sample; optimization sets $\Phi,\Psi$; penalties $\lambda_{p,F},\lambda_{t,F}>0$.
\begin{enumerate}[label=\arabic*:,leftmargin=1.8em,itemsep=0pt]
\item \textbf{Pooling step.} Using all source and target observations, obtain
\[
 \widehat\phi_F\in\arg\min_{\phi\in\Phi}\{L_F(\phi)+\lambda_{p,F}\|\beta\|_1\}.
\]
Let $\widehat\beta_{p,F}\in\R^p$ denote its coefficient block.
\item \textbf{Target correction step.} Using all target observations, obtain
\[
 \widehat\psi_F\in\arg\min_{\psi\in\Psi}\{L_t(\psi)+\lambda_{t,F}\|\beta-\widehat\beta_{p,F}\|_1\}.
\]
\end{enumerate}
\textbf{Ensure:} The target estimator $\widehat\psi_F$.
\end{algorithm}

The following proposition gives parameter-estimation and excess-risk bounds for the full-target algorithm.

\begin{proposition}[Full-target estimation and excess-risk bounds]\label{prop:fullTarget}
Assume C1--C4. Let $\bar\phi_F$ minimize $R_F$ over $\Phi$, and let $\bar\beta_F$ denote its coefficient block. Suppose $\bar\phi_F$ is interior and satisfies $\|\bar\beta_F-\beta_t^*\|_1\le C_{\rm pool}h$, with a fixed $C_{\rm pool}\ge1$, uniform in $n,p,s_1,h$ and the model parameters. This is the population stability condition for the full-target weights $n_s/n,n_t/n$.

For $\varepsilon\in(0,1/4)$ and a sufficiently large positive constant $A_\lambda$, set
\[
\lambda_{p,F}=A_\lambda\sqrt{\frac{\log(p/\varepsilon)}{n}},\qquad \lambda_{t,F}=A_\lambda\sqrt{\frac{\log(p/\varepsilon)}{n_t}}.
\]
If $\min(n_s,n_t)\ge C_n\log(p/\varepsilon)$ and
\[
\frac{s_1\log(p/\varepsilon)}{\sqrt{nn_t}}+h\sqrt{\frac{\log(p/\varepsilon)}{n_t}}\le c,
\]
for sufficiently large $C_n>0$ and sufficiently small $c>0$, then with probability at least $1-2\varepsilon$,
\begin{equation}\label{eq:fullTargetRate}
\|\widehat\psi_F-\psi^*\|_2^2+\mathcal E_t(\widehat\psi_F)\le C\left\{\frac{s_1\log(p/\varepsilon)}{\sqrt{nn_t}}+h\sqrt{\frac{\log(p/\varepsilon)}{n_t}}+\frac{(K_t-1)\log(1/\varepsilon)}{n_t}\right\}.
\end{equation}

If, in addition, $n_s\lesssim n_t$, then the estimation bounds \eqref{eq:targetrates} in Theorem \ref{thm2} and the excess-risk bound \eqref{eq:excess_risk_upper} in Theorem \ref{thm4} also hold with $\hat\psi$ replaced by $\widehat\psi_F$, with probability at least $1-2\varepsilon$.

For the expected bounds under $n_s\lesssim n_t$, take $\varepsilon=\{p(n_t+1)\}^{-4}$ in the penalties above and assume the preceding sample-size and radius conditions with this choice. Then the expected estimation bounds \eqref{eq:expectedRates} and the expected excess-risk bound \eqref{eq:expectedRisk} also hold with $\hat\psi$ replaced by $\widehat\psi_F$.
\end{proposition}
\begin{proof}
See Supplement~\ref{app:extensions}.
\end{proof}

\begin{remark}[Comparison of the two algorithms]
When $n_s/n_t\to\infty$, the coefficient term $s_1\log(p/\varepsilon)/\sqrt{nn_t}$ in \eqref{eq:fullTargetRate} exceeds the corresponding term $s_1\log(p/\varepsilon)/n$ in the sample-split excess-risk bound \eqref{eq:excess_risk_upper} by a factor of $\sqrt{n/n_t}$. This product term arises in our analysis because both steps reuse the target observations, so the correction sample is no longer independent of the pooling estimator.

However, our simulations in Section \ref{sec:Simulation} show that the full-target algorithm has better finite-sample performance than the sample-split algorithm. This advantage is likely due to more efficient use of the limited target data, with every target observation contributing to both pooling and correction. We therefore use the full-target algorithm in the real-data analysis in Section \ref{sec:Real-Data}.
\end{remark}

\section{Numerical Simulation}\label{sec:Simulation}
In this section, we assess our proposed full-target and sample-split algorithms for the two-stage ordinal transfer framework. We also implement the target-only and pooled algorithms as baselines.

\subsection{Data Generation and Evaluation}
We use logistic links in both domains, with $p=1500$, $K_s=5$, $K_t=3$, $\tau^{s*}=(-3,-1,1,3)^\top$ and $\tau^{t*}=(-1,1)^\top$. Covariates are independent $N(0,I_p)$ vectors, and latent errors have the standard logistic distribution. The target coefficient has support $\{1,\ldots,8\}$, with independent magnitudes uniform on $[0.5,1]$ and independent random signs.

Let $v\in\R^p$ have entries $(v_1,v_2,v_9,v_{10})=(0.30,-0.25,0.25,-0.20)$ and all other entries zero. For $h\in\{0,0.5,1,2,4,6\}$, set $\beta_s^*(h)=\beta_t^*+hv$. Then $\|v\|_1=1$ and $\|\beta_s^*(h)-\beta_t^*\|_1=h$. Therefore, the source coefficient $\beta_s^*$ has at most $s_1=10$ nonzero coefficients. 

We use $(n_s,n_t)\in\{(400,200),(1200,600),(2000,1000)\}$. Each setting has $100$ Monte Carlo replications and an independent target test set of size $1000$. Within a replication, all four methods share the training and test observations, and base random numbers are shared across $h$.

For a fitted target parameter $\hat\psi$, estimation performance is measured by $\|\hat\psi-\psi^*\|_2$, including both coefficients and target thresholds. Prediction performance is measured by test excess negative log-likelihood (NLL),
\[
 L_t^{\rm test}(\psi)=\frac{1}{1000}\sum_{i=1}^{1000}l_t(\psi;X_i^{\rm test},Y_i^{t,\rm test}),\qquad
 \widehat{\mathcal E}_{\rm NLL}=L_t^{\rm test}(\hat\psi)-L_t^{\rm test}(\psi^*).
\]

\subsection{Algorithms and Implementations}\label{subsec:methods}
We first list the four methods used in our simulation. 

\begin{itemize}
\item \textbf{Full-target.} Implement Algorithm \ref{alg:fullTarget} with penalties $\lambda_{p,F}=0.6\sqrt{\log p/n}$ and $\lambda_{t,F}=0.4\sqrt{\log p/n_t}$.
\item \textbf{Sample-split.} Implement Algorithm \ref{alg:transfer_learning} with $I_1=\{1,\ldots,\lfloor n_t/2\rfloor\}$ for pooling and its complement $I_2$ for correction. The penalties are $\lambda_{p,S}=0.6\sqrt{\log p/(n_s+n_{t,1})}$ and $\lambda_{t,S}=0.4\sqrt{\log p/n_{t,2}}$.
\item \textbf{Target-only.} Fit the target ordinal model by minimizing $L_t(\psi)+0.4\sqrt{\log p/n_t}\|\beta\|_1$ over $\Psi$.
\item \textbf{Pooled.} Only implement the first step of Algorithm \ref{alg:fullTarget} and use the target component of $\widehat\phi_F$. Use penalty $\lambda_{p,F}=0.6\sqrt{\log p/n}$.
\end{itemize}

All four methods use the same converged proximal-gradient solver. We optimize over $\mathcal B(10)$ for the coefficients $\beta_s,\beta_t$ and $\mathcal T_5(6,0.2),\mathcal T_3(6,0.2)$ for the thresholds $\tau^s,\tau^t$, respectively. It is straightforward to verify that the true parameters lie in the interiors of the corresponding spaces for every $h$ in the simulation grid.

\subsection{Estimation and Predictive Comparisons}
\begin{figure}[htbp]
\centering
\includegraphics[width=0.99\textwidth]{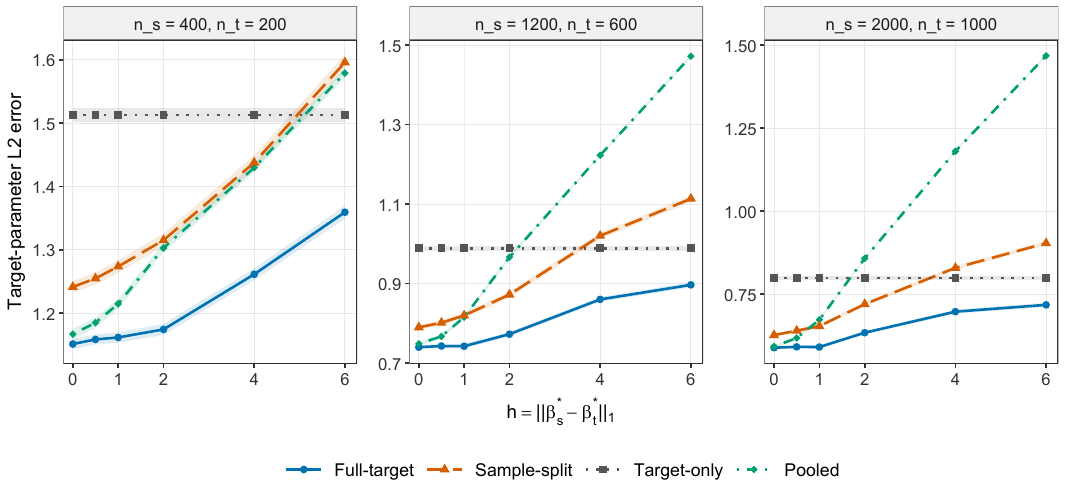}
\caption{Estimation error $\|\hat\psi-\psi^*\|_2$ for the four methods. Panels correspond to the three sample-size settings. Points average 100 replications; shaded bands show one Monte Carlo standard error on either side of the mean.}
\label{fig:l2_h}
\end{figure}

\begin{figure}[htbp]
\centering
\includegraphics[width=0.99\textwidth]{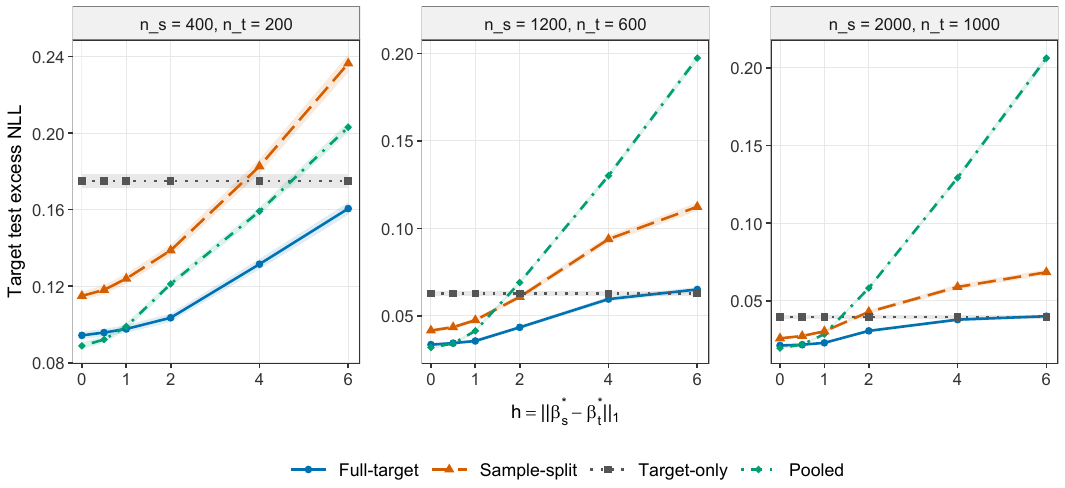}
\caption{Mean target test excess NLL $\widehat{\mathcal E}_{\rm NLL}$ under the same settings as Figure \ref{fig:l2_h}. Each replication uses 1000 independent target test observations. Shaded bands show one Monte Carlo standard error on either side of the mean.}
\label{fig:excess_nll_h}
\end{figure}

Figures \ref{fig:l2_h} and \ref{fig:excess_nll_h} show parameter estimation error and test excess NLL across the discrepancy grid. The full-target algorithm attains lower mean error and lower mean test loss than the sample-split algorithm in all eighteen settings. The difference between the two implementations grows as the source--target discrepancy increases. Larger samples improve both estimators, with the full-target algorithm retaining its advantage throughout the three sample-size settings.

The two-stage correction becomes particularly useful when the domains differ substantially. The pooled algorithm performs well for small $h$, but its error increases with the coefficient discrepancy. The full-target algorithm corrects much of this deterioration and remains close to the target-only algorithm in prediction at the largest discrepancy, while retaining lower parameter error. Using only half of the target observations for correction leaves the sample-split algorithm with higher estimation error and prediction loss than the full-target algorithm.

Table \ref{tab:simulation_comparison} gives direct comparisons at $h=0,2,6$. For example, with $(n_s,n_t)=(1200,600)$ and $h=6$, the full-target algorithm reduces mean Euclidean error from $1.1135$ to $0.8969$ and mean excess NLL from $0.1124$ to $0.0651$ relative to the sample-split algorithm, improvements of approximately $19\%$ and $42\%$, respectively. When the target sample grows from $200$ to $1000$, the Euclidean error of the full-target algorithm at $h=6$ decreases from $1.3593$ to $0.7175$, compared with $1.5959$ to $0.9033$ for the sample-split algorithm.

\begin{table}[htbp]
\centering
\caption{Comparisons between the full-target algorithm and the sample-split algorithm at representative discrepancy levels. Entries are Monte Carlo means, with standard errors in parentheses, over 100 replications.}
\label{tab:simulation_comparison}
\small
\setlength{\tabcolsep}{4pt}
\begin{tabular}{ccrrrr}
\toprule
 & & \multicolumn{2}{c}{Target-parameter Euclidean error} & \multicolumn{2}{c}{Test excess NLL} \\
\cmidrule(lr){3-4}\cmidrule(lr){5-6}
$(n_s,n_t)$ & $h$ & \shortstack{the full-target\\algorithm} & \shortstack{the sample-split\\algorithm} & \shortstack{the full-target\\algorithm} & \shortstack{the sample-split\\algorithm} \\
\midrule
\multirow{3}{*}{$(400,200)$} & 0 & 1.151 (0.008) & 1.241 (0.009) & 0.0943 (0.0019) & 0.1149 (0.0023) \\
 & 2 & 1.174 (0.009) & 1.315 (0.011) & 0.1035 (0.0022) & 0.1387 (0.0028) \\
 & 6 & 1.359 (0.010) & 1.596 (0.012) & 0.1605 (0.0032) & 0.2364 (0.0046) \\
\addlinespace
\multirow{3}{*}{$(1200,600)$} & 0 & 0.740 (0.005) & 0.790 (0.006) & 0.0335 (0.0009) & 0.0416 (0.0010) \\
 & 2 & 0.773 (0.006) & 0.872 (0.008) & 0.0434 (0.0010) & 0.0608 (0.0014) \\
 & 6 & 0.897 (0.007) & 1.113 (0.008) & 0.0651 (0.0014) & 0.1124 (0.0024) \\
\addlinespace
\multirow{3}{*}{$(2000,1000)$} & 0 & 0.589 (0.004) & 0.626 (0.005) & 0.0212 (0.0007) & 0.0258 (0.0007) \\
 & 2 & 0.634 (0.005) & 0.719 (0.006) & 0.0306 (0.0009) & 0.0428 (0.0011) \\
 & 6 & 0.718 (0.006) & 0.903 (0.007) & 0.0400 (0.0011) & 0.0683 (0.0018) \\
\bottomrule
\end{tabular}
\end{table}

These simulations show that the full-target algorithm provides better finite-sample estimation and prediction than the sample-split algorithm. We therefore use the full-target algorithm in the real-data analysis in Section \ref{sec:Real-Data}.

\section{Real Data Analysis}\label{sec:Real-Data}

We apply the heterogeneous ordinal transfer framework to hospital-mortality prediction in the SUPPORT2 study of seriously ill hospitalized adults \citep{knaus1995support,support2_data}. The source task records a five-level functional outcome two months after study entry, whereas the target task concerns the binary outcome of death during hospitalization. These related outcomes provide a clinical example of transfer between unaligned response spaces.

\subsection{Data and Prediction Tasks}
SUPPORT2 contains 9105 patients. We use the 7705 patients with both outcomes recorded, among whom the hospital-mortality rate is $30.20\%$. Data were obtained courtesy of the Vanderbilt University Department of Biostatistics. The source response has five categories: no moderate or severe functional disability, substantial limitations in activities of daily living, a Sickness Impact Profile score of at least 30, coma or intubation, and death before two months. The target response is survival to hospital discharge or hospital death. Both responses are related to disease severity and are measured over different follow-up periods. For example, 862 of the 3123 patients in the worst source category did not die during hospitalization.

Predictors include demographic characteristics, disease categories, comorbidities, vital signs, laboratory measurements and activities of daily living recorded around study day 3. We exclude follow-up outcomes, hospital costs, later treatment-intensity summaries, existing prognostic scores and physicians' survival estimates. Categorical predictors are encoded as indicators, and numerical predictors receive missingness indicators. Eight skewed numerical predictors are transformed by $\log(1+x)$. This construction gives 63 candidate columns. Within each training sample, numerical missing values are imputed by medians, continuous predictors are clipped at their training $0.5\%$ and $99.5\%$ quantiles, and all columns are standardized. 

\subsection{Training and Evaluation}
We apply the full-target, target-only and pooled algorithms described in Section \ref{sec:Simulation}. The penalties are chosen by the cross-validation procedure below. All three algorithms use the same transformed features. The transformation parameters are estimated from source and target training covariates and reapplied to validation and test observations.

We use logistic links with $K_s=5$ and $K_t=2$. We fix $n_t\in\{500,1000\}$ and vary $n_s\in\{500,1000,1500,2000,3000\}$, with 2500 target test patients in each repetition. The source, target-training and test samples consist of disjoint patients.

The penalties retain the sample-size scaling
\[
\lambda_{\rm pool}=C_p\sqrt{\frac{\log p}{n_s+n_t}},\qquad \lambda_t=C_t\sqrt{\frac{\log p}{n_t}}.
\]
We select constants by stratified five-fold cross-validation within the target training sample, minimizing validation NLL. The full-target algorithm jointly selects $(C_p,C_t)$, and the target-only and pooled algorithms select $C_t$ and $C_p$, respectively, over
\[
\begin{split}
C_p&\in\{0.15,0.3,0.6,1.2,2.4,4.8,9.6\},\\
C_t&\in\{0.05,0.1,0.15,0.3,0.6,1.2,2.4,4.8\}.
\end{split}
\]
Each validation fold is excluded from both estimation stages and from estimation of the feature transformation. Models are refitted on the complete training samples after tuning.

For every sample-size setting, we repeat patient sampling, tuning and fitting 60 times using seeds 9401--9460. All three algorithms share the same patients and validation folds within a repetition. At each fixed target sample size, the target-training and test patients are held fixed as the source sample size increases, using nested source samples from the same patient ordering. We evaluate test NLL, AUC and the Brier score. Tables report means with standard deviations across repeated sampling and splitting, and learning curves display mean prediction performance. To examine the consistency of the improvement, we also summarize paired differences
\[
\Delta_{\rm NLL}=\mathrm{NLL}_{\rm baseline}-\mathrm{NLL}_{\rm Full},\qquad \Delta_{\rm AUC}=\mathrm{AUC}_{\rm Full}-\mathrm{AUC}_{\rm baseline},
\]
so that positive values favor the full-target algorithm. These summaries describe variation under repeated sampling and splitting of the observed cohort.

\subsection{Prediction Performance and Stability}
\begin{figure}[!htbp]
\centering
\includegraphics[width=0.99\textwidth]{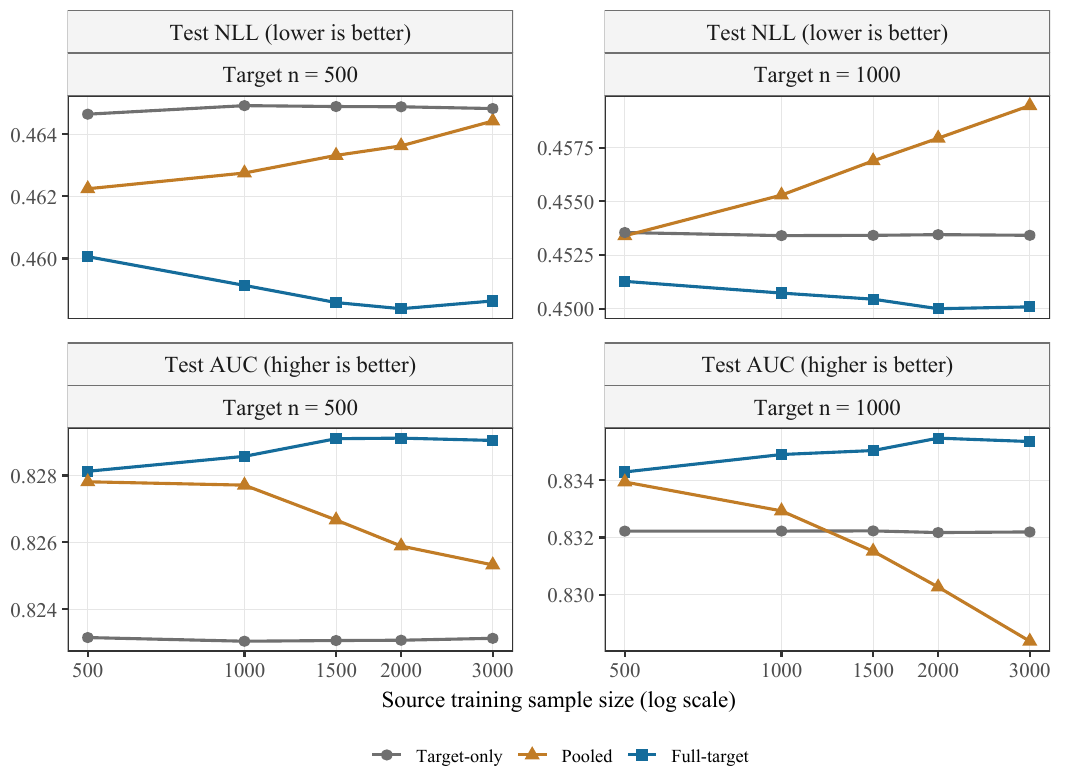}
\caption{SUPPORT2 test NLL (top) and AUC (bottom) as source sample size increases at fixed $n_t=500$ and $1000$. Points are means over 60 sampling and splitting repetitions, with tuning within the target training sample. Each panel has its own vertical scale.}
\label{fig:support_learning}
\end{figure}

Figure \ref{fig:support_learning} shows how additional source observations affect prediction at fixed target sample sizes.

The full-target algorithm has the lowest mean NLL and highest mean AUC at all ten settings. Its performance improves as $n_s$ grows to about 2000 and then levels off. The performance of the pooled algorithm deteriorates as the source sample receives greater weight, emphasizing the discrepancy between the source and target. The target correction adapts the shared information of hospital mortality and preserves the benefit of the additional source observations.

Table \ref{tab:real_data_perf} gives a representative comparison at $(n_s,n_t)=(3000,500)$. The full-target algorithm reduces mean NLL by $1.33\%$ and $1.25\%$ relative to the target-only and pooled algorithms, respectively, and also gives the best mean AUC and Brier score.

\begin{table}[!htbp]
\centering
\caption{SUPPORT2 test performance for $(n_s,n_t)=(3000,500)$ and 2500 test patients. Entries are means with standard deviations in parentheses over 60 sampling and splitting repetitions. Lower NLL and Brier scores and higher AUC are preferred.}
\label{tab:real_data_perf}
\small
\begin{tabular}{lccc}
\toprule
Method & NLL & AUC & Brier score \\
\midrule
the target-only algorithm & 0.4648 (0.0132) & 0.8231 (0.0108) & 0.1524 (0.0051) \\
the pooled algorithm & 0.4644 (0.0101) & 0.8253 (0.0078) & 0.1521 (0.0041) \\
the full-target algorithm & \textbf{0.4586 (0.0112)} & \textbf{0.8290 (0.0089)} & \textbf{0.1502 (0.0043)} \\
\bottomrule
\end{tabular}
\end{table}

\begin{figure}[!htbp]
\centering
\includegraphics[width=0.99\textwidth]{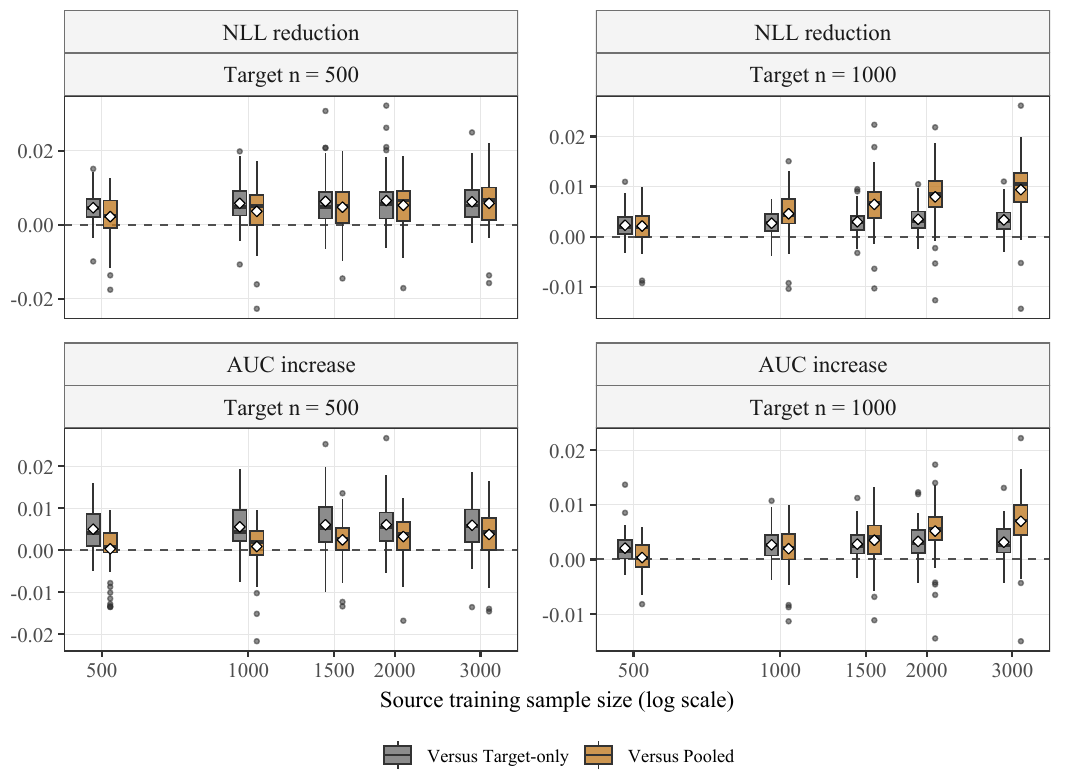}
\caption{Paired NLL reductions (top) and AUC increases (bottom) of the full-target algorithm over both baselines across 60 sampling and splitting repetitions. Positive values favor the full-target algorithm. Boxes show quartiles and medians, whiskers use $1.5$ times the interquartile range, dots show outliers and diamonds mark means. Each panel has its own vertical scale.}
\label{fig:support_paired}
\end{figure}

\Needspace{4\baselineskip}
Figure \ref{fig:support_paired} compares the NLL and AUC of the full-target algorithm against the target-only and pooled algorithms. The result shows the proposed full-target algorithm reduces NLL and improves AUC over the target-only and pooled algorithms across most of the 60 repetitions, with larger improvements over the pooled algorithm as $n_s$ increases. 

These results demonstrate the practical value of the proposed full-target algorithm for improving target prediction through transfer between unaligned ordinal response spaces.

Estimated target parameters from the three algorithms (Table~\ref{tab:methodEstimates}) and complementary target-size learning curves are provided in the \hyperlink{supplement:start}{Supplementary Material}.

\section{Discussion}\label{sec:Discussion}
In this paper, we introduced a high-dimensional transfer learning framework for structurally heterogeneous ordinal response models. Separate cumulative-link models with domain-specific regression coefficients, links and thresholds allow knowledge transfer across unaligned ordinal response spaces. We developed the sample-split and full-target algorithms based on $\ell_1$-penalized pooling and target correction. The sample-split algorithm assigns independent target parts to the two stages, while the full-target algorithm uses all target observations in both stages.

Theoretically, we characterized the estimation problem through a two-step transfer procedure and proved restricted strong convexity for the empirical losses. We derived $\ell_1$- and $\ell_2$-error bounds together with corresponding minimax lower bounds. The rates distinguish three sources of statistical error: pooled coefficient estimation, source--target discrepancy and target-threshold estimation. Under the stated conditions, the expected estimation error and excess risk rates for the sample-split algorithm are minimax optimal. When $n_s \lesssim n_t$, the full-target algorithm achieves the same optimal rates as the sample-split algorithm.
Numerical simulations illustrate the effect of coefficient discrepancy and favor the full-target algorithm over the sample-split algorithm. The real-data application further demonstrates the predictive gains of the full-target algorithm over the target-only and pooled baselines when transferring from a five-level functional outcome to binary hospital mortality.

Several directions remain open for future research. First, extending the framework to more general response structures, such as transfer between ordinal and nominal categorical outcomes, presents new modeling and theoretical challenges. Estimating domain-specific link shapes is another extension within the ordinal setting. Second, we characterize coefficient shifts through an $\ell_1$ radius, which aligns with the correction penalty. Developing methods and minimax theory under an $\ell_0$-sparse shift, as in high-dimensional linear regression and GLM transfer learning \citep{li2022linear,li2024glm}, is an interesting direction. Finally, constructing adaptive transfer procedures for severe or uninformative source shifts remains an important problem.

\section*{Acknowledgements}
This work is supported by the National Natural Science Foundation of China (Grant No. 12271286).

\bibliographystyle{plainnat}
\bibliography{ref}

@article{bastani2021,
  author = {Bastani, Hamsa},
  title = {Predicting with Proxies: Transfer Learning in High Dimension},
  journal = {Management Science}, volume = {67}, number = {5},
  pages = {2964--2984}, year = {2021}
}

@article{tcai2021,
  author = {Cai, Tony and Wei, Hongji},
  title = {Transfer Learning for Nonparametric Classification: Minimax Rate and Adaptive Classifier},
  journal = {The Annals of Statistics}, volume = {49}, number = {1},
  pages = {100--128}, year = {2021}
}

@misc{chang2024feature,
  author = {Chang, Jae Ho and Russo, Massimiliano and Paul, Subhadeep},
  title = {Heterogeneous Transfer Learning for High-Dimensional Regression with Feature Mismatch},
  howpublished = {arXiv preprint arXiv:2412.18081}, year = {2024}
}

@article{gu2025angle,
  author = {Gu, Tian and Han, Yi and Duan, Rui},
  title = {Robust Angle-Based Transfer Learning in High Dimensions},
  journal = {Journal of the Royal Statistical Society Series B: Statistical Methodology},
  volume = {87}, number = {3}, pages = {723--745}, year = {2025}
}

@article{kim2022medical,
  author = {Kim, Hee E and Cosa-Linan, Alejandro and Santhanam, Nandhini and Jannesari, Mahboubeh and Maros, Mate E and Ganslandt, Thomas},
  title = {Transfer Learning for Medical Image Classification: A Literature Review},
  journal = {BMC Medical Imaging}, volume = {22}, number = {1}, pages = {69}, year = {2022}
}

@article{kpotufe2021,
  author = {Kpotufe, Samory and Martinet, Guillaume},
  title = {Marginal Singularity, and the Benefits of Labels in Covariate-Shift},
  journal = {The Annals of Statistics}, volume = {49}, number = {6},
  pages = {3299--3323}, year = {2021}
}

@article{li2022linear,
  author = {Li, Sai and Cai, T. Tony and Li, Hongzhe},
  title = {Transfer Learning for High-Dimensional Linear Regression: Prediction, Estimation and Minimax Optimality},
  journal = {Journal of the Royal Statistical Society Series B}, volume = {84},
  number = {1}, pages = {149--173}, year = {2022}
}

@article{li2024glm,
  author = {Li, Sai and Zhang, Linjun and Cai, T. Tony and Li, Hongzhe},
  title = {Estimation and Inference for High-Dimensional Generalized Linear Models with Knowledge Transfer},
  journal = {Journal of the American Statistical Association}, volume = {119},
  number = {546}, pages = {1274--1285}, year = {2024}
}

@article{li2013computer,
  author = {Li, Wen and Duan, Lixin and Xu, Dong and Tsang, Ivor W.},
  title = {Learning with Augmented Features for Supervised and Semi-Supervised Heterogeneous Domain Adaptation},
  journal = {IEEE Transactions on Pattern Analysis and Machine Intelligence},
  volume = {36}, number = {6}, pages = {1134--1148}, year = {2013}
}

@misc{lin2022functional,
  author = {Lin, Haotian and Reimherr, Matthew},
  title = {On Hypothesis Transfer Learning of Functional Linear Models},
  howpublished = {arXiv preprint arXiv:2206.04277}, year = {2022}
}

@article{ma2023rkhs,
  author = {Ma, Cong and Pathak, Reese and Wainwright, Martin J.},
  title = {Optimally Tackling Covariate Shift in {RKHS}-Based Nonparametric Regression},
  journal = {The Annals of Statistics}, volume = {51}, number = {2},
  pages = {738--761}, year = {2023}
}

@article{negahban2012,
  author = {Negahban, Sahand N. and Ravikumar, Pradeep and Wainwright, Martin J. and Yu, Bin},
  title = {A Unified Framework for High-Dimensional Analysis of {$M$}-Estimators with Decomposable Regularizers},
  journal = {Statistical Science}, volume = {27}, number = {4},
  pages = {538--557}, year = {2012}
}

@article{peng2025hetero,
  author = {Peng, Yanjin and Wang, Lei},
  title = {Heterogeneity-Aware Transfer Learning for High-Dimensional Linear Regression Models},
  journal = {Computational Statistics \& Data Analysis}, volume = {206},
  pages = {108129}, year = {2025}
}

@inproceedings{shi2010transfer,
  author = {Shi, Xiaoxiao and Liu, Qiong and Fan, Wei and Yu, Philip S. and Zhu, Ruixin},
  title = {Transfer Learning on Heterogeneous Feature Spaces via Spectral Transformation},
  booktitle = {2010 IEEE International Conference on Data Mining},
  pages = {1049--1054}, publisher = {IEEE}, year = {2010}
}

@article{tian2023,
  author = {Tian, Ye and Feng, Yang},
  title = {Transfer Learning under High-Dimensional Generalized Linear Models},
  journal = {Journal of the American Statistical Association}, volume = {118},
  number = {544}, pages = {2684--2697}, year = {2023}
}

@book{np-tsybakov-book,
  author = {Tsybakov, Alexandre B.}, title = {Introduction to Nonparametric Estimation},
  publisher = {Springer}, year = {2009}
}

@inproceedings{wang2015speech,
  author = {Wang, Dong and Zheng, Thomas Fang},
  title = {Transfer Learning for Speech and Language Processing},
  booktitle = {2015 Asia-Pacific Signal and Information Processing Association Annual Summit and Conference (APSIPA)},
  pages = {1225--1237}, publisher = {IEEE}, year = {2015}
}

@article{zhao2017geo,
  author = {Zhao, Bei and Huang, Bo and Zhong, Yanfei},
  title = {Transfer Learning with Fully Pretrained Deep Convolution Networks for Land-Use Classification},
  journal = {IEEE Geoscience and Remote Sensing Letters}, volume = {14},
  number = {9}, pages = {1436--1440}, year = {2017}
}

@misc{zhao2023heterogeneous,
  author = {Zhao, Ruzhang and Kundu, Prosenjit and Saha, Arkajyoti and Chatterjee, Nilanjan},
  title = {Heterogeneous Transfer Learning for Building High-Dimensional Generalized Linear Models with Disparate Datasets},
  howpublished = {arXiv preprint arXiv:2312.12786}, year = {2023}
}

@misc{zheng2026hfs,
  author = {Zheng, Zejing and Huang, Rui and Zhao, Junlong},
  title = {Transfer Learning with Heterogeneous Feature Spaces in Linear Regression},
  howpublished = {arXiv preprint arXiv:2609.08526}, year = {2026}
}

@article{zheng2026discrete,
  author = {Zheng, Zejing and Zheng, Shengbing and Zhao, Junlong},
  title = {Transfer Learning for High Dimensional Data with Discrete Responses},
  journal = {Computational Statistics \& Data Analysis}, volume = {215},
  pages = {108292}, year = {2026}
}

@article{mccullagh1980,
  author = {McCullagh, Peter},
  title = {Regression Models for Ordinal Data},
  journal = {Journal of the Royal Statistical Society: Series B (Methodological)},
  volume = {42}, number = {2}, pages = {109--127}, year = {1980},
  doi = {10.1111/j.2517-6161.1980.tb01109.x}
}

@article{tutz2022ordinal,
  author = {Tutz, Gerhard},
  title = {Ordinal Regression: A Review and a Taxonomy of Models},
  journal = {WIREs Computational Statistics},
  volume = {14}, number = {2}, pages = {e1545}, year = {2022},
  doi = {10.1002/wics.1545}
}

@article{knaus1995support,
  author = {Knaus, W. A. and Harrell, F. E. and Lynn, J. and others},
  title = {The {SUPPORT} Prognostic Model: Objective Estimates of Survival for Seriously Ill Hospitalized Adults},
  journal = {Annals of Internal Medicine},
  volume = {122}, number = {3}, pages = {191--203}, year = {1995},
  doi = {10.7326/0003-4819-122-3-199502010-00007}
}

@misc{support2_data,
  author = {{Vanderbilt University Department of Biostatistics}},
  title = {{SUPPORT} Datasets: {SUPPORT2} Data and Variable Documentation},
  year = {2026},
  howpublished = {\url{https://hbiostat.org/data/repo/supportdesc}},
  note = {Accessed October 16, 2026}
}

\clearpage
\phantomsection
\hypertarget{supplement:start}{}
\begin{center}
\Large\bfseries Supplementary Material
\end{center}

\appendix
\numberwithin{equation}{section}
\numberwithin{table}{section}
\numberwithin{figure}{section}
\renewcommand{\theHequation}{\theHsection.\arabic{equation}}
\renewcommand{\theHtable}{\theHsection.\arabic{table}}
\renewcommand{\theHfigure}{\theHsection.\arabic{figure}}

\begingroup
\small
\setstretch{1.1}
\section{Additional Real-Data Results}\label{app:realData}
\subsection{Varying the Target Sample Size}\label{app:targetLearning}
These experiments investigate the performance under varying target sizes with a fixed source size. We hold the source sample size at $n_s=1000$ or $3000$ and vary $n_t\in\{100,200,500,1000,2000\}$. The data, clinical predictors, estimation procedures and training-only five-fold tuning are the same as in Section~\ref{sec:Real-Data} of the main paper. Each point averages 60 repeated patient splits with seeds 9401--9460 and 2500 target test patients.

\begin{figure}[!htbp]
\centering
\includegraphics[width=0.88\textwidth]{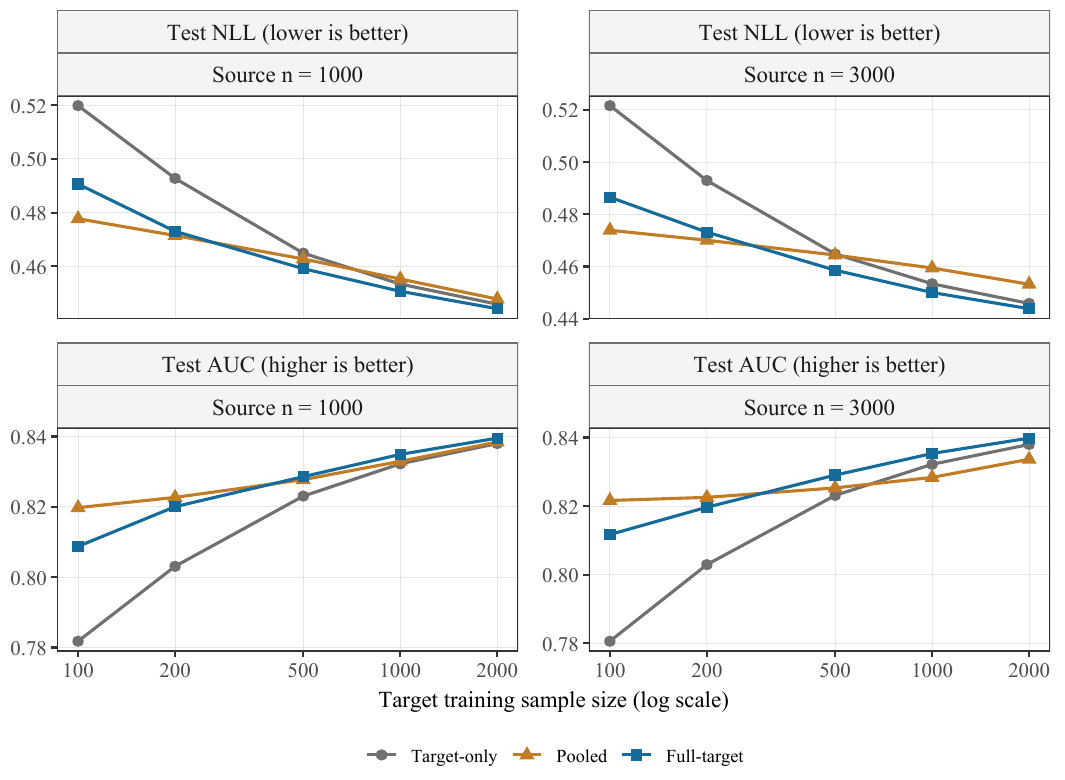}
\caption{Mean test NLL and AUC as the target sample size increases at fixed source sizes. Points average 60 repeated patient splits. Each panel uses its own vertical scale.}
\label{fig:suppTargetLearning}
\end{figure}

The full-target algorithm improves mean NLL over the target-only algorithm throughout the target-size grid. The pooled algorithm has the lowest mean NLL with 100 or 200 target observations; from $n_t=500$ onward, the full-target algorithm has lower mean NLL and higher mean AUC than both baselines at each source size. The larger target sample provides more information for correcting the pooled fit for hospital-mortality prediction.

\FloatBarrier
\clearpage
\subsection{Estimated Parameters of the Three Methods}\label{app:methodEstimates}
Table \ref{tab:methodEstimates} reports the target threshold and all 60 retained predictor coefficients for the target-only, pooled and full-target algorithms at $(n_s,n_t)=(3000,500)$. We use the first prespecified patient split, with seed 9401, and the training-only five-fold selections from the main experiment. The selected constants are $C_t=0.15$ for the target-only algorithm, $C_p=0.15$ for the pooled algorithm, and $(C_p,C_t)=(0.15,0.30)$ for the full-target algorithm. All three algorithms use the same training observations and feature transformation. Coefficients refer to the standardized predictors, and rows are ordered by the absolute coefficient estimated by the full-target algorithm.

\begingroup
\setstretch{1}
\small
\begin{longtable}{p{0.44\linewidth}rrr}
\caption{Estimated target parameters for the three algorithms on the same patient split.}\label{tab:methodEstimates}\\
\toprule
Parameter & \shortstack{the target-only\\algorithm} & \shortstack{the pooled\\algorithm} & \shortstack{the full-target\\algorithm} \\
\midrule
\endfirsthead
\multicolumn{4}{l}{Table \thetable\ (continued)}\\
\toprule
Parameter & \shortstack{the target-only\\algorithm} & \shortstack{the pooled\\algorithm} & \shortstack{the full-target\\algorithm} \\
\midrule
\endhead
\midrule
\multicolumn{4}{r}{Continued on the next page}\\
\endfoot
\bottomrule
\endlastfoot
Target threshold $\tau_1^t$ & 1.1691 & 1.1736 & 1.2134 \\
\texttt{adlp\_missing} & 0.4537 & 0.6169 & 0.6169 \\
\texttt{scoma} & 0.5582 & 0.4978 & 0.4978 \\
\texttt{dzgroup\_MOSF.w.Malig} & 0.3935 & 0.3069 & 0.3874 \\
\texttt{hday} & 0.3074 & 0.2147 & 0.3097 \\
\texttt{adlp} & 0.0000 & 0.2772 & 0.2772 \\
\texttt{edu\_missing} & 0.0524 & 0.2634 & 0.2634 \\
\texttt{age} & 0.1663 & 0.2588 & 0.2588 \\
\texttt{adls} & 0.0823 & 0.2212 & 0.2212 \\
\texttt{dzgroup\_COPD} & -0.2296 & -0.1438 & -0.1474 \\
\texttt{dzgroup\_CHF} & -0.1328 & -0.1443 & -0.1443 \\
\texttt{dzgroup\_Lung.Cancer} & 0.0000 & 0.1243 & 0.1243 \\
\texttt{hrt} & 0.0000 & 0.1171 & 0.1171 \\
\texttt{bili} & 0.0064 & 0.1129 & 0.1129 \\
\texttt{edu} & 0.1546 & 0.0000 & 0.0991 \\
\texttt{wblc} & 0.0000 & 0.0718 & 0.0718 \\
\texttt{ca\_metastatic} & 0.0000 & 0.0694 & 0.0694 \\
\texttt{meanbp} & -0.1404 & -0.0395 & -0.0676 \\
\texttt{sod} & 0.0000 & -0.0655 & -0.0655 \\
\texttt{num.co} & 0.0000 & 0.1237 & 0.0652 \\
\texttt{adls\_missing} & 0.0000 & 0.1080 & 0.0597 \\
\texttt{crea} & 0.0000 & 0.0545 & 0.0545 \\
\texttt{income\_Missing} & 0.2334 & 0.0502 & 0.0502 \\
\texttt{diabetes} & -0.0808 & -0.0496 & -0.0496 \\
\texttt{bun} & -0.0131 & -0.0486 & -0.0486 \\
\texttt{temp} & -0.0538 & -0.0477 & -0.0477 \\
\texttt{dementia} & -0.0171 & -0.0436 & -0.0436 \\
\texttt{income\_X..50k} & 0.0000 & -0.0406 & -0.0406 \\
\texttt{dzgroup\_Coma} & 0.0014 & 0.0399 & 0.0399 \\
\texttt{ca\_yes} & 0.0525 & 0.0120 & 0.0397 \\
\texttt{pafi} & -0.1006 & 0.0000 & -0.0382 \\
\texttt{alb} & 0.0000 & -0.0353 & -0.0353 \\
\texttt{resp} & 0.0158 & 0.0351 & 0.0351 \\
\texttt{race\_asian} & 0.1603 & 0.0331 & 0.0331 \\
\texttt{scoma\_missing} & 0.0000 & 0.0304 & 0.0304 \\
\texttt{dzgroup\_Colon.Cancer} & -0.2229 & -0.0169 & -0.0202 \\
\texttt{race\_black} & 0.0000 & 0.0202 & 0.0202 \\
\texttt{income\_X.25..50k} & 0.0000 & -0.0171 & -0.0171 \\
\texttt{urine} & 0.0000 & -0.0091 & -0.0091 \\
\texttt{income\_X.11..25k} & 0.0000 & -0.0087 & -0.0087 \\
\texttt{pafi\_missing} & -0.0412 & 0.0069 & 0.0069 \\
\texttt{race\_other} & 0.0000 & 0.0067 & 0.0067 \\
\texttt{sex\_male} & 0.0640 & 0.0035 & 0.0035 \\
\texttt{meanbp\_missing} & 0.0000 & 0.0034 & 0.0034 \\
\texttt{hrt\_missing} & 0.0000 & 0.0034 & 0.0034 \\
\texttt{resp\_missing} & 0.0000 & 0.0034 & 0.0034 \\
\texttt{temp\_missing} & 0.0000 & 0.0034 & 0.0034 \\
\texttt{sod\_missing} & 0.0000 & 0.0034 & 0.0034 \\
\texttt{ph} & 0.0000 & 0.0000 & 0.0000 \\
\texttt{glucose} & -0.0876 & 0.0000 & 0.0000 \\
\texttt{wblc\_missing} & 0.0000 & 0.0000 & 0.0000 \\
\texttt{alb\_missing} & 0.0000 & 0.0000 & 0.0000 \\
\texttt{bili\_missing} & -0.1366 & 0.0000 & 0.0000 \\
\texttt{crea\_missing} & 0.0000 & 0.0000 & 0.0000 \\
\texttt{ph\_missing} & 0.0000 & 0.0000 & 0.0000 \\
\texttt{glucose\_missing} & 0.0000 & 0.0000 & 0.0000 \\
\texttt{bun\_missing} & 0.0000 & 0.0000 & 0.0000 \\
\texttt{urine\_missing} & 0.0000 & 0.0000 & 0.0000 \\
\texttt{race\_hispanic} & 0.0312 & 0.0000 & 0.0000 \\
\texttt{race\_Missing} & 0.0000 & 0.0000 & 0.0000 \\
\end{longtable}
\endgroup
\endgroup
\clearpage

\section{Proof of Lemma~\ref{thm1}}\label{app:pool}
For each domain $d\in\{s,t\}$, recall that the negative log-likelihood loss for one observation $(x,y)\in\R^p\times[K_d]$ is
\[
 l_d(\beta,\tau;x,y)=-\log\{F_d(\tau_y-x^\top\beta)-F_d(\tau_{y-1}-x^\top\beta)\},
\]
where $\beta\in\R^p$, $\tau\in\R^{K_d-1}$ is an ordered threshold vector, $\tau_0=-\infty$ and $\tau_{K_d}=\infty$.
We first establish the derivative, score and curvature bounds under \texorpdfstring{\hyperlink{condition:C1}{C1}}{C1}--\texorpdfstring{\hyperlink{condition:C4}{C4}}{C4} in Section~\ref{subsec:assumptions} of the main paper, and then apply the penalized basic inequality to the coefficient and threshold blocks. For a domain $d\in\{s,t\}$, define $m_d=K_d-1$, and let $\mathbf1_{m_d}\in\R^{m_d}$ be the all-ones vector. Population centering gives the exact identity
\begin{equation}\label{eq:populationCentering}
 Z^{(d)}=X^{(d)}-\mu_d,\qquad a^d=\tau^d-\mathbf1_{m_d}\mu_d^\top\beta,\qquad l_d(\beta,\tau^d;X^{(d)},Y^d)=l_d(\beta,a^d;Z^{(d)},Y^d).
\end{equation}
The coefficient vector and threshold gaps are preserved. Estimation and parameter losses use the original coordinates. \texorpdfstring{\hyperlink{condition:C1}{C1}}{C1} also gives the uniform projection bound
\begin{equation}\label{eq:noncenteredProjection}
 \|v^\top X^{(d)}\|_{\psi_2}\le C(\sigma_x+M_\mu)\|v\|_2,\qquad v\in\R^p.
\end{equation}

For a pooled increment $v=(v_\beta^\top,v_{\tau^s}^\top,v_{\tau^t}^\top)^\top\in\R^{p+K_s+K_t-2}$, the blocks have dimensions $p,K_s-1,K_t-1$, respectively. Define its weighted norm by
\begin{equation}\label{eq:weightednorm}
 \|v\|_\rho^2=\|v_\beta\|_2^2+\rho_s\|v_{\tau^s}\|_2^2+\rho_t\|v_{\tau^t}\|_2^2,
\end{equation}
\noindent where $\rho_s=n_s/n_{\rm p}$ and $\rho_t=n_{t,1}/n_{\rm p}$, with $n_{\rm p}=n_s+n_{t,1}$ as in the main paper.
Write $v_\tau=(v_{\tau^s}^\top,v_{\tau^t}^\top)^\top\in\R^{K_s+K_t-2}$ and $\|v_\tau\|_\rho^2=\rho_s\|v_{\tau^s}\|_2^2+\rho_t\|v_{\tau^t}\|_2^2$.

For a convex differentiable empirical loss $\mathcal L$, when both $\vartheta$ and $\vartheta+w$ belong to its parameter space, define the Bregman increment $D_{\mathcal L}(w;\vartheta)=\mathcal L(\vartheta+w)-\mathcal L(\vartheta)-\langle\nabla\mathcal L(\vartheta),w\rangle$. Put $D_p(v)=D_{L_{\rm pool}}(v;\bar\phi)$ for $\bar\phi+v\in\Phi$, where $\bar\phi$ is the unique population pooled minimizer defined in condition \texorpdfstring{\hyperlink{condition:C5}{C5}}{C5}. Section~\ref{app:target} evaluates $D_{L_{t,2}}$ at feasible comparison centers fixed by the pooling observations.

\subsection{Properties of the Negative Log-Likelihood}
Fix a domain $d\in\{s,t\}$ and abbreviate its distribution function, density and negative log-density by $F,f,V$. Put $\eta=x^\top\beta$. For an interior category $1<y<K_d$, define $A_1=\tau_{y-1}-\eta$, $A_2=\tau_y-\eta$ and $D=A_2-A_1>0$. The probability of this category is $\pi_{12}=F(A_2)-F(A_1)$. Define the negative log-likelihood and its Hessian with respect to $(A_1,A_2)$ by
\[
 G(A_1,A_2)=-\log\pi_{12},\qquad H(A_1,A_2)=\nabla^2_{(A_1,A_2)}G(A_1,A_2)\in\R^{2\times2}.
\]
For $y=1$ and $y=K_d$, the integration intervals are $(-\infty,\tau_1-\eta]$ and $(\tau_{K_d-1}-\eta,\infty)$, respectively. Only one limit is finite in each case. The corresponding negative log-likelihood functions are $G_-(z)=-\log F(z)$ and $G_+(z)=-\log\{1-F(z)\}$, $z\in\R$. They are evaluated at $z=\tau_1-\eta$ for $y=1$ and $z=\tau_{K_d-1}-\eta$ for $y=K_d$.

\Needspace{6\baselineskip}
We use the notation and parameter spaces in Section~\ref{sec:Problem-Formulation} and the conditions in Section~\ref{subsec:assumptions} of the main paper. For a matrix $A=(A_{ij})$, $\|A\|_{\rm op}=\sup_{\|v\|_2=1}\|Av\|_2$ and $\|A\|_1=\max_j\sum_i|A_{ij}|$. The vector norms and Orlicz norms are those defined in Section~\ref{subsec:preliminaries} of the main paper. Unless stated otherwise, $c,C>0$ may change from line to line and are independent of $n_s,n_t,p,s_1,h,\varepsilon$.

\begin{lemma}\label{lem:linkRegularity}
Suppose \texorpdfstring{\hyperlink{condition:C4}{C4}}{C4} holds, and let $M$ be the upper bound on $V''$ in that condition.
\begin{enumerate}[label=(\roman*),leftmargin=2em]
\item The matrix $H(A_1,A_2)$ is positive definite for finite $A_1<A_2$. If $A_2-A_1\ge g$, then
\begin{equation}\label{eq:linkHessianUpper}
 \|H(A_1,A_2)\|_{\rm op}\le M+2g^{-2}.
\end{equation}
For each fixed $U>0$, its smallest eigenvalue has a positive lower bound over $-U\le A_1<A_2\le U$ with $A_2-A_1\ge g$.

\item The functions $G_-$ and $G_+$ satisfy
\[
 0<G_-''(z)\le M,\qquad 0<G_+''(z)\le M,\qquad z\in\R.
\]
For each fixed $U>0$, both second derivatives have positive lower bounds on $[-U,U]$:
\[
 \inf_{|z|\le U}\min\{G_-''(z),G_+''(z)\}>0.
\]

\item For $\tau\in\mathcal T_{K_d}(T,g)$ and $y\in[K_d]$, regard $l_d$ as a function of $\eta=x^\top\beta$, with $\tau$ and $y$ fixed. For every $\eta\in\R$,
\[
 0\le\partial_\eta^2l_d\le M.
\]
The first derivatives and the loss satisfy
\begin{equation}\label{eq:linkScoreLoss}
 \begin{aligned}
 |\partial_\eta l_d|+\|\nabla_\tau l_d\|_\infty&\le C(1+|\eta|),\\
 0\le l_d&\le C(1+\eta^2).
 \end{aligned}
\end{equation}
\end{enumerate}
\end{lemma}

\begin{proof}
Write $H=H(A_1,A_2)$, $r_i=f(A_i)/\pi_{12}$ and $\nu_i=V'(A_i)$ for $i=1,2$. Direct differentiation gives
\begin{equation}\label{eq:generalEndpointH}
 \nabla G(A_1,A_2)=(r_1,-r_2)^\top,\qquad H=\begin{pmatrix}r_1(r_1-\nu_1)&-r_1r_2\\-r_1r_2&r_2(r_2+\nu_2)\end{pmatrix}.
\end{equation}
Both diagonal entries of $H$ are positive. If $\nu_1>0$, strict convexity of $V$ gives $f(A_1+u)<f(A_1)e^{-\nu_1u}$ for $u>0$. Integrating gives $\pi_{12}<f(A_1)/\nu_1$, so $r_1>\nu_1$. If $\nu_1\le0$, then $r_1-\nu_1>0$ directly. If $\nu_2<0$, the corresponding inequality to the left of $A_2$ gives $\pi_{12}<f(A_2)/(-\nu_2)$, so $r_2+\nu_2>0$. If $\nu_2\ge0$, this inequality is immediate.

The determinant is
\begin{equation}\label{eq:generalEndpointDet}
 \det H=\frac{f(A_1)f(A_2)}{\pi_{12}^3}\{\nu_2f(A_1)-\nu_1f(A_2)-\nu_1\nu_2\pi_{12}\}.
\end{equation}
Since $V''>0$, we have $\nu_1<\nu_2$. If $\nu_1\le0\le\nu_2$, then $\nu_2f(A_1)-\nu_1f(A_2)-\nu_1\nu_2\pi_{12}>0$. When $\nu_1\nu_2>0$, differentiating $f/V'$ gives
\[
 \nu_2f(A_1)-\nu_1f(A_2)-\nu_1\nu_2\pi_{12}=\nu_1\nu_2\int_{A_1}^{A_2}\frac{f(z)V''(z)}{\{V'(z)\}^2}\,dz>0.
\]
Therefore, $H$ is positive definite. For the first and last categories, let $r_-(z)=f(z)/F(z)$ and $r_+(z)=f(z)/\{1-F(z)\}$. Differentiation gives
\[
 G_-''(z)=r_-(z)\{r_-(z)+V'(z)\},\qquad G_+''(z)=r_+(z)\{r_+(z)-V'(z)\}.
\]
The same strict-convexity inequalities, integrated over the two half-lines, show that $G_-''(z)>0$ and $G_+''(z)>0$. Continuity then gives positive uniform lower bounds on fixed compact sets.

To prove the upper bounds, let $Z$ have density $f$, and put $I=(A_1,A_2]$ and $e=(1,1)^\top$. Since $f$ is integrable and $V$ is strictly convex, there are $z_-<z_+$ such that $V'(z_-)<0<V'(z_+)$. Convexity at these two points gives exponential upper bounds for the tails of $f$. Also, $V''\le M$ implies $|V'(z)|\le|V'(0)|+M|z|$. These bounds justify differentiating the probability of a translated interval under the integral. We obtain
\begin{equation}\label{eq:locationCurvature}
 e^\top He=\E\{V''(Z)\mid Z\in I\}-\operatorname{Var}\{V'(Z)\mid Z\in I\}\le M.
\end{equation}
The quadratic form $e^\top He$ is nonnegative because $H$ is positive definite. Taking $I=(-\infty,A_2]$ or $I=(A_1,\infty)$ gives the upper bounds for $G_-''$ and $G_+''$. The same identities give $0\le\partial_\eta^2l_d\le M$.

Log-concavity of $f$ and Jensen's inequality give
\[
 \pi_{12}\ge D\int_0^1 f(A_1)^{1-u}f(A_2)^u\,du\ge D\sqrt{f(A_1)f(A_2)}.
\]
Hence $r_1r_2\le D^{-2}$. Since $H$ is positive definite,
\[
 \|H\|_{\rm op}\le\operatorname{tr}H=e^\top He+2r_1r_2\le M+2g^{-2},
\]
which proves \eqref{eq:linkHessianUpper}.

To bound the derivatives, integrate $f'=-V'f$ to obtain
\[
 r_1-r_2=\E\{V'(Z)\mid A_1<Z\le A_2\}.
\]
The absolute value of $r_1-r_2$ is at most $|V'(0)|+M\max(|A_1|,|A_2|)$. Since $r_1r_2\le g^{-2}$, we have $\min(r_1,r_2)\le g^{-1}$. Thus each $r_i$ is at most $g^{-1}+|r_1-r_2|$. For a finite threshold value $z$ in the first or last category, put $\delta_z=(1+|V'(z)|)^{-1}$. Taylor's inequality for $V$ on $[z,z+\delta_z]$ and $[z-\delta_z,z]$ gives
\[
 1-F(z)\ge e^{-1-M/2}\delta_zf(z),\qquad F(z)\ge e^{-1-M/2}\delta_zf(z).
\]
Thus $r_-(z)$ and $r_+(z)$ are at most $e^{1+M/2}(1+|V'(z)|)$. Since every finite argument of $F$ has absolute value at most $T+|\eta|$, the derivative bound in \eqref{eq:linkScoreLoss} follows.

Finally $V(z)\le V(0)+|V'(0)||z|+Mz^2/2$. Each category interval contains a subinterval of length $\min(g,1)$ whose points have absolute value at most $T+|\eta|+1$. Its probability is at least $\min(g,1)e^{-C(1+\eta^2)}$, proving the loss bound.
\end{proof}

For domain $d$, let $r_y=2$ when $1<y<K_d$ and $r_y=1$ when $y\in\{1,K_d\}$. For $1<y<K_d$, let $J_{d,y}(x)\in\R^{r_y\times(p+K_d-1)}$ be the Jacobian of $(A_1,A_2)^\top$ with respect to $(\beta,\tau)$. For $y=1$ or $y=K_d$, define $J_{d,y}(x)$ using the scalar $\tau_1-x^\top\beta$ or $\tau_{K_d-1}-x^\top\beta$, respectively. Define $H_{d,y}(\beta,\tau;x)\in\R^{r_y\times r_y}$ as $H(A_1,A_2)$ for an interior category, and as the $1\times1$ matrix containing $G_-''(\tau_1-\eta)$ or $G_+''(\tau_{K_d-1}-\eta)$ for the corresponding boundary category. The parameter Hessian is
\begin{equation}\label{eq:chainH}
 \nabla^2_{(\beta,\tau)}l_d(\beta,\tau;x,y)=J_{d,y}(x)^\top H_{d,y}(\beta,\tau;x)J_{d,y}(x).
\end{equation}
By \eqref{eq:chainH}, the negative log-likelihood is convex. Lemma \ref{lem:linkRegularity} bounds its derivatives in terms of $X$ and $X^\top\beta$. Together with \texorpdfstring{\hyperlink{condition:C1}{C1}}{C1}, these bounds justify differentiation under expectation for the population risks.

\subsection{Bounds for the Loss Gradients}
Fix $\varepsilon\in(0,1/4)$ as in Section~\ref{subsec:theorems} of the main paper and write
\[
 \Lambda_1=\log(p/\varepsilon),\qquad \Lambda_2=\log(1/\varepsilon),\qquad a=\lambda_{\rm pool}=A_\lambda\sqrt{\Lambda_1/n_{\rm p}},\qquad b=\lambda_t=A_\lambda\sqrt{\Lambda_1/n_{t,2}}.
\]
Choose $A_\tau>0$ sufficiently large and put
\[
 q_p=A_\tau\sqrt{(K_s+K_t-2)\Lambda_2/n_{\rm p}},\qquad q_t=A_\tau\sqrt{(K_t-1)\Lambda_2/n_{t,2}}.
\]
The next lemma bounds the pooled threshold gradient by $q_p$ and the centered target threshold gradient by $q_t$. Choose $A_\lambda,A_\tau$ sufficiently large for these bounds and the expected bounds in Section~\ref{app:target}. These constants are independent of $n_s,n_t,p,s_1,h,\varepsilon$.

After choosing $A_\lambda,A_\tau$, take $C_n>0$ sufficiently large and $c>0$ in \eqref{eq:rateConditions} sufficiently small. Since $n_{\rm p}\asymp n$ and $n_{t,2}\asymp n_t$, condition \eqref{eq:rateConditions} makes $s_1a^2$ and $hb$ sufficiently small. The pooling gradient and curvature bounds have total failure probability at most $\varepsilon$. The conditional target bounds have total failure probability at most another $\varepsilon$.

\begin{lemma}[Pooled and centered target gradients]\label{lem:grad_step1}
Assume \texorpdfstring{\hyperlink{condition:C1}{C1}}{C1}--\texorpdfstring{\hyperlink{condition:C5}{C5}}{C5} and the sample size condition in \eqref{eq:rateConditions}. Let $\eventdef{1}$ be the event defined by
\begin{equation}\label{eq:scores}
 \|\nabla_\beta L_{\rm pool}(\bar\phi)\|_\infty\le a/2,\qquad \Big(\sum_{d=s,t}\rho_d^{-1}\|\nabla_{\tau^d}L_{\rm pool}(\bar\phi)\|_2^2\Big)^{1/2}\le q_p.
\end{equation}
For sufficiently large $A_\lambda,A_\tau$, we have $\Pp(\eventref{1})\ge1-\varepsilon/2$.

Write $R_t(\psi)=\E L_{t,2}(\psi)$ for the target population loss. For any fixed $\psi_c\in\Psi$, define the centered correction gradient $G_t(\psi_c)=\nabla L_{t,2}(\psi_c)-\nabla R_t(\psi_c)\in\R^{p+K_t-1}$, with coefficient and threshold blocks $G_{t,\beta}\in\R^p$ and $G_{t,\tau}\in\R^{K_t-1}$. Let $\eventdef{2}$ be the event defined by
\begin{equation}\label{eq:centerScores}
 \|G_{t,\beta}(\psi_c)\|_\infty\le b/2,\qquad \|G_{t,\tau}(\psi_c)\|_2\le q_t.
\end{equation}
Then $\Pp(\eventref{2})\ge1-\varepsilon/4$. The constants do not depend on the choice of the fixed feasible center.
\end{lemma}
\begin{proof}
By Lemma \ref{lem:linkRegularity}, the derivative with respect to $\beta_j$ is bounded in absolute value by $C|X_j|(1+|X^\top\beta|)$. The sub-Gaussian projection bounds in \eqref{eq:noncenteredProjection} and the bounded coefficient norm therefore give a uniform bound on its sub-exponential norm. Here we use $\|UV\|_{\psi_1}\le C\|U\|_{\psi_2}\|V\|_{\psi_2}$ for real random variables $U,V$. The threshold derivatives have uniformly bounded sub-Gaussian norms. Since $\bar\phi$ is an interior minimizer of $R_{\rm pool}$, the expected pooled coefficient gradient and each expected threshold gradient are zero.

For each source or target observation, center its loss gradient by subtracting its expectation under the corresponding domain. Every observation has coefficient weight $1/n_{\rm p}$. Bernstein's inequality and a union bound over the $p$ coefficient coordinates give the first inequality in \eqref{eq:scores}. The standard deviations of the source and target threshold averages are bounded by $C\sqrt{\rho_s/n_{\rm p}}$ and $C\sqrt{\rho_t/n_{\rm p}}$, respectively. Sub-Gaussian concentration gives the second inequality in \eqref{eq:scores}. The two pooled gradient bounds hold together with failure probability at most $\varepsilon/2$.

At any fixed $\psi_c\in\Psi$, center the loss gradient of each target observation in $I_2$ by subtracting its population expectation. The same tail bounds hold uniformly over the choice of $\psi_c$. Bernstein's inequality for the coefficient coordinates and sub-Gaussian concentration for the threshold coordinates give \eqref{eq:centerScores} with failure probability at most $\varepsilon/4$. The condition $n_{t,2}\ge C_n\Lambda_1$ makes the linear term in Bernstein's inequality no larger than its square-root term.
\end{proof}
\subsection{Restricted Strong Convexity}
The next lemma gives restricted strong convexity (RSC) bounds for the empirical losses. The bounds hold uniformly over the parameter increments specified below.
\begin{lemma}\label{lem:rsc_step1}
Assume \texorpdfstring{\hyperlink{condition:C1}{C1}}{C1}--\texorpdfstring{\hyperlink{condition:C4}{C4}}{C4} and the sample size condition in \eqref{eq:rateConditions}. Let $\eventdef{3}$ be the event on which
the following inequality holds for every $v\in\R^{p+K_s+K_t-2}$ with $\bar\phi+v\in\Phi$:
\begin{equation}\label{eq:RSC}
 D_p(v)\ge\kappa\|v\|_\rho^2-C_Ra^2\|v_\beta\|_1^2.
\end{equation}
Then $\Pp(\eventref{3})\ge1-\varepsilon/2$.

For any fixed feasible target center $\psi_c\in\Psi$, let $\eventdef{4}$ be the event on which the following inequality holds for every $w=(w_\beta^\top,w_\tau^\top)^\top\in\R^{p+K_t-1}$ with $w_\beta\in\R^p$, $w_\tau\in\R^{K_t-1}$ and $\psi_c+w\in\Psi$:
\begin{equation}\label{eq:centerRSC}
 D_{L_{t,2}}(w;\psi_c)\ge\kappa\|w\|_2^2-C_Rb^2\|w_\beta\|_1^2.
\end{equation}
Then $\Pp(\eventref{4})\ge1-\varepsilon/4$. The constants $\kappa,C_R>0$ are independent of $n_s,n_t,p,s_1,h,\varepsilon$ and uniform over the choice of the fixed center. The final term in each inequality allows for the fact that the empirical Hessian need not be positive definite when $p$ exceeds the sample size.
\end{lemma}
\begin{proof}
Fix a domain $d$. For a sample of $N\ge1$ independent observations from this domain, define
\[
 \mathcal L_N(\beta,\tau)=N^{-1}\sum_{i=1}^N l_d(\beta,\tau;X_i,Y_i).
\]
Put $K=K_d$ and $m=K-1$. Let $\xi_*=(\beta_*^\top,\tau_*^\top)^\top$ be the true domain parameter, so $\beta_*=\beta_d^*$ and $\tau_*=\tau^{d*}$. Let $\xi_c=(\beta_c^\top,\tau_c^\top)^\top$ be any fixed point in $\mathcal B(B)\times\mathcal T_K(T,g)$. The center need not equal the true parameter. Let $R_{\rm diam}$ be an upper bound on the diameter of this parameter set. For a nonzero $\Delta=(\Delta_\beta^\top,\Delta_\tau^\top)^\top\in\R^{p+m}$ with $\xi_c+\Delta$ in the same set, write $r=\|\Delta\|_2\le R_{\rm diam}$ and $v=\Delta/r$. Fix $U,V>0$, to be chosen below. For $j\in[m]$, define
\begin{equation}\label{eq:truncatedclass}
 Z_j(v)=v_{\tau,j}-X^\top v_\beta,\quad W_j=\ind\{Y\in\{j,j+1\}\},\quad A_U=\{|X^\top\beta_*|\le U,\ |X^\top\beta_c|\le U\}.
\end{equation}
Define the bounded function
\[
\varphi_V(z)=\begin{cases}z^2,&|z|\le V,\\ z^2(2-|z|/V),&V<|z|<2V,\\ 0,&|z|\ge2V.\end{cases}
\]
This function is nonnegative, is no larger than $z^2$, and has Lipschitz constant at most $CV$. It keeps the quadratic contribution for $|z|\le V$ and removes large values of $|z|$.

Let $f_v(X,Y)=\ind_{A_U}\sum_{j=1}^mW_j\varphi_V(Z_j(v))$. If $f_v(X,Y)>0$, at least one $|Z_j(v)|$ is less than $2V$. Since $|v_{\tau,j}|\le1$, this implies $|X^\top v_\beta|\le2V+1$. On $A_U$, all finite arguments of $F$ along $\xi_c+t\Delta$, $0\le t\le1$, then have absolute value at most $T+U+R_{\rm diam}(2V+1)$. Lemma \ref{lem:linkRegularity} gives a fixed positive lower bound for the corresponding Hessians. Taylor's formula with an integral remainder and \eqref{eq:chainH} give
\begin{equation}\label{eq:lossCutoff}
 D_{l_d}(\Delta;\xi_c)\ge c_0 r^2 f_v(X,Y).
\end{equation}
If $f_v(X,Y)=0$, inequality \eqref{eq:lossCutoff} follows from convexity. Otherwise, the bound on the arguments of $F$ applies to every nonzero summand in $f_v$.

For every covariate value in $A_U$, the conditional probability $\Pp(Y\in\{j,j+1\}\mid X)$ is at least a fixed $w_0>0$. This follows from the bounded true thresholds, their positive gaps, and $|X^\top\beta_*|\le U$. Also,
\begin{equation}\label{eq:unitmoment}
 \E\sum_{j=1}^mZ_j(v)^2=m v_\beta^\top\Sigma_dv_\beta+\nrm{v_\tau-\mathbf1_m\mu_d^\top v_\beta}_2^2\ge c_1,\qquad \nrm v_2=1.
\end{equation}
The lower bound follows from \texorpdfstring{\hyperlink{condition:C1}{C1}}{C1} and $\|v_\tau\|_2^2\le2\|v_\tau-\mathbf1_m\mu_d^\top v_\beta\|_2^2+2mM_\mu^2\|v_\beta\|_2^2$. Thus the untruncated quadratic form is uniformly equivalent to $\|v_\beta\|_2^2+\|v_\tau\|_2^2$. Equation \eqref{eq:noncenteredProjection} also bounds all fourth moments uniformly. Put $Q(v)=\sum_{j=1}^mZ_j(v)^2$. For unit $v$, Cauchy--Schwarz and the sub-Gaussian tails give
\[
 \E[Q(v)\ind_{A_U^c}]\le\{\E Q(v)^2\}^{1/2}\Pp(A_U^c)^{1/2}\le C\Pp(A_U^c)^{1/2}.
\]
\[
 \E\sum_jZ_j(v)^2\ind\{|Z_j(v)|>V\}\le Ce^{-cV^2}.
\]
Choose $U$ and then $V$ sufficiently large that these two tail contributions are each at most $c_1/4$, uniformly over unit $v$. For this fixed $U$, the conditional probability bound $w_0$ is fixed and positive. Since $z^2-\varphi_V(z)\le z^2\ind\{|z|>V\}$,
\[
 \E f_v\ge w_0\{\E Q(v)-\E[Q(v)\ind_{A_U^c}]-\E\sum_jZ_j(v)^2\ind\{|Z_j(v)|>V\}\}\ge w_0c_1/2.
\]
Thus $\inf_{\nrm v_2=1}\E f_v\ge\mu$, where $\mu=w_0c_1/2>0$ is fixed independently of $n,p$.

Let $P$ be the single-domain observation distribution and $P_N$ its empirical measure, so $Pf=\E f(X,Y)$ and $P_Nf=N^{-1}\sum_{i=1}^Nf(X_i,Y_i)$. Write $A_{U,i}$ and $W_{ij}$ for $A_U$ and $W_j$ evaluated at observation $(X_i,Y_i)$. Let $\epsilon_i$ be independent Rademacher signs, independent of the observations, with $\Pp(\epsilon_i=1)=\Pp(\epsilon_i=-1)=1/2$. For $R\ge1$, define $\mathcal V_R=\{v\in\R^{p+m}:\|v\|_2=1,\ \|v_\beta\|_1\le R\}$. Symmetrization and scalar contraction, separately for $j\in[m]$, give
\begin{equation}\label{eq:cutoffcomplexity}
 \E\sup_{v\in\mathcal V_R}|(P_N-P)f_v|\le C\{R\sqrt{\log(2p)/N}+N^{-1/2}\}.
\end{equation}
Indeed the coefficient part is bounded by
\[
 R\cdot\E\nrm{N^{-1}\sum_i\epsilon_i\ind_{A_{U,i}}W_{ij}X_i}_\infty\le C R\sqrt{\log(2p)/N}.
\]
The summands $\epsilon_i\ind_{A_{U,i}}W_{ij}X_{ij'}$, for each coefficient coordinate $j'\in[p]$, are independent and have mean zero. Their sub-Gaussian norms are uniformly bounded because $|\ind_{A_{U,i}}W_{ij}X_{ij'}|\le|X_{ij'}|$. The part involving the threshold coordinates is at most $C/\sqrt N$. Since $f_v$ is bounded uniformly in $v$, the bounded-differences inequality adds $C\sqrt{u/N}$ to the expected supremum with failure probability at most $e^{-u}$.

Apply these bounds for $R=1,2,4,\ldots,2^{\lceil\log_2\sqrt p\rceil}$. There are $N_{\rm rad}=1+\lceil\log_2\sqrt p\rceil$ such values. Taking $u=\log(4N_{\rm rad}/\varepsilon)\le C\Lambda_1$ makes the total failure probability at most $\varepsilon/4$. For every unit vector $v$, one of these values is between $\max\{1,\|v_\beta\|_1\}$ and twice that quantity. A union bound therefore gives, simultaneously for all unit $v$,
\begin{equation}\label{eq:uniformcutoff}
 P_Nf_v\ge\mu-C\sqrt{\Lambda_1/N}\{\nrm{v_\beta}_1+1\}.
\end{equation}
Choose $C_n$ sufficiently large that $C\sqrt{\Lambda_1/N}\le\mu/4$. Young's inequality gives
\[
 C\sqrt{\Lambda_1/N}\|v_\beta\|_1\le\mu/4+(C^2/\mu)(\Lambda_1/N)\|v_\beta\|_1^2.
\]
Hence $P_Nf_v\ge\mu/2-C(\Lambda_1/N)\|v_\beta\|_1^2$. Multiplication by $c_0r^2$ in \eqref{eq:lossCutoff}, with $\kappa=c_0\mu/2$, gives
\begin{equation}\label{eq:domainRSC}
 D_{\mathcal L_N}(\Delta;\xi_c)\ge\kappa\nrm\Delta_2^2-C(\Lambda_1/N)\nrm{\Delta_\beta}_1^2.
\end{equation}
Apply \eqref{eq:domainRSC} to all source observations at $(\bar\beta,\bar\tau^s)$ and to the target observations in $I_1$ at $(\bar\beta,\bar\tau^t)$. Sum the inequalities with the pooling weights $\rho_s,\rho_t$. The quadratic terms give $\kappa\|v\|_\rho^2$. Each negative term has coefficient at most $C\Lambda_1/n_{\rm p}$. A union bound with total failure probability $\varepsilon/2$ gives \eqref{eq:RSC}. Applying the same argument to the $n_{t,2}$ observations in $I_2$ at a fixed $\psi_c\in\Psi$, with failure probability $\varepsilon/4$, gives \eqref{eq:centerRSC}. All constants are uniform over the fixed centers in the stated parameter spaces.
\end{proof}

\subsection{Error Bounds of the Pooled Estimator}
\begin{proof}[Completion of the proof of Lemma~\ref{thm1}]
Define $\eventdef{5}=\eventref{1}\cap\eventref{3}$. Then $\Pp(\eventref{5})\ge1-\varepsilon$. On $\eventref{5}$, set $v=\hat\phi-\bar\phi$, $S=\supp(\beta_s^*)$, $e_p=\|v\|_\rho$, and $H_S=\|\bar\beta_{S^c}\|_1$. Conditions \eqref{eq:bias} and \eqref{eq:shift} give $H_S\le(C_{\rm pool}+1)h$. Comparing the objective at $\hat\phi$ and $\bar\phi$, and applying the pooled gradient bounds, gives
\begin{equation}\label{eq:poolBI}
 D_p(v)+\tfrac a2\nrm{v_{\beta,S^c}}_1\le\tfrac{3a}2\nrm{v_{\beta,S}}_1+2aH_S+q_p\nrm{v_\tau}_\rho.
\end{equation}
Here we use $\|\bar\beta\|_1-\|\bar\beta+v_\beta\|_1\le\|v_{\beta,S}\|_1-\|v_{\beta,S^c}\|_1+2H_S$. Since $D_p(v)\ge0$ and $q_p/a\le C$ for fixed $K_s,K_t$, inequality \eqref{eq:poolBI} gives
\begin{equation}\label{eq:approxcone}
 \nrm{v_\beta}_1\le4\nrm{v_{\beta,S}}_1+4H_S+2(q_p/a)\nrm{v_\tau}_\rho\le C\sqrt{s_1}e_p+Ch.
\end{equation}
Combine \eqref{eq:poolBI}, \eqref{eq:RSC} and \eqref{eq:approxcone} to obtain
\begin{equation}\label{eq:poolsolve}
 \kappa e_p^2\le C\sqrt{s_1}a e_p+Cah+C a^2(s_1e_p^2+h^2).
\end{equation}
Condition \eqref{eq:rateConditions} makes $s_1a^2$ sufficiently small. Also $a\le b$, $ah\le bh\le c$ and $a^2h^2\le c ah$. Moving the term $Ca^2s_1e_p^2$ to the left-hand side and applying Young's inequality gives $e_p^2\le C(s_1a^2+ah)$. By \eqref{eq:approxcone}, $\|v_\beta\|_1\le C(s_1a+\sqrt{s_1ah}+h)\le C(s_1a+h)$. Finally, use $\|\bar\beta-\beta_t^*\|_1\le C_{\rm pool}h$, the triangle inequality and $ah\le(a^2+h^2)/2$ to obtain the bounds relative to $\beta_t^*$.
\end{proof}

\section{Proof of Theorem~\ref{thm2}}\label{app:target}
\subsection{Comparison Parameters for the Correction Step}
Let $\mathcal D_{\rm p}$ consist of all source observations and the target observations indexed by $I_1$. Work first on $\eventref{5}$ from Section~\ref{app:pool}, whose probability is at least $1-\varepsilon$. Define
\[
 v_\beta=\hat\beta_{\rm pool}-\bar\beta\in\R^p,\qquad \bar d=\beta_t^*-\bar\beta\in\R^p,\qquad u=\hat\beta_{\rm pool}-\beta_t^*=v_\beta-\bar d\in\R^p.
\]
Thus $v_\beta$ is the estimation error relative to $\bar\beta$, $\bar d$ is the difference between $\beta_t^*$ and $\bar\beta$, and $u$ is the estimation error relative to $\beta_t^*$. Lemma~\ref{thm1} and \texorpdfstring{\hyperlink{condition:C5}{C5}}{C5} give
\begin{equation}\label{eq:vproperties}
 \|v_\beta\|_2^2\le C(s_1a^2+ah),\qquad \|v_\beta\|_1\le C(s_1a+h),\qquad \|\bar d\|_1\le C_{\rm pool}h.
\end{equation}
For sufficiently small $c$ in \eqref{eq:rateConditions}, we have $\|v_\beta\|_2<r_0$. 

Define the following two comparison parameters:
\begin{equation}\label{eq:comparisonCenters}
 \psi_{c,0}=(\hat\beta_{\rm pool}^\top,\tau^{t*\top})^\top,\qquad \psi_{c,1}=((\beta_t^*+v_\beta)^\top,\tau^{t*\top})^\top.
\end{equation}
The parameters $\psi_{c,0}$ and $\psi_{c,1}$ serve as feasible comparison points for the target correction estimator. We compare the penalized target objective at $\hat\psi$ with its value at either parameter to derive the basic inequality for the correction step.
Both parameters in \eqref{eq:comparisonCenters} belong to $\Psi$ on $\eventref{5}$. Their distances from $\psi^*$ are $\|u\|_2$ and $\|v_\beta\|_2$, respectively. Conditional on $\mathcal D_{\rm p}$, they are fixed, and the observations in $I_2$ remain independent target observations. In the following argument, $\psi_c=(\beta_c^\top,\tau^{t*\top})^\top$ denotes the selected parameter.

\subsection{Error Bounds for the Correction Step}\label{app:conditional}
\texorpdfstring{\hyperlink{condition:C1}{C1}}{C1} gives $\E X^{(t)}X^{(t)\top}=\Sigma_t+\mu_t\mu_t^\top$ with operator norm at most $C_\Sigma+M_\mu^2$. Together with \eqref{eq:chainH} and the Hessian bounds in Lemma \ref{lem:linkRegularity}, this gives $\|\nabla^2R_t(\psi)\|_{\rm op}\le C$ uniformly on $\Psi$. Correct specification gives $\nabla R_t(\psi^*)=0$, so integration along the feasible segment yields
\begin{equation}\label{eq:meanScore}
 \|\nabla R_t(\psi_c)\|_2\le C\|\psi_c-\psi^*\|_2.
\end{equation}
Conditional on any pooling sample in $\eventref{5}$, apply Lemmas \ref{lem:grad_step1} and \ref{lem:rsc_step1} at the selected $\psi_c$. Define $\eventdef{6}=\eventref{2}\cap\eventref{4}$, where both events are evaluated at this $\psi_c$. Each event has conditional failure probability at most $\varepsilon/4$. It follows that
\[
 \Pp(\eventref{5}\cap\eventref{6})\ge1-2\varepsilon.
\]
All subsequent deterministic bounds hold on $\eventref{5}\cap\eventref{6}$.

Put $w=\hat\psi-\psi_c=(w_\beta^\top,w_\tau^\top)^\top$, where $w_\beta\in\R^p$ and $w_\tau\in\R^{K_t-1}$. Let $e_c=\|w\|_2$ and $r_c=\|\nabla R_t(\psi_c)\|_2+q_t$. If $\psi_c=\psi_{c,0}$, then $\beta_c-\hat\beta_{\rm pool}=0$ and we put $H_c=0$. If $\psi_c=\psi_{c,1}$, then $\beta_c-\hat\beta_{\rm pool}=\bar d$ and we put $H_c=\|\bar d\|_1$.

Optimality at the feasible center, the centered gradient bounds and \eqref{eq:meanScore} imply
\begin{equation}\label{eq:correctionBI}
 D_{L_{t,2}}(w;\psi_c)+\tfrac b2\|w_\beta\|_1\le2bH_c+r_ce_c.
\end{equation}
At $\psi_{c,1}$, we use $\|\bar d\|_1-\|\bar d+w_\beta\|_1\le2\|\bar d\|_1-\|w_\beta\|_1$. At $\psi_{c,0}$, the coefficient penalty is zero. Since the loss is convex, $D_{L_{t,2}}(w;\psi_c)\ge0$, and hence
\[
 \|w_\beta\|_1\le4H_c+2r_ce_c/b.
\]
Substitute this bound into \eqref{eq:centerRSC} and \eqref{eq:correctionBI} to obtain
\[
 \kappa e_c^2\le2bH_c+r_ce_c+C b^2H_c^2+C r_c^2e_c^2.
\]
If $r_c^2$ and $bH_c$ are sufficiently small, move the term $Cr_c^2e_c^2$ to the left-hand side and apply Young's inequality. This gives
\begin{equation}\label{eq:correctionOracle}
 e_c^2\le C(bH_c+r_c^2),\qquad \|w_\beta\|_1\le C(H_c+r_c^2/b).
\end{equation}
For the second inequality, use $r_c\sqrt{bH_c}/b\le C(H_c+r_c^2/b)$.

\subsection{High-Probability Error Bounds}\label{app:transferRates}
If $h\le b$, choose $\psi_{c,0}$. Equation \eqref{eq:vproperties}, \texorpdfstring{\hyperlink{condition:C5}{C5}}{C5} and $ah\le(a^2+h^2)/2$ give
\[
 \|u\|_2^2\le C(s_1a^2+h^2),\qquad \|u\|_1\le C(s_1a+h).
\]
We have $r_c^2\le C(s_1a^2+h^2+q_t^2)$. This is sufficiently small because $h^2\le hb$ and $n_{t,2}\ge C\Lambda_1$. By the triangle inequality and \eqref{eq:correctionOracle}, the squared Euclidean error relative to $\psi^*$ is at most $C(s_1a^2+h^2+q_t^2)$. For the coefficient $\ell_1$ error, use $a\le b$ and $h^2/b\le h$ to obtain the bound $C(s_1a+h+q_t^2/b)$.

If $h>b$, choose $\psi_{c,1}$. Here $r_c^2\le C(s_1a^2+ah+q_t^2)$ is sufficiently small because $ah\le bh\le c$. Also $H_c\le C_{\rm pool}h$. By \eqref{eq:correctionOracle} and the triangle inequality, the squared Euclidean error relative to $\psi^*$ is at most $C(s_1a^2+hb+q_t^2)$. Adding $\|v_\beta\|_1$ and using $(s_1a^2+ah)/b\le s_1a+h$ gives the same coefficient $\ell_1$ bound as in the case $h\le b$.

Consequently, on $\eventref{5}\cap\eventref{6}$,
\begin{equation}\label{eq:targetSquared}
 \|\hat\psi-\psi^*\|_2^2\le C\{s_1a^2+(hb\wedge h^2)+q_t^2\}.
\end{equation}
The threshold $\ell_1$ error is at most $\sqrt{K_t-1}\|\hat\psi-\psi^*\|_2\le C(s_1a+h+q_t)$ for fixed $K_t$, using $\sqrt{hb}\wedge h\le h$. Combining the blocks gives the $\ell_1$ bound $C(s_1a+h+q_t+q_t^2/b)$.

Because $\Lambda_2\le\Lambda_1$ and $K_t$ is fixed,
\[
 q_t+q_t^2/b\le C(K_t-1)\sqrt{\log(1/\varepsilon)/n_t}.
\]
Substituting the definitions of $a,b,q_t$ and using $n_{\rm p}\asymp n$ and $n_{t,2}\asymp n_t$ gives the explicit rates in \eqref{eq:targetrates}.

\subsection{Expected Error Bounds}\label{app:expectedBounds}
Fix $A_0=4$ and set $\varepsilon_E=\{p(n_t+1)\}^{-A_0}$. Write $\Lambda_{1,E}=\log(p/\varepsilon_E)$, $a_E=A_\lambda\sqrt{\Lambda_{1,E}/n_{\rm p}}$ and $b_E=A_\lambda\sqrt{\Lambda_{1,E}/n_{t,2}}$. These are the penalties in \eqref{eq:tuning} with $\varepsilon=\varepsilon_E$, and
\[
 \Lambda_{1,E}=\log p+A_0\log\{p(n_t+1)\}\asymp\log\{p(n_t+1)\}.
\]
Choose $A_\lambda,A_\tau,C_n$ sufficiently large and $c$ sufficiently small for the gradient and curvature bounds in Section~\ref{app:pool}.
Let $\eventdef{7}$ denote $\eventref{5}$ constructed with $\varepsilon=\varepsilon_E$ and penalties $a_E,b_E$. Then $\Pp(\eventref{7})\ge1-\varepsilon_E$.

On $\eventref{7}$, choose $\psi_c=\psi_{c,0}$ if $h\le b_E$ and $\psi_c=\psi_{c,1}$ otherwise. Given this pooling sample, let $\eventdef{8}$ be the event on which $\|G_{t,\beta}(\psi_c)\|_\infty\le b_E/2$ and \eqref{eq:centerRSC}, with $b=b_E$, holds for every $w\in\R^{p+K_t-1}$ such that $\psi_c+w\in\Psi$. Lemmas \ref{lem:grad_step1} and \ref{lem:rsc_step1} give $\Pp(\eventref{8}^c\mid\mathcal D_{\rm p})\le\varepsilon_E/2$ on $\eventref{7}$. Define
\[
 Q_c=\|\nabla_{\tau^t} L_{t,2}(\psi_c)-\nabla_{\tau^t} R_t(\psi_c)\|_2
\]
on $\eventref{7}$, and set $Q_c=0$ on its complement. Conditional on a pooling sample in $\eventref{7}$, $\psi_c$ is a fixed point in $\Psi$. The centered threshold derivatives are uniformly sub-Gaussian, so
\begin{equation}\label{eq:thresholdMoment}
 \E(Q_c^2\mid\mathcal D_{\rm p})\le C(K_t-1)/n_{t,2},\qquad \Pp(Q_c>q_0\mid\mathcal D_{\rm p})\le Ce^{-cn_{t,2}},
\end{equation}
where $q_0>0$ is a sufficiently small fixed constant. Define $\eventdef{9}=\eventref{7}\cap\eventref{8}\cap\{Q_c\le q_0\}$. On $\eventref{9}$, the inequalities in Sections~\ref{app:conditional} and \ref{app:transferRates} hold with $a_E,b_E,Q_c$ in place of $a,b,q_t$. Taking conditional expectations and then expectations over $\mathcal D_{\rm p}$ gives $\E[\ind_{\eventref{7}}Q_c^2]\le C(K_t-1)/n_{t,2}$.

For sufficiently large $C_n$, condition $n_{t,2}\ge C_n\Lambda_{1,E}$ gives $\Pp(\eventref{9}^c)\le C\{p(n_t+1)\}^{-A_0}$. Since $\hat\psi,\psi^*\in\Psi$, their squared Euclidean distance is at most $C$ and their $\ell_1$ distance is at most $C\sqrt p$. The contributions from $\eventref{9}^c$ are therefore at most $C/n_{t,2}$ and $C/\sqrt{n_{t,2}}$, respectively. Taking expectations of the squared-error bound gives
\begin{equation}\label{eq:expectedSquared}
 \E\|\hat\psi-\psi^*\|_2^2\le C\{s_1a_E^2+(hb_E\wedge h^2)+(K_t-1)/n_{t,2}\}.
\end{equation}
For the $\ell_1$ terms, Cauchy--Schwarz gives $\E Q_c\le C\sqrt{(K_t-1)/n_{t,2}}$, and $\E Q_c^2/b_E\le C(K_t-1)/\sqrt{n_{t,2}}$ since $\Lambda_{1,E}\ge1$. Jensen's inequality and the deterministic $\ell_1$ bounds yield \eqref{eq:expectedRates}, using $n_{\rm p}\asymp n$ and $n_{t,2}\asymp n_t$.

\section{Proof of Lower Bounds in Theorems~\ref{thm3} and \ref{thm4}}\label{app:lower}
This section proves the parameter-estimation lower bounds in Theorem~\ref{thm3} and the excess-risk lower bound in Theorem~\ref{thm4}. Section~\ref{app:KLHellinger} compares parameter distances with KL divergence and Hellinger distance. Section~\ref{app:lowerProofs} constructs three finite parameter families and derives the lower bounds. Write $\Theta=\Theta(s_1,h)$ and let $\psi(\theta)$ denote the target parameter, as in Section~\ref{subsec:assumptions} of the main paper.

\Needspace{10\baselineskip}
\subsection{KL and Hellinger Bounds}\label{app:KLHellinger}
The following proposition supplies the distributional bounds used in the parameter constructions. Its KL upper bound controls the difficulty of distinguishing the candidate parameters from the observed samples. Its Hellinger lower bound will be used for probability prediction.

\begin{proposition}[Bounds for KL Divergence and Hellinger Distance]\label{prop:KL}
Assume \texorpdfstring{\hyperlink{condition:C1}{C1}}{C1}, \texorpdfstring{\hyperlink{condition:C3}{C3}}{C3} and \texorpdfstring{\hyperlink{condition:C4}{C4}}{C4}. Fix a domain $d\in\{s,t\}$, put $K=K_d$, and write $\xi=(\beta^\top,\tau^\top)^\top\in\R^{p+K-1}$. In this subsection, write $F=F_d$ and $f=f_d$, and abbreviate $l_d$ and $\pi_{d,y}$ by $l$ and $\pi_y$. Let $P_\xi$ denote the joint distribution of one observation $(X,Y)$ under parameter $\xi$. The marginal distribution $P_X$ of $X$ is the same for all $\xi$, so $P_\xi(dx,\{y\})=\pi_y(x;\xi)P_X(dx)$.

For distributions $P,Q$ dominated by a measure $\nu$, define $H^2(P,Q)=\int(\sqrt{dP/d\nu}-\sqrt{dQ/d\nu})^2\,d\nu$ and $\operatorname{TV}(P,Q)=\frac12\int|dP/d\nu-dQ/d\nu|\,d\nu$. The following bound holds uniformly over $\xi_0\in\mathcal B(B-r_0)\times\mathcal T_K(T-r_0,g+r_0)$ and $\xi_1\in\mathcal B(B)\times\mathcal T_K(T,g)$:
\begin{equation}\label{eq:KLequiv}
 c\|\xi_1-\xi_0\|_2^2\le\KL(P_{\xi_0},P_{\xi_1})\le C\|\xi_1-\xi_0\|_2^2.
\end{equation}
For arbitrary $\xi_0,\xi_1\in\mathcal B(B)\times\mathcal T_K(T,g)$,
\begin{equation}\label{eq:Hequiv}
 H^2(P_{\xi_0},P_{\xi_1})\ge c\|\xi_1-\xi_0\|_2^2.
\end{equation}
\end{proposition}

\begin{proof}
Put $m=K-1$, $\Delta=\xi_1-\xi_0=(\Delta_\beta^\top,\Delta_\tau^\top)^\top$, and $\xi_u=\xi_0+u\Delta=(\beta_u^\top,\tau_u^\top)^\top$ for $0\le u\le1$. Here $\Delta_\beta\in\R^p$ and $\Delta_\tau\in\R^m$. Write $\E_{\xi_0}$ for expectation under $P_{\xi_0}$, and $\E_X$ or $\E$ for expectation over $X\sim P_X$. The coefficient ball and threshold set are convex, so the whole segment belongs to $\mathcal B(B)\times\mathcal T_K(T,g)$.

We first derive the KL identity. Since the covariate distribution is the same under both parameters,
\[
 \begin{aligned}\KL(P_{\xi_0},P_{\xi_1})&=\E_{\xi_0}\log\frac{\pi_Y(X;\xi_0)}{\pi_Y(X;\xi_1)}\\&=\E_{\xi_0}\{l(\xi_1;X,Y)-l(\xi_0;X,Y)\}.\end{aligned}
\]
The conditional mean of the gradient at the true parameter is zero:
\[
 \E_{\xi_0}\{\nabla l(\xi_0;X,Y)\mid X\}=-\sum_{y=1}^K\pi_y(X;\xi_0)\nabla\log\pi_y(X;\xi_0)=-\nabla\sum_{y=1}^K\pi_y(X;\xi_0)=0.
\]
Lemma~\ref{lem:linkRegularity} and \texorpdfstring{\hyperlink{condition:C1}{C1}}{C1} justify differentiation under expectation. Taylor's formula with an integral remainder therefore gives
\begin{equation}\label{eq:integralTaylor}
 \KL(P_{\xi_0},P_{\xi_1})=\int_0^1(1-u)\E_{\xi_0}[\Delta^\top\nabla^2l(\xi_u;X,Y)\Delta]\,du.
\end{equation}

For the upper bound, the Jacobian formula \eqref{eq:chainH} and the bounds in Lemma~\ref{lem:linkRegularity} give
\[
 \Delta^\top\nabla^2l(\xi_u;X,Y)\Delta\le C\sum_{j=1}^m(\Delta_{\tau,j}-X^\top\Delta_\beta)^2\le C\{(X^\top\Delta_\beta)^2+\|\Delta_\tau\|_2^2\}.
\]
By \texorpdfstring{\hyperlink{condition:C1}{C1}}{C1}, $\E XX^\top=\Sigma_d+\mu_d\mu_d^\top$ has operator norm at most $C_\Sigma+M_\mu^2$. Taking expectations and using $\int_0^1(1-u)\,du=1/2$ proves the upper bound in \eqref{eq:KLequiv}.

For the lower bound, choose a fixed $U>0$ and restrict attention to
\[
 \mathcal A=\{|X^\top\beta_0|\le U,\ |X^\top\beta_1|\le U\}.
\]
On $\mathcal A$, convexity gives $|X^\top\beta_u|\le U$ for every $u\in[0,1]$. Each finite argument $\tau_{u,j}-X^\top\beta_u$ is then in $[-T-U,T+U]$, and adjacent arguments are separated by at least $g$. Lemma~\ref{lem:linkRegularity} gives positive lower bounds for the eigenvalues of $H(A_1,A_2)$ and for $G_-''$ and $G_+''$ on this bounded set. The true conditional class probabilities $\pi_y(X;\xi_0)$ also have a positive lower bound on $\mathcal A$. For the first and last categories this follows from positivity of $F(-T-U)$ and $1-F(T+U)$. For an interior category it follows from positivity of $f$ on a bounded interval and a category interval of length at least $g$.

To see how these bounds act on the parameter direction, put $z_j=\Delta_{\tau,j}-X^\top\Delta_\beta$, $j\in[m]$. For $1<y<K$, the directional Hessian equals $(z_{y-1},z_y)H(A_1,A_2)(z_{y-1},z_y)^\top$. For $y=1$ and $y=K$, it equals $G_-''(\tau_{u,1}-X^\top\beta_u)z_1^2$ and $G_+''(\tau_{u,m}-X^\top\beta_u)z_m^2$, respectively. Thus every $z_j^2$ occurs in the contributions of categories $j$ and $j+1$. Averaging over $Y$ conditional on $X$ gives a constant $c_U>0$ such that
\begin{equation}\label{eq:truncCurv}
 \E_{\xi_0}[\Delta^\top\nabla^2l(\xi_u;X,Y)\Delta\mid X]\ind_{\mathcal A}\ge c_U\sum_{j=1}^m(\Delta_{\tau,j}-X^\top\Delta_\beta)^2\ind_{\mathcal A}.
\end{equation}
This inequality holds for every $u\in[0,1]$.

We next show that truncation preserves a fixed fraction of the expected quadratic form. Define $Q_\Delta(X)=\sum_{j=1}^m(\Delta_{\tau,j}-X^\top\Delta_\beta)^2$. Expanding around $\mu_d$ gives
\[
 \E Q_\Delta(X)=m\Delta_\beta^\top\Sigma_d\Delta_\beta+\|\Delta_\tau-\mathbf1_m\mu_d^\top\Delta_\beta\|_2^2.
\]
The covariance lower bound controls $\|\Delta_\beta\|_2^2$. Also,
\[
 \|\Delta_\tau\|_2^2\le2\|\Delta_\tau-\mathbf1_m\mu_d^\top\Delta_\beta\|_2^2+2mM_\mu^2\|\Delta_\beta\|_2^2.
\]
Together with the covariance upper bound, these inequalities imply $c\|\Delta\|_2^2\le\E Q_\Delta\le C\|\Delta\|_2^2$. The projection bound \eqref{eq:noncenteredProjection} gives $\E Q_\Delta^2\le C\|\Delta\|_2^4$. It also gives $\Pp(\mathcal A^c)\le Ce^{-cU^2}$ uniformly over the bounded coefficient vectors. Consequently,
\[
 \E[Q_\Delta\ind_{\mathcal A^c}]\le(\E Q_\Delta^2)^{1/2}\Pp(\mathcal A^c)^{1/2}\le C\|\Delta\|_2^2e^{-cU^2}.
\]
Choose $U$ sufficiently large that this is at most half of the lower bound for $\E Q_\Delta$. Then $\E[Q_\Delta\ind_{\mathcal A}]\ge c\|\Delta\|_2^2$. The loss is convex, so its directional Hessian is nonnegative outside $\mathcal A$. Combining \eqref{eq:truncCurv} with \eqref{eq:integralTaylor} now proves the KL lower bound.

The same calculation also bounds the population Hessian at any candidate parameter $\xi\in\mathcal B(B)\times\mathcal T_K(T,g)$. The conditional Hessian calculation leading to \eqref{eq:truncCurv} holds for any direction. Take $\xi_1=\xi$, $u=1$ and a direction $v=(v_\beta^\top,v_\tau^\top)^\top\in\R^{p+m}$, and define $Q_v(X)=\sum_{j=1}^m(v_{\tau,j}-X^\top v_\beta)^2$. The moment and truncation bounds above apply to $Q_v$ as well. Thus
\[
 c\|v\|_2^2\le v^\top\E_{\xi_0}\nabla^2l(\xi;X,Y)v\le C\|v\|_2^2,\qquad v\in\R^{p+m}.
\]

It remains to prove \eqref{eq:Hequiv}. Both parameters may now range over the outer parameter set. Use the same event $\mathcal A$ and choose $U$ as above. Given $X=x$, let $p_i(x)=(\pi_1(x;\xi_i),\ldots,\pi_K(x;\xi_i))$, $i=0,1$. For each $j\in[m]$, the difference between their cumulative probabilities is
\[
 \sum_{y=1}^j\{\pi_y(x;\xi_1)-\pi_y(x;\xi_0)\}=F(\tau_{1,j}-x^\top\beta_1)-F(\tau_{0,j}-x^\top\beta_0).
\]
On $\mathcal A$, the mean-value theorem and the positive minimum of $f$ on $[-T-U,T+U]$ bound the absolute value of this difference below by $c|\Delta_{\tau,j}-x^\top\Delta_\beta|$. Its absolute value is at most $\operatorname{TV}(p_0(x),p_1(x))$. Cauchy--Schwarz gives $\operatorname{TV}(p_0(x),p_1(x))\le H(p_0(x),p_1(x))$. Squaring and summing over the fixed number $m$ of cumulative probabilities yields
\begin{equation}\label{eq:cumulativeH}
 H^2(p_0(X),p_1(X))\ge c\sum_{j=1}^m(\Delta_{\tau,j}-X^\top\Delta_\beta)^2\ind_{\mathcal A}.
\end{equation}
Because the marginal distribution of $X$ is common to both parameters, $H^2(P_{\xi_0},P_{\xi_1})=\E_X H^2(p_0(X),p_1(X))$. Taking expectations in \eqref{eq:cumulativeH} and using the bound for $\E[Q_\Delta\ind_{\mathcal A}]$ proves \eqref{eq:Hequiv}.
\end{proof}

\subsection{Proofs of the Minimax Lower Bounds}\label{app:lowerProofs}
We now begin the main proof of Theorem~\ref{thm3} and the lower bound in Theorem~\ref{thm4}.
\begin{proof}
Fix the two Gaussian covariate distributions $Q_d=N(\mu_d,I_p)$, $d=s,t$, as in Theorem~\ref{thm3}. For a full parameter $\theta\in\Theta$, let $P_{d,\theta}$ denote the joint distribution of one observation in domain $d$. Let $\Pp_\theta$ denote the distribution of all training observations and $\E_\theta$ expectation under $\Pp_\theta$. Sample independence gives
\[
 \Pp_\theta=P_{s,\theta}^{\otimes n_s}\otimes P_{t,\theta}^{\otimes n_t},\qquad \KL(\Pp_\theta,\Pp_{\theta'})=n_s\KL(P_{s,\theta},P_{s,\theta'})+n_t\KL(P_{t,\theta},P_{t,\theta'}).
\]
We use this identity and Proposition~\ref{prop:KL} to control the sample KL divergences. Put $b_0=\sqrt{\log p/n_t}$. Define the Hamming distance $d_{\rm Ham}(\omega,\omega')=\#\{j:\omega_j\ne\omega_j'\}$. 

\subsubsection{Parameter Constructions and Verification of C5}\label{app:packingMembership}
Fix baseline thresholds $\tau^{d,0}$ in the interior of $\mathcal T_{K_d}(T-r_0,g+r_0)$, $d=s,t$. For each construction, $\theta^\omega$ denotes the full parameter with blocks $\beta_s^\omega,\tau^{s,\omega},\beta_t^\omega,\tau^{t,\omega}$. We use the following three constructions, with a separate choice of amplitude $\gamma$ in each. The positive constants $c_0,c_1,c_2,c_3$ are chosen sufficiently small in the corresponding lower-bound arguments.
\begin{enumerate}[label=(\roman*),ref=\roman*,leftmargin=2em]
\item\label{constr:shared}(Shared Coefficient) For $\omega\in\{0,1\}^p$ with $\|\omega\|_0=s_1$, set $\gamma=c_0\sqrt{\log(ep/s_1)/n}$ and
\[
 \beta_s^\omega=\beta_t^\omega=\gamma\omega,\qquad \tau^{s,\omega}=\tau^{s,0},\qquad \tau^{t,\omega}=\tau^{t,0}.
\]
This construction varies the common sparse coefficient and keeps both threshold vectors fixed.
\item\label{constr:shift}(Coefficient Shift) Set
\[
 (k,\gamma)=\begin{cases}(\max\{1,\lfloor c_1h/b_0\rfloor\},c_2b_0),&h\ge b_0,\\(1,c_2h),&0\le h<b_0,\end{cases}
\]
For $\omega\in\{0,1\}^p$ with $\|\omega\|_0=k$, define
\[
 \beta_s^\omega=0_p,\qquad \beta_t^\omega=\gamma\omega,\qquad \tau^{s,\omega}=\tau^{s,0},\qquad \tau^{t,\omega}=\tau^{t,0}.
\]
The source distribution is identical for all indices, so only target observations distinguish these candidates. For $h=0$, all candidates reduce to the same zero-coefficient parameter.
\item\label{constr:thresholds}(Varying Target Thresholds) For $\omega\in\{0,1\}^{K_t-1}$, set $\gamma=c_3/\sqrt{n_t}$ and
\[
 \beta_s^\omega=\beta_t^\omega=0_p,\qquad \tau^{s,\omega}=\tau^{s,0},\qquad \tau^{t,\omega}=\tau^{t,0}+\gamma\omega.
\]
This construction varies only the target thresholds.
\end{enumerate}

We now verify that the three constructions satisfy \texorpdfstring{\hyperlink{condition:C5}{C5}}{C5}. Fix a candidate. Write its true parameter blocks as $\beta_s^*$, $\tau^{s*}$, $\beta_t^*$ and $\tau^{t*}$. If $\beta_s^*=\beta_t^*$, each domain risk is minimized at the common coefficient and its own true thresholds. Their positively weighted sum has the same minimizer. The true parameters have margin $r_0$ in $\Phi$, so the pooled minimizer is interior and its coefficient bias is zero. This verifies \texorpdfstring{\hyperlink{condition:C5}{C5}}{C5} for constructions~(\ref{constr:shared}) and (\ref{constr:thresholds}).

For construction~(\ref{constr:shift}), put $\zeta=\beta_t^*$ and recall that $\beta_s^*=0_p$. Consider candidates with $\|\zeta\|_1\le h$ and $\|\zeta\|_2\le B-r_0$. We derive a fixed radius for $\|\zeta\|_2$ that ensures \texorpdfstring{\hyperlink{condition:C5}{C5}}{C5}; Section~\ref{app:targetShift} verifies this radius condition for the chosen amplitudes. The case $\zeta=0_p$ was covered above. Suppose $\zeta\ne0_p$. Set $m_d=K_d-1$, $Z^{(d)}=X^{(d)}-\mu_d$ and $a_d^*=\tau^{d*}-\mathbf1_{m_d}\mu_d^\top\beta_d^*$. Define
\[
 R_d(\beta,\tau)=\E l_d(\beta,\tau;X^{(d)},Y^d),\qquad \widetilde R_d(\beta,a)=\E l_d(\beta,a;Z^{(d)},Y^d).
\]
The centering identity \eqref{eq:populationCentering} gives $R_d(\beta,\tau)=\widetilde R_d(\beta,\tau-\mathbf1_{m_d}\mu_d^\top\beta)$, and $Z^{(d)}\sim N(0,I_p)$. Work first on the larger compact convex set
\begin{equation}\label{eq:auxiliaryPoolSpace}
 \widetilde T=T+M_\mu B,\qquad \widetilde\Phi=\mathcal B(B)\times\mathcal T_{K_s}(\widetilde T,g)\times\mathcal T_{K_t}(\widetilde T,g).
\end{equation}
Every centered image of a point in $\Phi$ belongs to $\widetilde\Phi$: subtracting a common scalar preserves the threshold gaps, and the threshold magnitudes are at most $T+M_\mu B$. Moreover,
\[
 \|a_d^*\|_\infty\le T-r_0+M_\mu(B-r_0)=\widetilde T-(1+M_\mu)r_0,
\]
and the true centered threshold gaps are at least $g+r_0$.

Let $\check\phi=(\check\beta^\top,\check a^{s\top},\check a^{t\top})^\top\in\R^{p+m_s+m_t}$ minimize $\sum_{d=s,t}\rho_d\widetilde R_d$ over $\widetilde\Phi$. The population Hessian argument in the proof of Proposition~\ref{prop:KL}, applied with centered covariates and bound $\widetilde T$, gives strict convexity and hence uniqueness. If an orthogonal matrix $O\in\R^{p\times p}$ satisfies $O\zeta=\zeta$, then $OZ^{(d)}$ has the same distribution as $Z^{(d)}$ and leaves each true linear predictor unchanged. The domain risks and $\widetilde\Phi$ are therefore invariant under $\beta\mapsto O\beta$. Uniqueness implies $O\check\beta=\check\beta$ for every such $O$. In particular, reflecting any direction perpendicular to $\zeta$ leaves $\check\beta$ unchanged, so $\check\beta=t\zeta$ for some scalar $t$.

For each domain define the scalar function obtained by minimizing over its thresholds,
\[
 \widetilde r_d(t)=\min_{a\in\mathcal T_{K_d}(\widetilde T,g)}\widetilde R_d(t\zeta,a),\qquad |t|\|\zeta\|_2\le B.
\]
To verify strict convexity, take distinct $t_1,t_2$ and minimizing threshold vectors $a_1,a_2$. Their convex combination is feasible, and strict joint convexity gives, for $u\in(0,1)$,
\[
 \widetilde r_d((1-u)t_1+ut_2)\le\widetilde R_d(((1-u)t_1+ut_2)\zeta,(1-u)a_1+ua_2)<(1-u)\widetilde r_d(t_1)+u\widetilde r_d(t_2).
\]
Correct specification places the minimum of $\widetilde r_s$ at $t=0$ and that of $\widetilde r_t$ at $t=1$. If $t<0$, moving to zero decreases both domain risks; if $t>1$, moving to one decreases both. Their positively weighted sum is therefore minimized at $t\in[0,1]$. In particular, $\|\check\beta\|_2\le\|\zeta\|_2$ and $\|\check\beta-\beta_d^*\|_2\le\|\zeta\|_2$.

For fixed $\check\beta$, each $\check a^d$ minimizes its own domain risk, because $\rho_d>0$. Comparing it with the feasible $a_d^*$ and applying both KL bounds in Proposition~\ref{prop:KL} gives positive constants $c_*,C_*$ such that
\begin{equation}\label{eq:profileStability}
 \begin{aligned}
 c_*\|\check a^d-a_d^*\|_2^2&\le\widetilde R_d(\check\beta,\check a^d)-\widetilde R_d(\beta_d^*,a_d^*)\\
 &\le\widetilde R_d(\check\beta,a_d^*)-\widetilde R_d(\beta_d^*,a_d^*)\le C_*\|\check\beta-\beta_d^*\|_2^2.
 \end{aligned}
\end{equation}
The first inequality drops the nonnegative coefficient part of the KL lower bound. The last inequality compares parameters with identical thresholds. Put $C_{\rm pr}=\max\{1,\sqrt{C_*/c_*}\}$ and return to the original coordinates by setting $\tau^{d,\dagger}=\check a^d+\mathbf1_{m_d}\mu_d^\top\check\beta$. Then
\[
 \tau^{d,\dagger}-\tau^{d*}=(\check a^d-a_d^*)+\mathbf1_{m_d}\mu_d^\top(\check\beta-\beta_d^*).
\]
Using \eqref{eq:profileStability} and $\|\mu_d\|_2\le M_\mu$ gives
\begin{equation}\label{eq:originalProfileBounds}
 \|\tau^{d,\dagger}-\tau^{d*}\|_\infty\le(C_{\rm pr}+M_\mu)\|\zeta\|_2,\qquad \tau^{d,\dagger}_{j+1}-\tau^{d,\dagger}_j\ge g+r_0-2C_{\rm pr}\|\zeta\|_2.
\end{equation}
The mean shift cancels when taking adjacent threshold differences. Define
\begin{equation}\label{eq:meanPackingRadius}
 r_\mu=\min\{B-r_0,\ r_0/[2(C_{\rm pr}+M_\mu)],\ r_0/(4C_{\rm pr})\}>0.
\end{equation}
If $\|\zeta\|_2\le r_\mu$, the mapped thresholds have magnitudes at most $T-r_0/2$ and gaps at least $g+r_0/2$. Also $\|\check\beta\|_2\le B-r_0$. Thus the mapped minimizer is interior to $\Phi$. It minimizes the centered risk over a superset of the centered image of $\Phi$, so it also minimizes the original pooled risk. Finally,
\[
 \bar\beta=t\zeta,\qquad \|\bar\beta-\beta_t^*\|_1=(1-t)\|\zeta\|_1\le h.
\]
This verifies \texorpdfstring{\hyperlink{condition:C5}{C5}}{C5} with $C_{\rm pool}=1$. We will check $\|\zeta\|_2\le r_\mu$ for the target-shift constructions below.

The next three subsubsections prove the $\ell_2$ and $\ell_1$ estimation lower bounds for constructions~(\ref{constr:shared})--(\ref{constr:thresholds}), respectively. Section~\ref{app:sharedSparse} gives the shared-coefficient terms $\sqrt{s_1\log p/n}$ and $s_1\sqrt{\log p/n}$. Section~\ref{app:targetShift} gives the coefficient-shift terms $\sqrt{hb_0}\wedge h$ and $h$. Section~\ref{app:orderedThresholds} gives the threshold terms $\sqrt{(K_t-1)/n_t}$ and $(K_t-1)/\sqrt{n_t}$. Section~\ref{app:lowerCombination} then uses the same constructions for the excess-risk lower bound.

\subsubsection{Estimation Lower Bounds for the Shared Coefficient}\label{app:sharedSparse}
We first record the finite sets of binary vectors used here and in Section~\ref{app:targetShift}. Let $\alpha\in(0,1)$ be the exponent in \eqref{eq:lowerRegime}. For an integer $k$ in the range $1\le k\le cp^\alpha$, there is a set $\Omega_k\subseteq\{0,1\}^p$ such that
\[
 \|\omega\|_0=k,\qquad d_{\rm Ham}(\omega,\omega')\ge k/2\ \ (\omega\ne\omega'),\qquad \log|\Omega_k|\ge \tfrac12 k\log(ep/k).
\]
One construction chooses vectors successively from all $\binom pk$ supports and removes those within Hamming distance $k/2$ of each chosen vector. For a fixed support, the number removed is at most
\[
 \sum_{r=0}^{\lfloor k/4\rfloor}\binom kr\binom{p-k}{r}\le C^k(p/k)^{k/4}.
\]
Indeed, $r$ coordinates are deleted and $r$ are added, so the Hamming distance is $2r$. The term for $r=0$ is one. For $r\ge1$, use $\binom kr\le2^k$ and $\binom{p-k}{r}\le(ep/r)^r$, and sum over $r\le k/4$. Since $\binom pk\ge(p/k)^k$, the number selected is at least $C^{-k}(p/k)^{3k/4}$. Taking the constant $c$ in \eqref{eq:lowerRegime} sufficiently small makes $p/k$ sufficiently large and gives the displayed cardinality bound. The same choice makes $\tfrac12\log(e/c)\ge\log3$, so $|\Omega_k|\ge3$ for every such $k$. Also $1\le s_1\le cp^\alpha$ implies $p\ge c^{-1/\alpha}\ge3$. Thus all finite families used with \citet[Theorem 2.5]{np-tsybakov-book} below have at least three members. Their observation distributions are mutually absolutely continuous because the Gaussian covariate densities and the conditional class probabilities are positive.

Use construction~(\ref{constr:shared}) with the index set $\Omega_{s_1}$. The squared coefficient norm is $s_1\gamma^2$. Condition \eqref{eq:lowerRegime} makes it at most $(B-r_0)^2$ when $c_0$ and the constant in that condition are sufficiently small. The coefficients have the required sparsity and zero discrepancy, and Section~\ref{app:packingMembership} verifies \texorpdfstring{\hyperlink{condition:C5}{C5}}{C5}. Hence every $\theta^\omega$ belongs to $\Theta$.

For distinct indices, $\|\beta_t^\omega-\beta_t^{\omega'}\|_2\ge\gamma\sqrt{s_1/2}$ and $\|\beta_t^\omega-\beta_t^{\omega'}\|_1\ge\gamma s_1/2$. On the other hand, the sample KL identity and \eqref{eq:KLequiv} give
\[
 \KL(\Pp_{\theta^\omega},\Pp_{\theta^{\omega'}})\le Cn\gamma^2d_{\rm Ham}(\omega,\omega')\le Cn s_1\gamma^2=Cc_0^2s_1\log(ep/s_1).
\]
Choosing $c_0$ sufficiently small makes this upper bound at most $\frac1{16}\log(|\Omega_{s_1}|-1)$. The separation and KL conditions of \citet[Theorem 2.5]{np-tsybakov-book} therefore hold. Applied with each of the distances $\|\psi-\psi'\|_2$ and $\|\psi-\psi'\|_1$, that theorem gives a positive lower bound on the probability of an error of size $c\gamma\sqrt{s_1}$ and $c\gamma s_1$, respectively. The inequality $\E Z\ge a\Pp(Z\ge a)$ for $Z\ge0$ and $a>0$ then gives
\[
 \inf_{\hat\psi}\sup_{\theta\in\Theta}\E_\theta\|\hat\psi-\psi(\theta)\|_2\ge c\sqrt{s_1\log p/n},\qquad \inf_{\hat\psi}\sup_{\theta\in\Theta}\E_\theta\|\hat\psi-\psi(\theta)\|_1\ge cs_1\sqrt{\log p/n}.
\]
Here $s_1\le cp^\alpha$ implies $\log(ep/s_1)\asymp\log p$.

\subsubsection{Estimation Lower Bounds for the Coefficient Shift}\label{app:targetShift}
Use construction~(\ref{constr:shift}). Only the $n_t$ target observations contribute to the sample KL divergences because the source distribution is the same for all candidates.

First suppose $h\ge b_0$, and take $\omega\in\Omega_k$ with $k,\gamma$ defined in construction~(\ref{constr:shift}).
The condition $h/b_0\le cp^\alpha$ in \eqref{eq:lowerRegime} places $k$ in the range of the finite-set construction above. If $k=1$, then $k\gamma\le h$ for $c_2\le1$; if $k>1$, then $k\gamma\le c_1c_2h$. Thus the coefficient discrepancy is at most $h$. Also $k\gamma^2\le C h b_0$. Since $h\ge b_0$, the minimum $hb_0\wedge h^2$ in \eqref{eq:lowerRegime} is $hb_0$. Taking its small constant sufficiently small gives $\|\beta_t^\omega\|_2\le r_\mu$. Section~\ref{app:packingMembership} then verifies \texorpdfstring{\hyperlink{condition:C5}{C5}}{C5} and all candidates belong to $\Theta$.

The target distances are at least $c\gamma\sqrt k\asymp\sqrt{hb_0}$ in $\ell_2$ and $c\gamma k\asymp h$ in $\ell_1$. For the sample distributions,
\[
 \KL(\Pp_{\theta^\omega},\Pp_{\theta^{\omega'}})\le Cn_tk\gamma^2=Cc_2^2 k\log p.
\]
Moreover, $\log|\Omega_k|\ge ck\log(ep/k)\asymp k\log p$. Choosing $c_2$ sufficiently small verifies the KL condition of \citet[Theorem 2.5]{np-tsybakov-book}. Its separation condition is given by the target distances just calculated. Using $\E Z\ge a\Pp(Z\ge a)$ again yields lower bounds $c\sqrt{hb_0}$ and $ch$.

Next suppose $0<h<b_0$. Use $p$ target alternatives $\beta_t^j=c_2h e_j$, $j\in[p]$. Their pairwise distances are $\sqrt2c_2h$ and $2c_2h$ in $\ell_2$ and $\ell_1$. Condition \eqref{eq:lowerRegime} now bounds $h^2$, so the candidates again satisfy $\|\beta_t^j\|_2\le r_\mu$ and belong to $\Theta$. Their sample KL divergences satisfy
\[
 \KL(\Pp_{\theta^j},\Pp_{\theta^{j'}})\le Cn_tc_2^2h^2\le Cc_2^2\log p.
\]
\citet[Theorem 2.5]{np-tsybakov-book}, with $c_2$ sufficiently small, gives lower bounds $ch$ for both expected norm errors when $0<h<b_0$. For $h=0$, the shift contribution is zero. Combining the regimes gives
\[
 \inf_{\hat\psi}\sup_{\theta\in\Theta}\E_\theta\|\hat\psi-\psi(\theta)\|_2\ge c(\sqrt{hb_0}\wedge h),\qquad \inf_{\hat\psi}\sup_{\theta\in\Theta}\E_\theta\|\hat\psi-\psi(\theta)\|_1\ge ch.
\]

\subsubsection{Estimation Lower Bounds for the Varying Target Thresholds}\label{app:orderedThresholds}
Use construction~(\ref{constr:thresholds}) and put $m=K_t-1$. Since the baseline vector $\tau^{t,0}$ is interior to the true-parameter threshold set, there is a fixed $\delta_\tau>0$ such that $|\tau_j^{t,0}|\le T-r_0-\delta_\tau$ and $\tau_{j+1}^{t,0}-\tau_j^{t,0}\ge g+r_0+2\delta_\tau$ for all applicable indices. Consider
\begin{equation}\label{eq:orderedpacking}
 \tau^{t,\omega}=\tau^{t,0}+\gamma\omega,\qquad \omega\in\{0,1\}^m,\qquad \gamma=c_3/\sqrt{n_t}.
\end{equation}
Choose $0<c_3<\delta_\tau/4$. Since $n_t\ge1$, each perturbation has magnitude at most $\delta_\tau/4$ and changes an adjacent gap by at most $\delta_\tau/4$. Thus every candidate belongs to the true-parameter threshold set. With both coefficient vectors equal to zero, Section~\ref{app:packingMembership} verifies \texorpdfstring{\hyperlink{condition:C5}{C5}}{C5} as well.

For neighboring indices with $d_{\rm Ham}(\omega,\omega')=1$, the source distributions coincide and \eqref{eq:KLequiv} gives
\[
 \KL(\Pp_{\theta^\omega},\Pp_{\theta^{\omega'}})\le Cn_t\gamma^2=Cc_3^2.
\]
Pinsker's inequality, \citet[Lemma 2.5(i)]{np-tsybakov-book}, therefore gives $\operatorname{TV}(\Pp_{\theta^\omega},\Pp_{\theta^{\omega'}})\le1/2$ when $c_3$ is sufficiently small. Apply \citet[Theorem 2.12(ii)]{np-tsybakov-book} to these full-sample distributions. It follows that every estimator $\hat\omega$ taking values in $\{0,1\}^m$ satisfies
\[
 \max_{\omega\in\{0,1\}^m}\E_{\theta^\omega}d_{\rm Ham}(\hat\omega,\omega)\ge m/4.
\]
For any estimator $\hat\tau\in\R^m$ of the target thresholds, define $\hat\omega_j=\ind\{\hat\tau_j>\tau_j^{t,0}+\gamma/2\}$ and let $N_{\rm err}=d_{\rm Ham}(\hat\omega,\omega)$. Each incorrectly selected coordinate has $|\hat\tau_j-\tau_j^{t,\omega}|\ge\gamma/2$. Hence
\[
 \|\hat\tau-\tau^{t,\omega}\|_1\ge\gamma N_{\rm err}/2,\qquad \|\hat\tau-\tau^{t,\omega}\|_2\ge\gamma\sqrt{N_{\rm err}}/2\ge\gamma N_{\rm err}/(2\sqrt m).
\]
Taking expectations and then the maximum over $\omega$ gives lower bounds $cm/\sqrt{n_t}$ and $c\sqrt{m/n_t}$ for the threshold $\ell_1$ and $\ell_2$ errors, respectively. Since the norm error of the full target parameter is at least its threshold-block error and $m=K_t-1$, we obtain
\[
 \inf_{\hat\psi}\sup_{\theta\in\Theta}\E_\theta\|\hat\psi-\psi(\theta)\|_2\ge c\sqrt{(K_t-1)/n_t},\qquad \inf_{\hat\psi}\sup_{\theta\in\Theta}\E_\theta\|\hat\psi-\psi(\theta)\|_1\ge c(K_t-1)/\sqrt{n_t}.
\]

We now combine the parameter-estimation lower bounds in Sections~\ref{app:sharedSparse}--\ref{app:orderedThresholds}. Each finite parameter family is contained in $\Theta$, so the minimax error for each norm is bounded below by the maximum of its three contributions:
\[
 \begin{aligned}
 \inf_{\hat\psi}\sup_{\theta\in\Theta}\E_\theta\|\hat\psi-\psi(\theta)\|_2&\ge c\left\{\sqrt{\frac{s_1\log p}{n}}+(\sqrt{hb_0}\wedge h)+\sqrt{\frac{K_t-1}{n_t}}\right\},\\
 \inf_{\hat\psi}\sup_{\theta\in\Theta}\E_\theta\|\hat\psi-\psi(\theta)\|_1&\ge c\left\{s_1\sqrt{\frac{\log p}{n}}+h+\frac{K_t-1}{\sqrt{n_t}}\right\}.
 \end{aligned}
\]
Substituting $b_0=\sqrt{\log p/n_t}$ proves \eqref{eq:lowerRates} in the main paper and completes the proof of Theorem~\ref{thm3}.

\subsubsection{Excess Risk Lower Bound for Theorem~\ref{thm4}}\label{app:lowerCombination}
We now use the same parameter families for an arbitrary data-dependent predictor $\hat q:\R^p\to\Delta^{K_t-1}$, with components $\hat q_y(x)$ for $y\in[K_t]$. Combine $\hat q$ with the fixed target covariate distribution to define $\widehat P_t(dx,\{y\})=\hat q_y(x)Q_t(dx)$. For each candidate $\theta^\omega$,
\[
 \E_X\KL\{\pi_t(X;\theta^\omega),\hat q(X)\}=\KL(P_{t,\theta^\omega},\widehat P_t)\ge H^2(P_{t,\theta^\omega},\widehat P_t),
\]
where the last inequality is \citet[Lemma 2.4]{np-tsybakov-book}. These are identities and inequalities conditional on the training observations. The remaining expectation in the minimax risk is over those observations.

For a given finite family, choose $\hat\omega$ to minimize $H(P_{t,\theta^\omega},\widehat P_t)$, with a fixed rule for ties. The triangle inequality and the minimizing property imply
\[
 H(P_{t,\theta^\omega},P_{t,\theta^{\hat\omega}})\le H(P_{t,\theta^\omega},\widehat P_t)+H(\widehat P_t,P_{t,\theta^{\hat\omega}})\le2H(P_{t,\theta^\omega},\widehat P_t).
\]
For the shared-coefficient family, \eqref{eq:Hequiv} gives squared target Hellinger separation at least $c s_1\gamma^2\asymp s_1\log p/n$. For the coefficient-shift families, the squared separations are at least $ck\gamma^2\asymp hb_0$ when $h\ge b_0$, and $ch^2$ when $0<h<b_0$. Sections~\ref{app:sharedSparse} and \ref{app:targetShift} verify the corresponding full-sample KL bounds. For each family, \citet[Theorem 2.5]{np-tsybakov-book} gives a positive lower bound on the worst-case probability of $\{\hat\omega\ne\omega\}$. On this event, the displayed triangle inequality bounds the squared Hellinger prediction error below by one quarter of the corresponding squared separation. Taking expectations gives the excess-risk contributions $c s_1\log p/n$ and $c(hb_0\wedge h^2)$, respectively. For $h=0$, the shift contribution is zero.

For construction~(\ref{constr:thresholds}), put $m=K_t-1$. Equation \eqref{eq:Hequiv} gives
\[
 H^2(P_{t,\theta^\omega},P_{t,\theta^{\omega'}})\ge c\gamma^2d_{\rm Ham}(\omega,\omega').
\]
Combining this separation bound with the triangle inequality above gives $H^2(P_{t,\theta^\omega},\widehat P_t)\ge c\gamma^2d_{\rm Ham}(\hat\omega,\omega)$. The neighboring full-sample total variation distances are at most $1/2$, as proved in Section~\ref{app:orderedThresholds}. Thus \citet[Theorem 2.12(ii)]{np-tsybakov-book} yields
\[
 \max_{\omega\in\{0,1\}^m}\E_{\theta^\omega}H^2(P_{t,\theta^\omega},\widehat P_t)\ge c\gamma^2m=c(K_t-1)/n_t.
\]
This proves the third excess-risk contribution for all predictors taking values in the simplex.

Each finite parameter family is contained in $\Theta$. Taking the maximum of the three excess-risk contributions gives
\begin{equation}\label{eq:combineLower}
 \inf_{\hat q}\sup_{\theta\in\Theta}\E_\theta\E_X\KL\{\pi_t(X;\theta),\hat q(X)\}\ge c\left\{\frac{s_1\log p}{n}+(hb_0\wedge h^2)+\frac{K_t-1}{n_t}\right\}.
\end{equation}
Substituting $b_0=\sqrt{\log p/n_t}$ proves \eqref{eq:excess_risk_lower} in the main paper and completes the proof of the lower bound in Theorem~\ref{thm4}.
\end{proof}

\section{Proof of the Upper Bounds in Theorem~\ref{thm4}}\label{app:risk}
\begin{proof}
Write $\psi=(\beta^\top,\tau^{t\top})^\top\in\R^{p+K_t-1}$ for a target parameter and $\psi^*=(\beta_t^{*\top},\tau^{t*\top})^\top$ for its true value. Recall that the estimator $\hat\psi$ belongs to $\Psi=\mathcal B(B)\times\mathcal T_{K_t}(T,g)$. For $y\in[K_t]$, let $\pi_{t,y}(x;\psi)$ denote the conditional probability of $Y^t=y$ given $X=x$ under $\psi$. Let $P_X$ be the target covariate distribution and define the joint distribution
\[
 P_\psi(dx,\{y\})=\pi_{t,y}(x;\psi)P_X(dx).
\]
The marginal distribution $P_X$ is the same for all $\psi$.

Conditional on the training observations, $\hat\psi$ is fixed. Let $(X,Y^t)$ be a new target observation, independent of the training observations, with distribution $P_{\psi^*}$. Write $\E_X$ for expectation under $P_X$. The risk in \eqref{eq:populationrisk} of the main paper is $R_t(\psi)=\E[-\log\pi_{t,Y^t}(X;\psi)]$, where the expectation is over this new observation. Hence
\[
 \mathcal E_t(\hat\psi)=R_t(\hat\psi)-R_t(\psi^*)=\E_X\sum_{y=1}^{K_t}\pi_{t,y}(X;\psi^*)\log\frac{\pi_{t,y}(X;\psi^*)}{\pi_{t,y}(X;\hat\psi)}=\KL(P_{\psi^*},P_{\hat\psi}).
\]
Conditions \texorpdfstring{\hyperlink{condition:C2}{C2}}{C2} and \texorpdfstring{\hyperlink{condition:C3}{C3}}{C3} place $\psi^*$ in $\mathcal B(B-r_0)\times\mathcal T_{K_t}(T-r_0,g+r_0)$. Applying the upper bound in \eqref{eq:KLequiv} with $d=t$, $\xi_0=\psi^*$ and $\xi_1=\hat\psi$ gives, for every realization of the training observations,
\begin{equation}\label{eq:riskParameterComparison}
 0\le\mathcal E_t(\hat\psi)\le C\|\hat\psi-\psi^*\|_2^2,
\end{equation}
where $C$ is independent of $n_s,n_t,p,s_1,h,\varepsilon$.

Under the conditions for the probability bounds in Theorem~\ref{thm2}, \eqref{eq:targetSquared} holds on $\eventref{5}\cap\eventref{6}$ with probability at least $1-2\varepsilon$. Substitute the definitions of $a,b,q_t$ into that bound and use $n_{\rm p}\asymp n$ and $n_{t,2}\asymp n_t$. Equation \eqref{eq:riskParameterComparison} then yields
\[
 \mathcal E_t(\hat\psi)\le C\left\{\frac{s_1\log(p/\varepsilon)}{n}+\left(h\sqrt{\frac{\log(p/\varepsilon)}{n_t}}\wedge h^2\right)+\frac{(K_t-1)\log(1/\varepsilon)}{n_t}\right\}
\]
on the same event. This proves \eqref{eq:excess_risk_upper}.

For the expected bound, use the penalties and sample size conditions in Section~\ref{app:expectedBounds}. Taking expectations over the training observations in \eqref{eq:riskParameterComparison} and applying \eqref{eq:expectedSquared} gives
\[
 \E\mathcal E_t(\hat\psi)\le C\E\|\hat\psi-\psi^*\|_2^2\le C\{s_1a_E^2+(hb_E\wedge h^2)+(K_t-1)/n_{t,2}\}.
\]
Since $a_E^2\asymp\log\{p(n_t+1)\}/n$ and $b_E^2\asymp\log\{p(n_t+1)\}/n_t$, it follows that
\[
 \E\mathcal E_t(\hat\psi)\le C\left\{\frac{s_1\log\{p(n_t+1)\}}{n}+\left(h\sqrt{\frac{\log\{p(n_t+1)\}}{n_t}}\wedge h^2\right)+\frac{K_t-1}{n_t}\right\}.
\]
This proves \eqref{eq:expectedRisk}. The lower bound \eqref{eq:excess_risk_lower} was proved in Section~\ref{app:lowerCombination}.
\end{proof}

\section{Additional Illustration and Extensions}\label{app:extensions}
In this section, we prove the two sufficient conditions for \texorpdfstring{\hyperlink{condition:C5}{C5}}{C5} stated in Remark~\ref{rem1}. We then provide the upper and lower bounds for an alternative version which imposes sparsity on the target coefficient $\beta_t$. Finally we prove Proposition~\ref{prop:fullTarget} for the full-target algorithm.

\subsection{Profile Hessians and Pooling Stability}
This section gives a sufficient condition for \texorpdfstring{\hyperlink{condition:C5}{C5}}{C5}. We first minimize the population loss over the thresholds. The Hessian of the resulting function gives a bound on the pooled coefficient bias.

\begin{proposition}[Condition on the Profile Hessians]\label{prop:profilebias}
Assume \texorpdfstring{\hyperlink{condition:C1}{C1}}{C1}--\texorpdfstring{\hyperlink{condition:C4}{C4}}{C4}, and let $\bar\phi$ minimize $R_{\rm pool}$ over $\Phi$. For each domain, let $R_d(\beta,\tau)=\E L_d(\beta,\tau)$. For fixed $\beta\in\mathcal B(B)$, let $\tau_d(\beta)$ be the threshold vector that minimizes $R_d$. Define the profiled population risk by
\[
 \tau_d(\beta)=\arg\min_{\tau\in\mathcal T_{K_d}(T,g)}R_d(\beta,\tau),\qquad r_d(\beta)=R_d(\beta,\tau_d(\beta)).
\]
Let $\operatorname{int}$ denote the interior of a set, and define
\[
 \mathcal U_d=\{\beta\in\operatorname{int}\mathcal B(B):\tau_d(\beta)\in\operatorname{int}\mathcal T_{K_d}(T,g)\}.
\]
The set \ $\mathcal U_d$ contains coefficient vectors for which both the coefficient and minimizing threshold vector lie in the interiors of their parameter spaces. Suppose $\bar\beta\in\operatorname{int}\mathcal B(B)$ and, for every $u\in[0,1]$, 
\[
 \beta_t^*+u(\bar\beta-\beta_t^*)\in\mathcal U_s\cap\mathcal U_t,\qquad \beta_t^*+u(\beta_s^*-\beta_t^*)\in\mathcal U_s.
\]
For $\beta\in\mathcal U_d$, partition the population Hessian $H^d=\nabla^2R_d(\beta,\tau_d(\beta))$ into coefficient and threshold blocks. The blocks $H_{\beta\beta}^d$, $H_{\beta\tau}^d$, $H_{\tau\beta}^d$ and $H_{\tau\tau}^d$ have dimensions $p\times p$, $p\times (K_d-1)$, $(K_d-1)\times p$ and $(K_d-1)\times (K_d-1)$, respectively. On $\mathcal U_d$, define the profiled coefficient Hessian $M_d(\beta)\in\R^{p\times p}$ by
\begin{equation}\label{eq:profileH}
 M_d(\beta)=\nabla^2r_d(\beta)=H_{\beta\beta}^d-H_{\beta\tau}^d(H_{\tau\tau}^d)^{-1}H_{\tau\beta}^d.
\end{equation}
With the pooling weights $\rho_s,\rho_t$, let
\begin{equation}\label{eq:profileMatrices}
 \overline M=\sum_{d=s,t}\rho_d\int_0^1M_d(\beta_t^*+u(\bar\beta-\beta_t^*))\,du,\qquad M_s^*=\int_0^1M_s(\beta_t^*+u(\beta_s^*-\beta_t^*))\,du.
\end{equation}
If $\|\overline M^{-1}M_s^*\|_1\le C_{\rm pool}$, then \texorpdfstring{\hyperlink{condition:C5}{C5}}{C5} holds, with the more precise bound
\[
 \|\bar\beta-\beta_t^*\|_1\le \rho_s C_{\rm pool}h.
\]
\end{proposition}
\begin{proof}
The population Hessian bound proved in Section~\ref{app:KLHellinger} gives fixed $c_H,C_H>0$ such that $c_HI\preceq\nabla^2R_d\preceq C_HI$ on the parameter space. In particular, the threshold block is invertible and the threshold minimizer is unique. If $\beta\in\mathcal U_d$, it satisfies $\nabla_\tau R_d(\beta,\tau_d(\beta))=0$. Compactness and uniqueness imply continuity of $\tau_d$: any convergent subsequence of minimizing threshold vectors must minimize the limiting population loss. The minimizing thresholds remain interior in a neighborhood of each $\beta\in\mathcal U_d$. The implicit function theorem therefore gives
\[
 D\tau_d(\beta)=-(H_{\tau\tau}^d)^{-1}H_{\tau\beta}^d,\qquad \nabla r_d(\beta)=\nabla_\beta R_d(\beta,\tau_d(\beta)).
\]
Here $D\tau_d(\beta)$ is the Jacobian of the threshold minimizer. Differentiating $\nabla r_d(\beta)$ gives \eqref{eq:profileH}. The Schur complement in that equation is bounded below by $c_HI$ and above by $C_HI$. Indeed, its quadratic form is obtained by minimizing the joint Hessian quadratic form over the threshold direction.

Correct specification and identifiability give $\tau_d(\beta_d^*)=\tau^{d*}$ and $\nabla r_d(\beta_d^*)=0$. The pooled optimum has $\bar\tau^d=\tau_d(\bar\beta)$ and satisfies $\sum_d \rho_d\nabla r_d(\bar\beta)=0$. The two line segments lie in the sets $\mathcal U_d$ specified in the proposition. Integrating their Hessians gives
\[
 \overline M(\bar\beta-\beta_t^*)=-\rho_s\nabla r_s(\beta_t^*)=\rho_s M_s^*(\beta_s^*-\beta_t^*).
\]
The matrices $M_d(\beta)$ are positive definite, so $\overline M$ is invertible. Using $\|Av\|_1\le\|A\|_1\|v\|_1$ and the discrepancy bound $\|\beta_s^*-\beta_t^*\|_1\le h$ in \eqref{eq:shift} of the main paper gives
\[
 \begin{aligned}\|\bar\beta-\beta_t^*\|_1&=\rho_s\|\overline M^{-1}M_s^*(\beta_s^*-\beta_t^*)\|_1\\&\le\rho_s\|\overline M^{-1}M_s^*\|_1\|\beta_s^*-\beta_t^*\|_1\le\rho_s C_{\rm pool}h.\end{aligned}
\] Since $\bar\beta\in\mathcal U_s\cap\mathcal U_t$, both minimizing threshold vectors are interior. Thus $\bar\phi$ is interior to $\Phi$, which completes the verification of \texorpdfstring{\hyperlink{condition:C5}{C5}}{C5}.
\end{proof}

\paragraph{A sufficient condition for the matrix bound.}
Suppose $\overline M$ and $M_s^*$ are block diagonal for a common partition of the coefficient coordinates, with at most $q$ coordinates per block, where $q$ is fixed independently of $p$. The population curvature bounds in the proof of Proposition \ref{prop:profilebias} give
\[
 \|\overline M^{-1}M_s^*\|_1\le\sqrt q\,\|\overline M^{-1}M_s^*\|_{\rm op}\le C\sqrt q.
\]
Hence $C_{\rm pool}\ge\max\{1,C\sqrt q\}$ gives the required matrix norm bound. The case $q=1$ corresponds to diagonal matrices.

\subsection{A Gaussian Sufficient Condition for C5}
\begin{example}[A Gaussian sufficient condition for C5]\label{ex:C5Gaussian}
Assume \texorpdfstring{\hyperlink{condition:C1}{C1}}{C1}--\texorpdfstring{\hyperlink{condition:C4}{C4}}{C4}. Suppose $X^{(d)}\sim N(\mu_d,\sigma_d^2I_p)$ for $d=s,t$, with the means and positive variances satisfying \texorpdfstring{\hyperlink{condition:C1}{C1}}{C1}, and let
\[
 \beta_s^*=\alpha_s v,\qquad \beta_t^*=\alpha_t v,\qquad v\in\R^p,\quad\|v\|_2=1,\quad\alpha_s,\alpha_t\in\R.
\]
There is a fixed $r_{\rm dir}>0$, independent of $n_s,n_t,p,s_1,h$, such that if $\|\beta_s^*-\beta_t^*\|_2\le r_{\rm dir}$, the pooled population minimizer lies in the interior of $\Phi$ and has coefficient vector
\[
 \bar\beta=\bar\alpha v,\qquad \bar\alpha\in[\min(\alpha_s,\alpha_t),\max(\alpha_s,\alpha_t)].
\]
Thus $\|\bar\beta-\beta_t^*\|_1\le\|\beta_s^*-\beta_t^*\|_1\le h$, and \texorpdfstring{\hyperlink{condition:C5}{C5}}{C5} holds with $C_{\rm pool}=1$. The two coefficient vectors are collinear, but the means, spherical variances, links, class counts and threshold vectors may differ between domains.
\end{example}
\begin{proof}
Use the centering identity \eqref{eq:populationCentering} and the larger parameter set \eqref{eq:auxiliaryPoolSpace}. Here $Z^{(d)}\sim N(0,\sigma_d^2I_p)$ and $a_d^*=\tau^{d*}-\mathbf1_{K_d-1}\mu_d^\top\beta_d^*$. The true centered thresholds belong to $\mathcal T_{K_d}(\widetilde T-r_0,g+r_0)$. The centered pooled population risk is strictly convex, so it has a unique minimizer $(\check\beta,\check a^s,\check a^t)$.

Orthogonal transformations fixing $v$ preserve the centered covariate distributions, the response distributions and the auxiliary set. Uniqueness implies $\check\beta=\bar\alpha v$. As proved for the scalar profiles in Section~\ref{app:packingMembership}, partial minimization over the convex threshold sets preserves strict convexity. The source and target scalar profiles have their minima at $\alpha_s$ and $\alpha_t$, respectively, so their positively weighted sum is minimized between these values. Therefore, with $\Delta=\|\beta_s^*-\beta_t^*\|_2$,
\[
 \|\check\beta\|_2\le B-r_0,\qquad \|\check\beta-\beta_d^*\|_2\le\Delta,\quad d=s,t.
\]
For each domain, compare the minimizing thresholds with $a_d^*$ at the coefficient $\check\beta$. Proposition \ref{prop:KL}, applied to the centered covariates and enlarged threshold bound, gives $\|\check a^d-a_d^*\|_2\le C_{\rm pr}\Delta$ for a fixed $C_{\rm pr}\ge1$. The bound does not depend on the positive pooling weights.

Return to the original thresholds by setting $\tau^{d,\dagger}=\check a^d+\mathbf1_{K_d-1}\mu_d^\top\check\beta$. As in \eqref{eq:originalProfileBounds},
\[
 \|\tau^{d,\dagger}-\tau^{d*}\|_\infty\le(C_{\rm pr}+M_\mu)\Delta,\qquad \tau^{d,\dagger}_{j+1}-\tau^{d,\dagger}_j\ge g+r_0-2C_{\rm pr}\Delta.
\]
Choose
\[
 r_{\rm dir}=\min\{r_0/[2(C_{\rm pr}+M_\mu)],\ r_0/(4C_{\rm pr})\}>0.
\]
For $\Delta\le r_{\rm dir}$, the mapped point is interior to the original $\Phi$, with threshold magnitudes at most $T-r_0/2$ and gaps at least $g+r_0/2$. It minimizes the centered risk over a superset of the centered image of $\Phi$, and hence is the original pooled minimizer. Finally,
\[
 \|\bar\beta-\beta_t^*\|_1=|\bar\alpha-\alpha_t|\|v\|_1\le|\alpha_s-\alpha_t|\|v\|_1=\|\beta_s^*-\beta_t^*\|_1\le h.
\]
\end{proof}

\subsection{Target-Sparse Minimax Bounds}\label{app:targetSparse}
The main paper assumes that the source coefficient is sparse, $\|\beta_s^*\|_0\le s_1$, and that $\|\beta_t^*-\beta_s^*\|_1\le h$. This section places the sparsity condition on the target coefficient, $\|\beta_t^*\|_0\le s_1$, while retaining the same discrepancy bound. This is the sparsity formulation used by \citet[Section 2.2]{tian2023}. The following result gives the corresponding estimation and excess-risk bounds for comparison.
\begin{corollary}[Target-Sparse Minimax Bounds]\label{cor:targetSparse}
Let $\Theta_t(s_1,h)$ be the class of two-domain parameters satisfying $\|\beta_t\|_0\le s_1$, $\|\beta_t-\beta_s\|_1\le h$, the coefficient and threshold bounds in \eqref{eq:parameterSpace}, and \texorpdfstring{\hyperlink{condition:C5}{C5}}{C5}. For $\theta\in\Theta_t(s_1,h)$, let $\psi(\theta)$ denote its target parameter, as in the main paper. The upper bounds in Lemma~\ref{thm1} and Theorems~\ref{thm2} and \ref{thm4} hold over this class with the same penalties and sample-size conditions.

For each fixed pair $X^{(d)}\sim N(\mu_d,I_p)$, $d=s,t$, with $\|\mu_d\|_2\le M_\mu$, put $b_0=\sqrt{\log p/n_t}$. Suppose $s_1\le cp^\alpha$ for a fixed $\alpha\in(0,1)$, $s_1\log p/n\le c(B-r_0)^2$ and $(s_1b_0^2\wedge hb_0\wedge h^2)\le c(B-r_0)^2$. Assume also that $c_3/\sqrt{n_t}$ is less than one quarter of the fixed threshold margin used in Section~\ref{app:orderedThresholds}. Then
\begin{align}\label{eq:targetSparseLower}
 \inf_{\hat\psi}\sup_{\theta\in\Theta_t}\E_\theta\|\hat\psi-\psi(\theta)\|_2&\ge c\{\sqrt{s_1\log p/n}+(\sqrt{s_1}b_0\wedge\sqrt{hb_0}\wedge h)+\sqrt{(K_t-1)/n_t}\},\\
 \inf_{\hat\psi}\sup_{\theta\in\Theta_t}\E_\theta\|\hat\psi-\psi(\theta)\|_1&\ge c\{s_1\sqrt{\log p/n}+(s_1b_0\wedge h)+(K_t-1)/\sqrt{n_t}\}.\nonumber
\end{align}
The minimax excess-risk lower bound is
\begin{equation}\label{eq:targetSparseRisk}
 \inf_{\hat q}\sup_{\theta\in\Theta_t}\E_\theta\E_X\KL\{\pi_t(X;\theta),\hat q(X)\}\ge c\{s_1\log p/n+(s_1b_0^2\wedge hb_0\wedge h^2)+(K_t-1)/n_t\}.
\end{equation}
Under the conditions for the expected bounds in Theorem~\ref{thm2}, $h\le s_1b_0$ and $\log n_t\lesssim\log p$, the upper bounds match these lower bounds. 
\end{corollary}
\begin{proof}
For the upper bounds take $S=\supp(\beta_t^*)$. \texorpdfstring{\hyperlink{condition:C5}{C5}}{C5} gives $\|\bar\beta_{S^c}\|_1\le\|\bar\beta-\beta_t^*\|_1\le C_{\rm pool}h$. This is the bound on the coefficients outside $S$ needed in \eqref{eq:poolBI}; the pooling and conditional correction arguments therefore give the stated upper bounds.

The shared-coefficient and ordered-threshold packings in Section~\ref{app:lower} satisfy target sparsity and give the first and third lower-bound components. For the shift component set $\beta_s=0_p$. If $0<h<b_0$, use the one-coordinate alternatives $\beta_t=c_2h e_j$, giving norm separations of order $h$ and squared Hellinger separation of order $h^2$. For $h\ge b_0$, use a finite set of binary vectors with exactly $k$ nonzero coordinates, pairwise Hamming distance at least $ck$, and
\[
 k=\min\{s_1,\max(1,\lfloor c_1h/b_0\rfloor)\},\qquad \gamma=c_2b_0,
\]
where $c_1,c_2>0$ are sufficiently small. Set $\beta_t^\omega=\gamma\omega$ for these binary vectors $\omega$. Each candidate has at most $s_1$ nonzero coordinates and $\|\beta_t^\omega\|_1=k\gamma\le h$. Its squared coefficient norm satisfies
\[
 \|\beta_t^\omega\|_2^2=k\gamma^2\le Cc_2^2(s_1b_0^2\wedge hb_0)\le(B-r_0)^2,
\]
where the last inequality follows from the norm condition in the corollary and $h\ge b_0$. Thus the candidates belong to the coefficient ball $\mathcal B(B-r_0)$. The finite family has log cardinality at least $ck\log(ep/k)$, and its target-sample KL divergence is at most $Cn_tk\gamma^2$. Since $k\le s_1\le cp^\alpha$, \citet[Theorem 2.5]{np-tsybakov-book} gives $\ell_2$ and $\ell_1$ lower bounds of orders $\sqrt{s_1}b_0\wedge\sqrt{hb_0}$ and $s_1b_0\wedge h$, respectively. If $h=0$, the shift contribution is zero. The Hellinger argument gives the corresponding excess-risk lower bound. Equations \eqref{eq:auxiliaryPoolSpace}--\eqref{eq:meanPackingRadius} verify \texorpdfstring{\hyperlink{condition:C5}{C5}}{C5} and the interior threshold constraints for these parameters. Combining the three submodels gives the stated bounds.

If $h\le s_1b_0$, then
\[
 s_1b_0^2\wedge hb_0\wedge h^2=hb_0\wedge h^2,\qquad s_1b_0\wedge h=h.
\]
The coefficient-shift terms in the $\ell_2$ and excess-risk lower bounds therefore reduce to $\sqrt{hb_0}\wedge h$ and $hb_0\wedge h^2$, respectively, and the $\ell_1$ term reduces to $h$. These are the corresponding terms in the expected upper bounds when $\log n_t\lesssim\log p$.
\end{proof}

\subsection{Proof of Proposition~\ref{prop:fullTarget}}
\begin{proof}
Write $a_F=\lambda_{p,F}=A_\lambda\sqrt{\Lambda_1/n}$, $b_F=\lambda_{t,F}=A_\lambda\sqrt{\Lambda_1/n_t}$ and $q_F=A_\tau\sqrt{(K_t-1)\Lambda_2/n_t}$, with $\Lambda_1,\Lambda_2,A_\tau$ as in Section~\ref{app:pool}. The sample-size and radius conditions of Proposition~\ref{prop:fullTarget} make $s_1a_Fb_F+hb_F$ sufficiently small.

Apply the pooling gradient, restricted strong convexity (RSC) and basic-inequality arguments of Section~\ref{app:pool} with domain weights $n_s/n,n_t/n$ and all target observations in the pooled loss. The full-target population stability condition gives $\nabla R_F(\bar\phi_F)=0$ and the required bound on the coefficients outside the source support, because $\bar\phi_F$ is interior. The threshold bound is $A_\tau\sqrt{(K_s+K_t-2)\Lambda_2/n}$. Since $a_F\le b_F$, making $s_1a_Fb_F+hb_F$ sufficiently small also makes $s_1a_F^2$ and $ha_F$ sufficiently small. Let $\eventdef{10}$ be the intersection of the full-target counterparts of $\eventref{1}$ and $\eventref{3}$. Then $\Pp(\eventref{10})\ge1-\varepsilon$. On $\eventref{10}$,
\[
\|\widehat\beta_{p,F}-\bar\beta_F\|_1\le C(s_1a_F+h),\qquad H_F:=\|\beta_t^*-\widehat\beta_{p,F}\|_1\le C(s_1a_F+h).
\]

\Needspace{4\baselineskip}
Let $\eventdef{11}$ be the event on which the following gradient bounds hold at the true target parameter:
\[
\|\nabla_\beta L_t(\psi^*)\|_\infty\le b_F/2,\qquad \|\nabla_{\tau^t}L_t(\psi^*)\|_2\le q_F,
\]
and the following RSC bound holds for every $z=(z_\beta^\top,z_\tau^\top)^\top\in\R^{p+K_t-1}$ with $\psi^*+z\in\Psi$:
\[
D_{L_t}(z;\psi^*)\ge\kappa\|z\|_2^2-C_Rb_F^2\|z_\beta\|_1^2.
\]
Here $z_\beta\in\R^p$ and $z_\tau\in\R^{K_t-1}$. The gradient and uniform RSC arguments of Section~\ref{app:pool}, applied at the fixed parameter $\psi^*$ using all $n_t$ target observations, give $\Pp(\eventref{11})\ge1-\varepsilon$. Both $\eventref{10}$ and $\eventref{11}$ involve the full target sample. The union bound gives
\[
 \Pp(\eventref{10}\cap\eventref{11})\ge1-\Pp(\eventref{10}^c)-\Pp(\eventref{11}^c)\ge1-2\varepsilon.
\]

Work on $\eventref{10}\cap\eventref{11}$. Put $z=\widehat\psi_F-\psi^*$, $e=\|z\|_2$ and $u=\|z_\beta\|_1$. Optimality at the feasible comparator $\psi^*$ and the inequalities defining $\eventref{11}$ give
\[
D_{L_t}(z;\psi^*)\le\tfrac{b_F}{2}u+q_Fe+b_F\{H_F-\|\beta_t^*-\widehat\beta_{p,F}+z_\beta\|_1\}.
\]
The triangle inequality gives $H_F-\|\beta_t^*-\widehat\beta_{p,F}+z_\beta\|_1\le2H_F-u$. Therefore,
\[
D_{L_t}(z;\psi^*)+\tfrac{b_F}{2}u\le2b_FH_F+q_Fe,\qquad u\le4H_F+2(q_F/b_F)e.
\]
The second inequality uses convexity, which makes the Bregman increment nonnegative. Combining it with RSC yields
\[
\kappa e^2\le2b_FH_F+q_Fe+32C_Rb_F^2H_F^2+8C_Rq_F^2e^2.
\]
The sample size condition $\min(n_s,n_t)\ge C_n\Lambda_1$, with $C_n$ sufficiently large, gives $8C_Rq_F^2\le\kappa/4$. Also $b_FH_F\le C(s_1a_Fb_F+hb_F)$ is sufficiently small when $c$ is chosen sufficiently small, so $(b_FH_F)^2\le Cb_FH_F$. Young's inequality absorbs $q_Fe$ and proves
\[
e^2\le C(b_FH_F+q_F^2)\le C(s_1a_Fb_F+hb_F+q_F^2).
\]
The gradient bounds are evaluated at $\psi^*$, and the RSC bound holds simultaneously for all $z\in\R^{p+K_t-1}$ with $\psi^*+z\in\Psi$. The pooled estimator enters the basic inequality through the penalty comparison. The population KL bound in Proposition \ref{prop:KL} gives $\mathcal E_t(\widehat\psi_F)\le Ce^2$ and proves \eqref{eq:fullTargetRate}.

If $n_s/n_t\le M_n$ for a fixed $M_n>0$, then $b_F/a_F\le\sqrt{1+M_n}$ and $s_1a_Fb_F\le\sqrt{1+M_n}\,s_1a_F^2$. When $h\le b_F$, we have $hb_F\le b_F^2\le(1+M_n)s_1a_F^2$, since $s_1\ge1$; when $h>b_F$, we have $hb_F\wedge h^2=hb_F$. Hence, on $\eventref{10}\cap\eventref{11}$, whose probability is at least $1-2\varepsilon$,
\[
e^2+\mathcal E_t(\widehat\psi_F)\le C\{s_1a_F^2+(hb_F\wedge h^2)+q_F^2\},
\]
where $C$ is independent of $n_s,n_t,p,s_1,h,\varepsilon$ for each fixed $M_n$. The same inequalities also give $u\le C(s_1a_F+h+q_F^2/b_F)$. The threshold $\ell_1$ error is at most $\sqrt{K_t-1}\,e$. Since $K_t$ is fixed and $\Lambda_2\le\Lambda_1$, these bounds give the parameter-error bounds \eqref{eq:targetrates} and the excess-risk bound \eqref{eq:excess_risk_upper} in the main paper.

For the expected bounds, assume $n_s/n_t\le M_n$ and take
\[
\varepsilon_E=\{p(n_t+1)\}^{-4},\qquad \Lambda_E=\log(p/\varepsilon_E)\asymp\log\{p(n_t+1)\}.
\]
Use the penalties
\[
a_{F,E}=\lambda_{p,F}=A_\lambda\sqrt{\Lambda_E/n},\qquad b_{F,E}=\lambda_{t,F}=A_\lambda\sqrt{\Lambda_E/n_t},
\]
and impose the sample-size and radius conditions of Proposition~\ref{prop:fullTarget} with $\varepsilon=\varepsilon_E$. Construct $\eventref{10}$ with $\varepsilon=\varepsilon_E$. Define $\eventdef{12}$ as the intersection of this event and the event on which the following bounds hold:
\[
\begin{aligned}
\|\nabla_\beta L_t(\psi^*)\|_\infty&\le b_{F,E}/2,\\
D_{L_t}(z;\psi^*)&\ge\kappa\|z\|_2^2-C_Rb_{F,E}^2\|z_\beta\|_1^2.
\end{aligned}
\]
The second inequality holds simultaneously for all $z=(z_\beta^\top,z_\tau^\top)^\top\in\R^{p+K_t-1}$ such that $\psi^*+z\in\Psi$, with $z_\beta\in\R^p$ and $z_\tau\in\R^{K_t-1}$. The arguments defining $\eventref{10}$ and $\eventref{11}$ give $\Pp(\eventref{12}^c)\le C\varepsilon_E$. On $\eventref{12}$, we have $H_F\le C(s_1a_{F,E}+h)$.

Define the target threshold gradient norm by
\[
Q_F=\|\nabla_{\tau^t}L_t(\psi^*)\|_2.
\]
The gradient at the fixed true parameter has mean zero. Independence of the target observations and the uniform sub-Gaussian threshold-gradient bounds from Section~\ref{app:pool} give
\[
\E Q_F^2\le C(K_t-1)/n_t,\qquad \Pp(Q_F>q_0)\le Ce^{-cn_t},
\]
where $q_0>0$ is a sufficiently small fixed constant. Choose $q_0$ so that $8C_Rq_0^2\le\kappa/4$. Define $\eventdef{13}=\eventref{12}\cap\{Q_F\le q_0\}$. For sufficiently large $C_n$, the condition $n_t\ge C_n\Lambda_E$ gives $\Pp(\eventref{13}^c)\le C\varepsilon_E$.

On $\eventref{13}$, the basic inequality and RSC argument above apply with $b_{F,E}$ and $Q_F$ in place of $b_F$ and $q_F$. Thus
\[
e^2\le C(b_{F,E}H_F+Q_F^2),\qquad u\le4H_F+2(Q_F/b_{F,E})e\le C(H_F+Q_F^2/b_{F,E}).
\]
The last inequality follows by substituting the bound for $e$ and applying Young's inequality. Since $n_s/n_t\le M_n$, the same two cases $h\le b_{F,E}$ and $h>b_{F,E}$ yield
\[
e^2\le C\{s_1a_{F,E}^2+(hb_{F,E}\wedge h^2)+Q_F^2\}.
\]
Taking expectations on $\eventref{13}$ uses only the unconditional second-moment bound for $Q_F$. Compactness and the population KL upper bound give $e^2+\mathcal E_t(\widehat\psi_F)\le C$ everywhere, so the exceptional contribution is at most $C\varepsilon_E\le C/n_t$. Consequently,
\[
\E\|\widehat\psi_F-\psi^*\|_2^2+\E\mathcal E_t(\widehat\psi_F)\le C\{s_1a_{F,E}^2+(hb_{F,E}\wedge h^2)+(K_t-1)/n_t\}.
\]
Jensen's inequality gives the expected Euclidean-error bound. Substituting the penalties gives the full-target counterparts of the expected $\ell_2$ bound in \eqref{eq:expectedRates} and the expected excess-risk bound in \eqref{eq:expectedRisk} of the main paper.

For the $\ell_1$ error, let $\widehat\tau_F^t$ denote the threshold block of $\widehat\psi_F$. On $\eventref{13}$ we have
\[
u\le C\{s_1a_{F,E}+h+Q_F^2/b_{F,E}\}.
\]
Because $K_t$ is fixed, $s_1\ge1$ and $\sqrt{hb_{F,E}\wedge h^2}\le h$, the threshold block satisfies
\[
\|\widehat\tau_F^t-\tau^{t*}\|_1\le\sqrt{K_t-1}\,e\le C(s_1a_{F,E}+h)+C\sqrt{K_t-1}\,Q_F.
\]
Moreover, Cauchy--Schwarz and $\Lambda_E\ge1$ give
\[
\sqrt{K_t-1}\,\E Q_F\le C(K_t-1)/\sqrt{n_t},\qquad \E Q_F^2/b_{F,E}\le C(K_t-1)/\sqrt{n_t}.
\]
Compactness bounds the full $\ell_1$ error by $C\sqrt p$ on $\eventref{13}^c$, whose contribution is at most $C\sqrt p\,\varepsilon_E\le C/\sqrt{n_t}$. Combining the coefficient and threshold blocks proves
\[
\E\|\widehat\psi_F-\psi^*\|_1\le C\left\{s_1\sqrt{\frac{\log\{p(n_t+1)\}}{n}}+h+\frac{K_t-1}{\sqrt{n_t}}\right\}.
\]
This is the full-target counterpart of the expected $\ell_1$ bound in \eqref{eq:expectedRates} of the main paper.
\end{proof}

\end{document}